\documentclass[10pt,twocolumn,letterpaper]{article}

\usepackage{iccv}
\usepackage{times}
\usepackage{epsfig}
\usepackage{graphicx}
\usepackage{amsmath}
\usepackage{amssymb}

\usepackage{algorithm}
\usepackage{algorithmic}

\usepackage[pagebackref=false,breaklinks=true,letterpaper=true,colorlinks,citecolor=black,linkcolor=black,bookmarks=false]{hyperref}

\iccvfinalcopy 

\def\iccvPaperID{****} 

\ificcvfinal\fi
\begin{document}

\title{Video Deblurring via Semantic Segmentation and Pixel-Wise Non-Linear Kernel}

\author{Wenqi Ren$^{1,2}$, Jinshan Pan$^{3}$, Xiaochun Cao$^{1,4}$\thanks{Corresponding author.}, ~and Ming-Hsuan Yang$^{5}$\\
$^{1}$State Key Laboratory of Information Security(SKLOIS), IIE, CAS\\
$^{2}$Tencent AI Lab \\
$^{3}$School of Computer Science and Engineering, Nanjing University of Science and Technology\\
$^{4}$School of Cyber Security, University of Chinese Academy of Sciences\\
$^{5}$Electrical Engineering and Computer Science, University of California, Merced
}

\maketitle

\begin{abstract}
Video deblurring is a challenging problem as the blur is
complex and usually caused by the combination of camera shakes, object motions, and depth variations.
Optical flow can be used for kernel estimation since it predicts motion trajectories.
However, the estimates are often inaccurate in complex scenes at object boundaries, which are crucial in kernel estimation.
In this paper, we exploit semantic segmentation in each blurry frame to understand the scene contents and
use different motion models for image regions to guide optical flow estimation.
While existing pixel-wise blur models assume that the blur kernel is the same as optical flow during the exposure time, this assumption does not hold
when the motion blur trajectory at a pixel is different from the estimated linear optical flow.
We analyze the relationship between motion blur trajectory and optical flow,
and present a novel pixel-wise non-linear kernel model to account for motion blur.
The proposed blur model is based on the non-linear optical flow, which describes complex motion blur more effectively.
Extensive experiments on challenging blurry videos demonstrate
the proposed algorithm performs favorably against the state-of-the-art methods.
\end{abstract}

\section{Introduction}
\begin{figure}[t]\small
	\begin{center}
		\begin{tabular}{@{}cc@{}}
			\includegraphics[width = 0.5\linewidth, height = 0.31\linewidth]{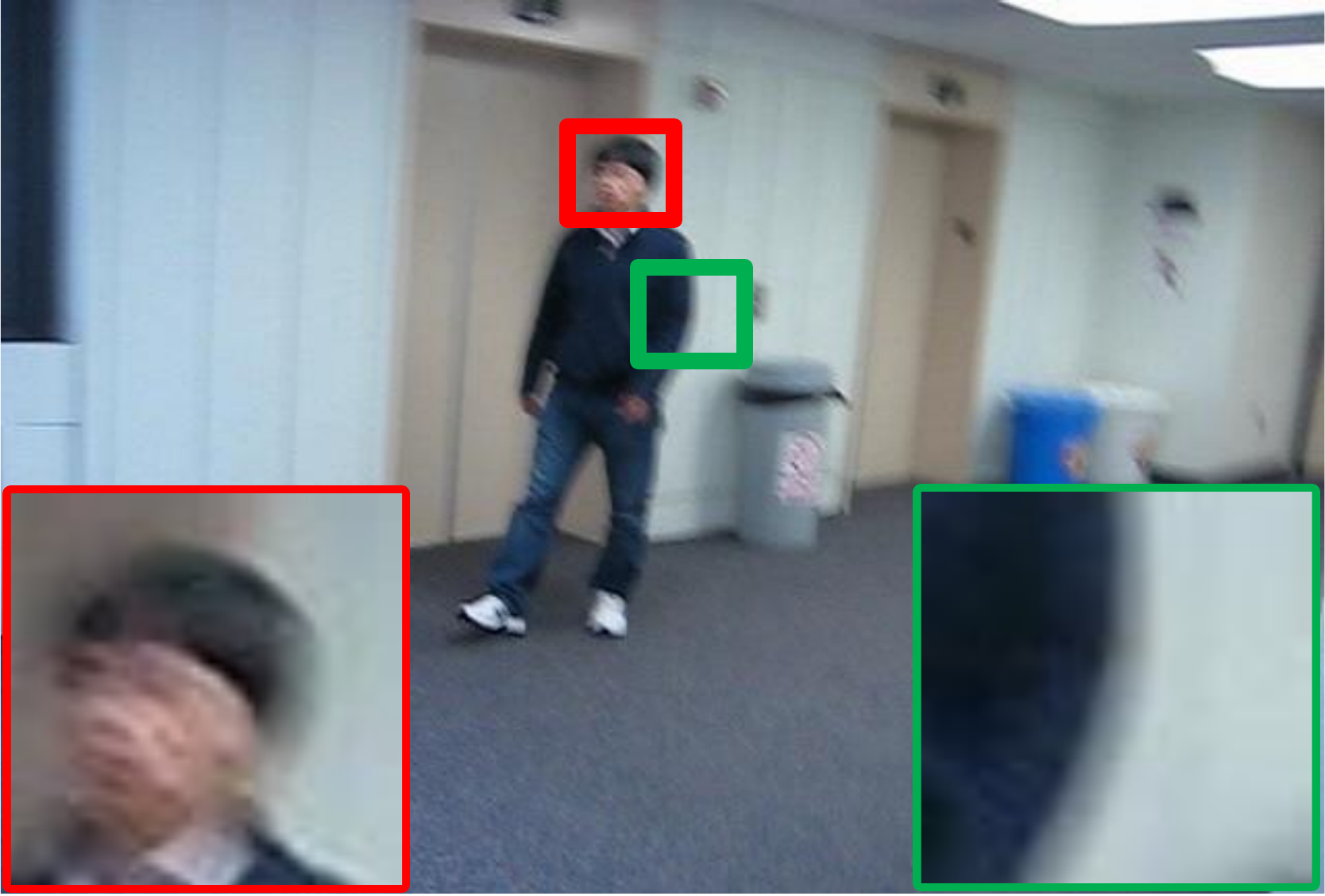} & \hspace{-0.3cm}
			\includegraphics[width = 0.5\linewidth, height = 0.31\linewidth]{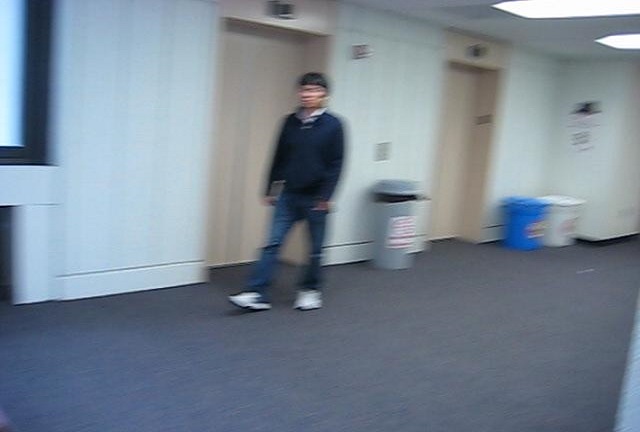} \\
			(a) Frame $t$ &
			(b) Frame $t+1$ \\	
			\includegraphics[width = 0.5\linewidth, height = 0.31\linewidth]{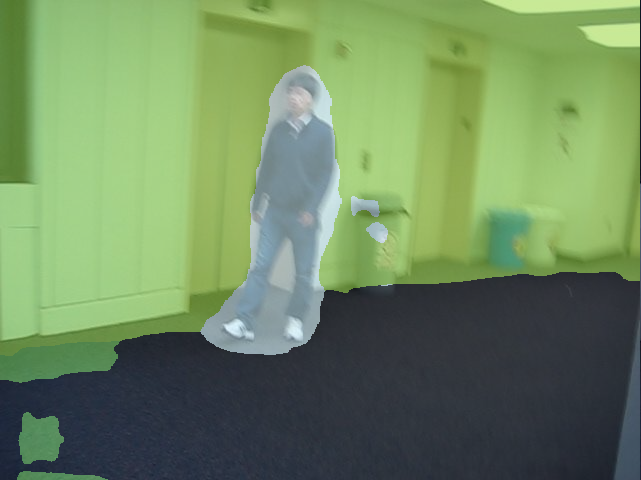} & \hspace{-0.3cm}
			\includegraphics[width = 0.5\linewidth, height = 0.31\linewidth]{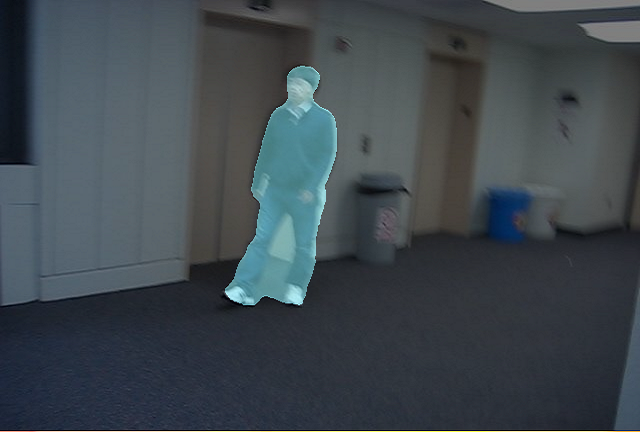} \\
			(c) Initial segmentation~\cite{ghiasi2016laplacian}  &
			(d) Our segmentation \\
			\includegraphics[width = 0.5\linewidth, height = 0.31\linewidth]{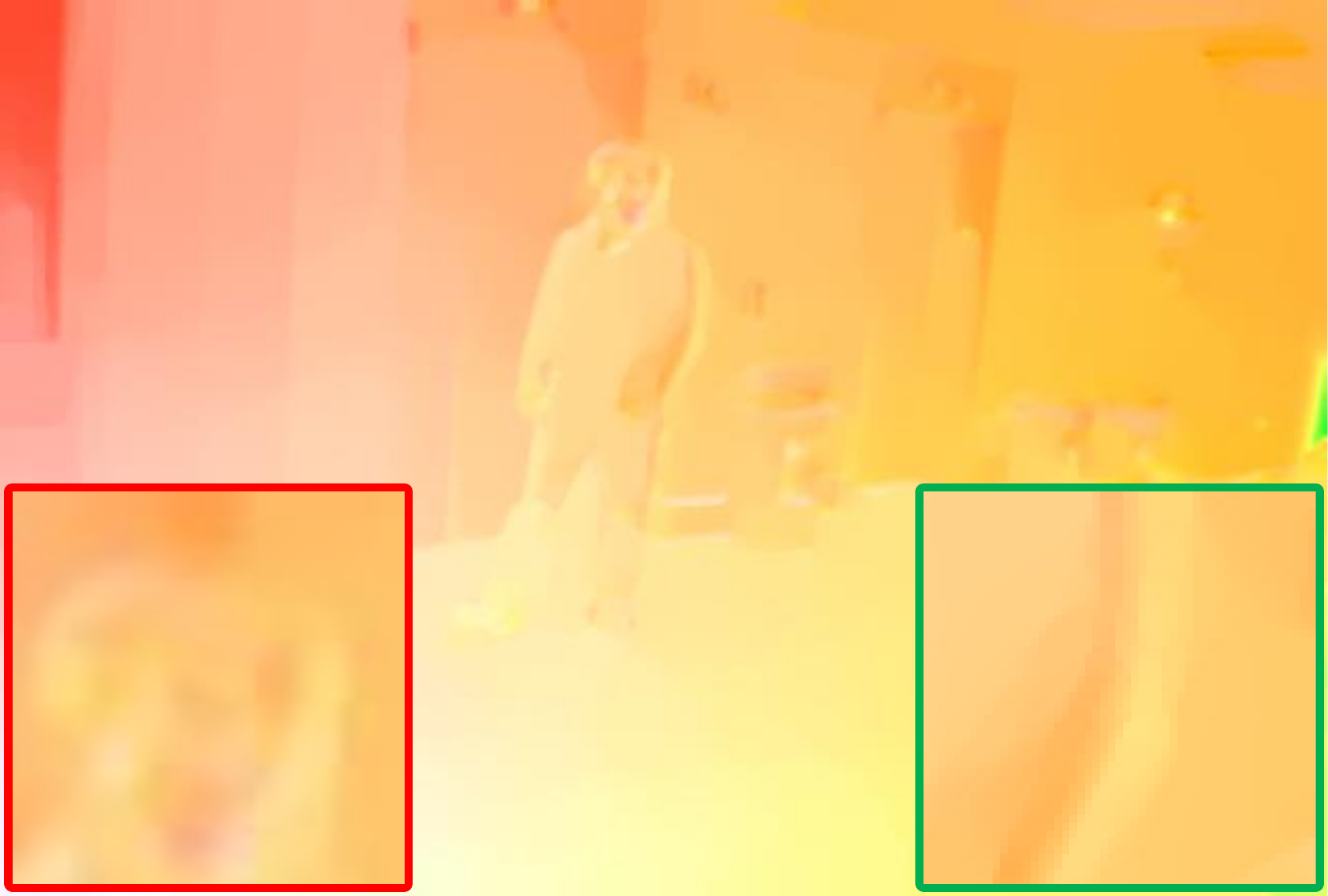} & \hspace{-0.3cm}
			\includegraphics[width = 0.5\linewidth, height = 0.31\linewidth]{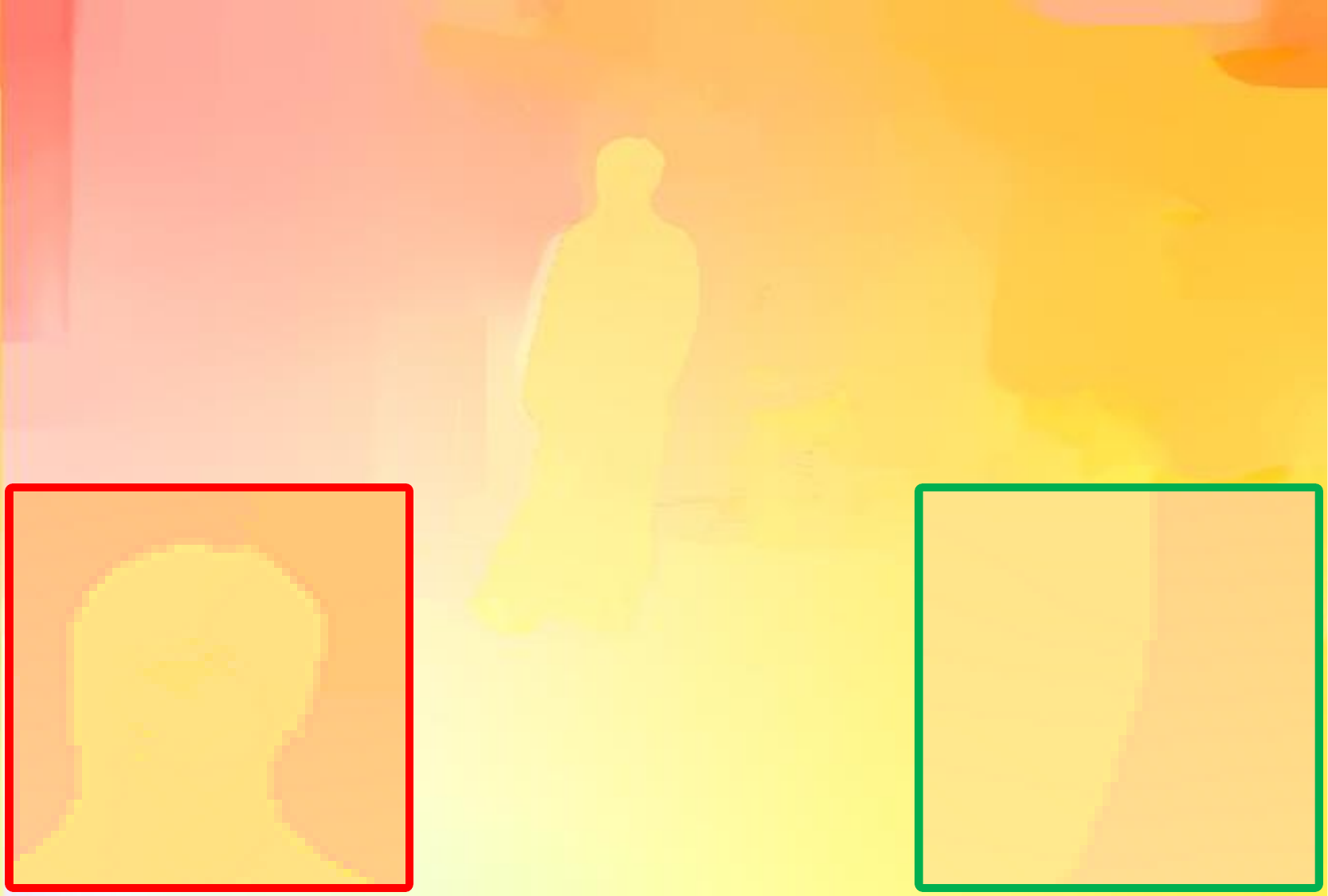} \\
			(e) Optical flow~\cite{kim2015generalized} &
			(f) Our optical flow \\
			\includegraphics[width = 0.5\linewidth, height = 0.31\linewidth]{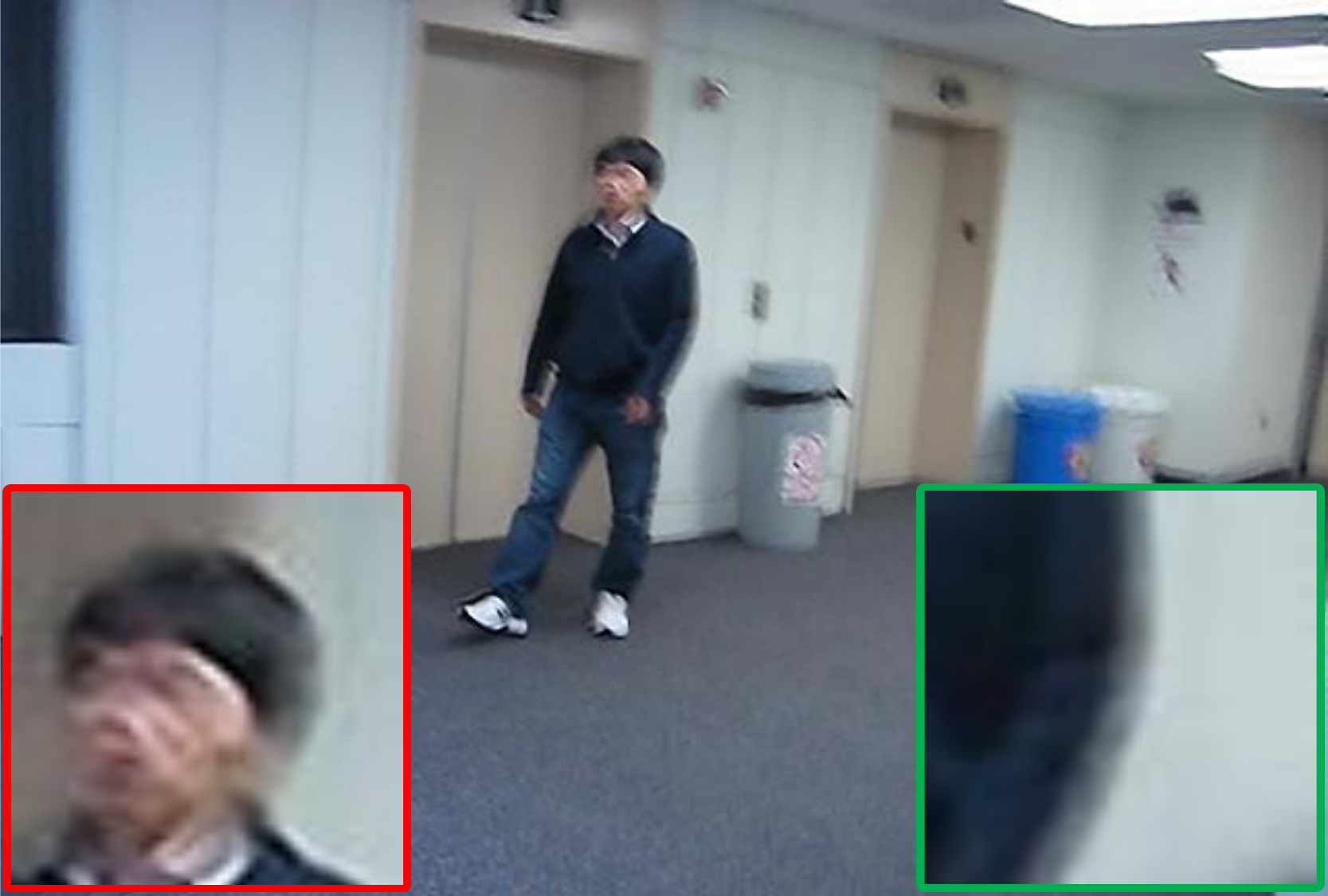} & \hspace{-0.3cm}
			\includegraphics[width = 0.5\linewidth, height = 0.31\linewidth]{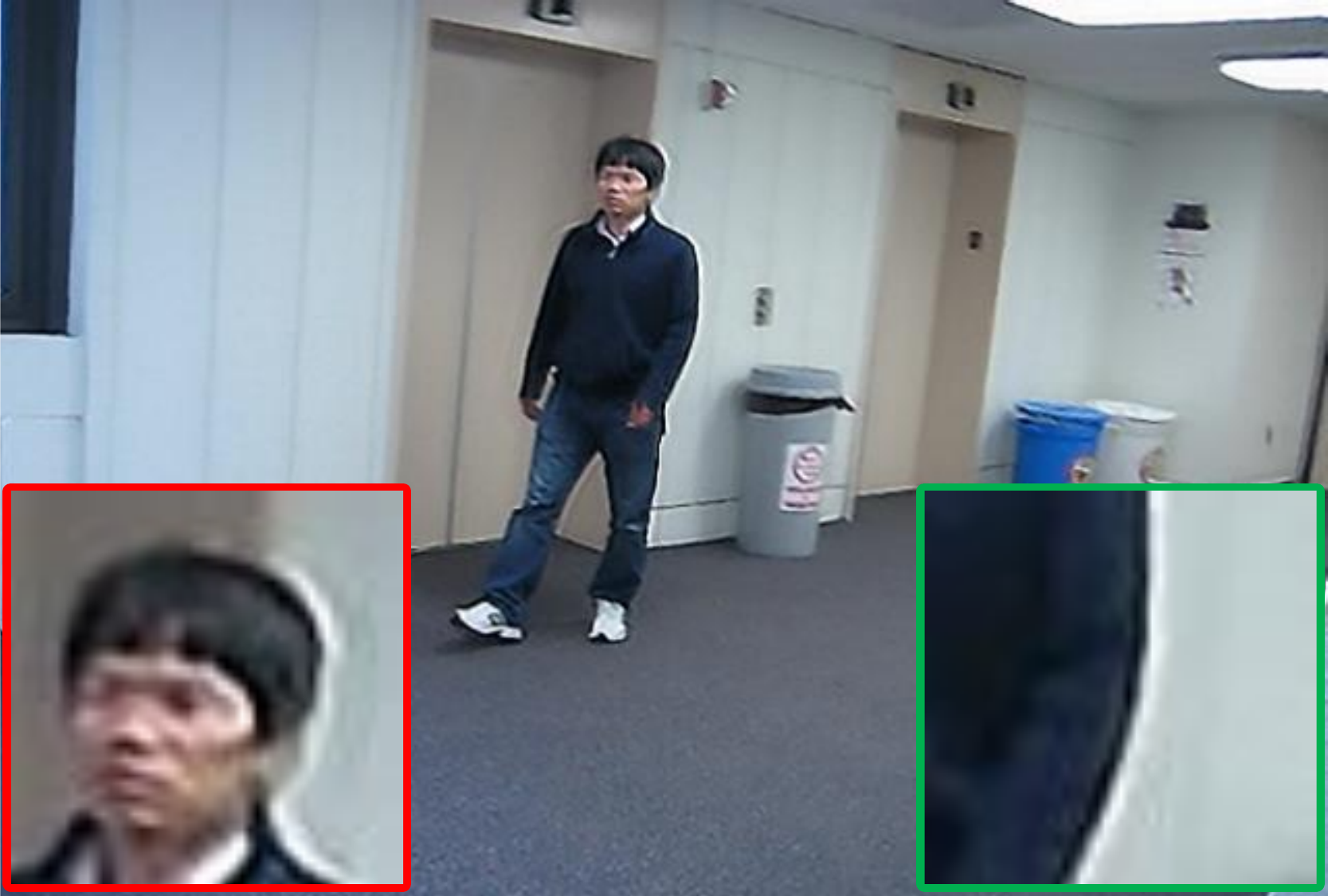} \\
			(g) Deblurred result~\cite{kim2015generalized} &
			(h) Our deblurred result \\
		\end{tabular}
	\end{center}
	%
	\caption{(a)-(b) Consecutive frames.
		(c) Semantic segmentation by~\cite{ghiasi2016laplacian}.
		(d) Segmentation of by the proposed algorithm, which is more accurate at the object boundary.
		(e) Optical flow by~\cite{kim2015generalized} from frame $t$ to $t+1$.
		(f) Optical flow by the proposed algorithm, which is more accurate around the object and background.
		(g) Deblurred result by~\cite{kim2015generalized}.
		(h) Deblurred image by the proposed algorithm. }
	\label{fig-example}
\end{figure}

The recent years have witnessed significant advances in image deblurring with
numerous applications \cite{ren2016image,yue2014hybrid}.
However, most deblurring methods are developed for single images
\cite{cao2015scene,michaeli2014blind,yan2017image} and
considerably less attention has been paid to videos  \cite{kim2015generalized,tai2008image,wulff2014modeling},
where the blur is caused by camera shakes, object motions, and depth variations, as illustrated by an example in
Figure~\ref{fig-example}.
Due to interacting and complex motions,
video deblurring cannot be modeled well by conventional uniform~\cite{fergus2006removing} or non-uniform blur \cite{whyte2012non}
models.
On the other hand, as most existing methods for video deblurring assume that the captured scenes are static \cite{lee2013dense,paramanand2013non},
these approaches do not handle blurs caused by abrupt motions and usually generate deblurred results with significant artifacts.

\begin{figure}[t]\small
	\centering
	\includegraphics[width = 0.45\textwidth]{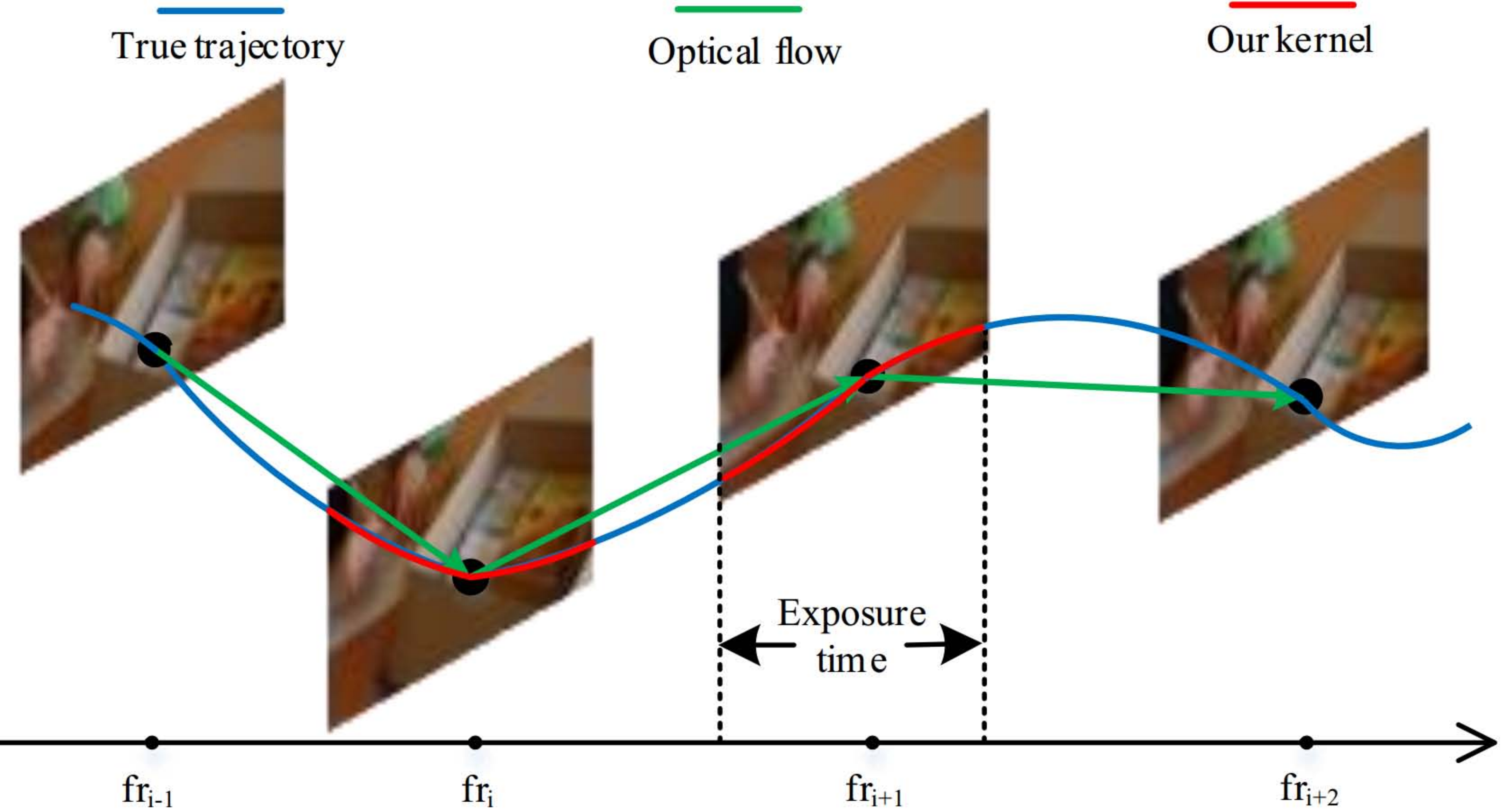}
	\caption{Video motion blur.
		The green line represents the true motion blur trajectory of the highlighted pixel. The blue line denotes the estimated optical flow. The ground truth motion blur trajectory is smooth and different from optical flow.
		Based on this observation, we approximate the true motion blur trajectory using the PWNLK model (red line) obtained from a quadratic function of optical flow.
	}
	\label{fig:ker-from-flow}
\end{figure}

To address these issues, deblurring algorithms based on segmentation \cite{bar2007variational,cho2007removing}
and motion transformation \cite{matsushita2005full, cho2012video} have been proposed.
However, segmentation based algorithms~\cite{bar2007variational,cho2007removing} require accurate object segments for kernel estimation.
In addition, transformation based methods~\cite{matsushita2005full, cho2012video} depend heavily on
whether sharp image patches can be extracted across frames for restoration.
Recently, Kim and Lee \cite{kim2015generalized} use
the bidirectional optical flow to estimate pixel-wise blur kernels, which is able to
handle generic blur in videos.
However, the deblurred results still contain some artifacts which can be attributed to two reasons.
First, the estimated optical flow may contain significant errors, particularly due to
large displacements or blurred edges \cite{portz2012optical,tsaivideo}.
Second, the pixel-wise linear blur kernel is assumed to be the same as the bidirectional optical flow.
This assumption does not usually hold for real images as illustrated in Figure~\ref{fig:ker-from-flow}.

In this work, we propose an efficient algorithm to estimate optical flow and semantic segmentation for video deblurring.
If the semantic segmentation of the scene is known,  optical flow within the same object should be smooth but flow across the boundary needs not be smooth,
and such constraints facilitate accurate blur kernel estimation.
On the other hand, accurate optical flow and segmentations are crucial to restore sharp frames.
Hence, accurate semantic segmentations and optical flow facilitate to recover accurate sharp frames and vice versa.
In addition, as blur kernel is caused by a complicated combination of camera shakes and objects motions,
it is different from the estimated linear optical flow as shown in Figure~\ref{fig:ker-from-flow}.
Although some non-linear optical flow methods~\cite{yuan2007simultaneous} have been developed,
these approaches focus on restoring complex flow structure, \eg, vortex and vanishing divergence,
and the estimated optical flow is still a straight line for each pixel.
To deal with various blurs in real scenes, we  propose a motion blur model
using a quadratic function to model optical flow and approximate the pixel-wise blur kernel
based on the non-linearity assumption.
Extensive experiments on challenging blurry videos demonstrate
the proposed algorithm performs favorably against the state-of-the-art methods.

The contributions of this work are summarized as follows.
First, we propose a novel algorithm to solve semantic segmentation, optical flow estimation, and video deblurring simultaneously in a unified framework.
Second, we exploit semantic segmentation to account for occlusions and blurry edges for accurate optical flow estimation.
Third, we propose a pixel-wise non-linear kernel (PWNLK)  model to approximate motion trajectories in videos,
where the blur kernel is estimated from optical flow under the non-linearity assumption.
We show that motion blur cannot be simply modeled by optical flow,
and the non-linearity assumption of optical flow is important for video deblurring.

\section{Related Work}

{\flushleft \textbf{Deblurring based on motion transformation.}}
Video deblurring based on  motion transformation
detects sharp images or patches by computing the absolute displacements of pixels between adjacent frames, from which the
clear contents are restored~\cite{lee2013video}.
Matsushita~\etal \cite{matsushita2005full} transfer and interpolate sharper image pixels of neighboring frames for deblurring.
Clear regions in a blurry video are detected to restore blurry regions of the same content in nearby frames~\cite{cho2012video}.
A multi-image enhancement method based on a unified Bayesian framework is proposed by Sunkavalli~\etal \cite{sunkavalli2012video}
to establish correspondence among neighboring frames.
However, these transformation based methods do not involve deconvolution and rely
on sharp patches from nearby frames which may not exist.

{\flushleft \textbf{Deblurring based on deconvolution.}}
Deconvolution based methods~\cite{delbracio2015hand} can be categorized into three approaches
based on uniform kernel,
layered blur model, and pixel-wise kernel.
Uniform kernel based methods \cite{cai2009blind,vsroubek2012robust} assume that the blur in each frame is spatial invariant. These methods are less effective for complex scenes with spatially variant blurs.

To deal with complex motion blurs, layered blur model is developed in the deblurring problem to handle locally varying blurs \cite{cho2007removing,wulff2014modeling}.
Cho~\etal~\cite{cho2007removing} simultaneously estimate multiple object motions,
blur kernels, and the associated image segmentations to solve video deblurring problem.
Kim~\etal~\cite{kim2013dynamic} adopt a nonlocal regularization on the
estimated residual and blurred image to handle object segmentation for dynamic scene deblurring.
A layered motion model is proposed by Bar~\etal \cite{bar2007variational} to segment images into foreground as well as background layers, and estimate a linear blur kernel for the foreground layer.
Wulff and Black \cite{wulff2014modeling} extend this layered model to segment images into foreground and background regions from where the global motion blur kernels are estimated based on affine motion.
However, these methods depend heavily on whether accurate segments can be obtained or not since each region is deblurred based on the segmentation.

To address this issue,
Li~\etal~\cite{li2010generating} parameterize the observed frames
in a blurry video by homography and recover sharp contents by jointly estimating blur kernels, camera duty cycles, and latent images.
In~\cite{zhang2015intra}, a projective motion path model~\cite{tai2011richardson} is used to estimate blur kernels
by exploiting inter-frame misalignments between frames.
However, blur models based on homography and projection are designed to account for global camera motions, which cannot model complex object motion and depth variations.
To solve this problem, Kim and Lee \cite{kim2014segmentation} propose a segmentation-free algorithm by using bidirectional optical flow to model motion blurs for dynamic scene deblurring.
This method is extended to generalized video deblurring in \cite{kim2015generalized} by alternatively estimating optical flow and latent frames.
Although promising results have been obtained, the assumption that motion blur
is same as optical flow does not hold in complex scenes as illustrated in Figure \ref{fig:ker-from-flow} especially when the camera duty cycle is large.

Different from these methods, we take scene semantics and objects into account and use the segmentation to improve optical flow estimation rather than direct deblurring.
We then use the estimated optical flow to compute pixel-wise kernel based on non-linear assumption.

{\flushleft \textbf{Deblurring based on deep learning.}}
Recently, image or video restoration algorithms that aim to recover the
underlying sharp contents based on convolutional neural networks, have emerged.
In \cite{schuler2016learning},
deep neural networks are used for single image deblurring using synthetic training data.
Su~\etal~\cite{su2016deep} propose a deep encoder-decoder network to address real world video deblurring problems.
Nevertheless, when images are heavily blurred, this method may
introduce temporal artifacts that become more visible after stabilization.

{\flushleft \textbf{Semantic segmentation.}}
Semantic segmentation~\cite{liang2015deep,liu2015matching,liu2016surveillance} aims to cluster image pixels of the same object class with assigned labels. Numerous recent methods use semantic segmentation to resolve ambiguities in road signs detection~\cite{maldonado2007road}, 3D reconstruction~\cite{hane2013joint}, and optical flow estimation by using different motion models at different object regions~\cite{sevilla2016optical}.

\section{Proposed Algorithm}

The use of semantic information facilitates modeling optical flow for each region and results in better estimates of pixel movements,
especially at motion boundaries.
In addition, the proposed PWNLK model is designed to estimate blur kernels more accurately.
In this section, we  analyze the relationship between optical flow and motion blur trajectory,
and present a video deblurring algorithm based on semantic segmentation and non-linear kernels.

\subsection{Motion Blur Model from Optical Flow}
The main challenge of video deblurring is how to estimate pixel-wise blur kernels from images.
As shown in Figure~\ref{fig:ker-from-flow}, optical flow (green line) reflects the moving linear direction of a pixel
between adjacent frames which may be different from the motion trajectory (blue line).
Thus, it is less accurate to model motion blur using optical flow based on linear assumption.
A motion blur trajectory is usually smooth and its shape can be approximated by a quadratic function.
To model motion blur trajectories $t$, we use the following parametric PWNLK model:
\begin{equation}
{\small
	t(f) = a f^2 + b f + c,
	\label{eq-motion-blur-model}
}	
\end{equation}
where $f=(u,v)$ is the estimated optical flow of adjacent frames, and $a$, $b$, as well as $c$ are parameters to be determined.
We find that the motion blur trajectory can be approximated well with this model as shown in Figure~\ref{fig:ker-from-flow}.
We parameterize each kernel $k_{i}(x)$ at pixel $x$ of frame $i$ as a quadratic function of bidirectional optical flow~\cite{kim2014segmentation, kim2015generalized},
\begin{equation}
{\small
	\begin{split}
	& k_{i}(x) = \\
	&\left\{
	\begin{aligned}
	&\frac{\delta(uv_{i, i +1} {\rm{-}} vu_{i, i +1})}{2\tau_i ( a_{i, i\rm{+}1}\|f_{i, i +1}\|^2  {\rm{+}} b_{i, i + 1} \|f_{i, i +1}\|  {\rm{+}} c_{i, i\rm{+}1})} \  \text{if} \ f {\rm{\in}} [0,\tau_i f_{i, i\rm{+}1}],\\
	&\frac{\delta(uv_{i, i -1} {\rm{-}} vu_{i, i -1})}{2\tau_i ( a_{i, i-1}\|f_{i, i-1}\|^2 {\rm{+}} b_{i, i-1} \|f_{i, i-1}\| {\rm{+}} c_{i, i-1})} \  \text{if} \ f {\rm{\in}} (0,\tau_i f_{i, i-1}],\\
	&0, \qquad \qquad \qquad \qquad \qquad \qquad \qquad \qquad \qquad \quad \text{otherwise.}
	\end{aligned}
	\right.
	\end{split}
	\label{eq-pwnl}
}
\end{equation}
With the blur kernel $k_i$, the blurry frame $y_i$ can be formulated as
\begin{equation}
y_i = k_il_i + \varepsilon,
\label{equ-motion-blur-model-1}
\end{equation}
where $l_i$ denotes the $i$-th latent frame, and $\varepsilon$ denotes noise.
Based on the blur model~\eqref{equ-motion-blur-model-1}, we present an effective video deblurring method
and present detailed analysis of the algorithm in the following sections.

\subsection{Proposed Video Deblurring Model}
Based on the PWNLK model~\eqref{eq-motion-blur-model}, blur formulation~\eqref{equ-motion-blur-model-1} and the standard maximum a posterior framework \cite{krishnan2011blind}, our video deblurring model is defined as
\begin{eqnarray}
{\small
	\begin{split}
	E(l,k,f,s) & = \sum_i \{E_{d}(l_i, k_i, y_i) + E_{m}(f_{ik}, s_{ik}) \\
	& + E_{t}(l_i, f_i, s_i) + E_{s}(l_i, f_i,s_i) \} ,
	\end{split}
	\label{equ-map2}
}
\end{eqnarray}
where $f_{ik}=(u_{ik},v_{ik})$ and $s_{ik}$ denote optical flow and segmentation in the $k$-th layer of $i$-th frame, respectively.
The first term $E_{d}$ in \eqref{equ-map2} is the data fidelity term, \ie, the deblurred frame $l_i$ should be consistent with the observation $y_i$.
The second term $E_{m}$ denotes a motion term which encodes two assumptions.
First, neighboring pixels should have similar motion if they belong to the same semantic segmentation layer.
Second, pixels from each layer $k$ should share a global motion model $f(\theta_{ik})$, where $\theta_{ik}$ is parameter that changes over time and depends on the object class $k$.
The third term $E_{t}$ is the temporal regularization term, which is used
to ensure the brightness constancy between adjacent frames.
The last term $E_{s}$ denotes the spatial regularization term of latent images and optical flow.
The details of each term in \eqref{equ-map2} are described below.

{\flushleft \textbf{Data term based on the PWNLK model.}}
It has been shown that using gradients of latent and blurry images in the data term can reduce ringing artifacts \cite{kim2014segmentation,kim2015generalized}.
Thus, our data fidelity term is defined as
\begin{equation}
{\small
	E_{d}(l_i, k_i, y_i)=
	\sum_i \lambda \|\nabla (k_i l_i) - \nabla y_i\|_2^2,
	\label{eq-data}
}
\end{equation}
As blur kernel $k_i$ is computed according to the motion blur trajectory in~\eqref{eq-motion-blur-model},
the data fidelity term~\eqref{eq-data} involves parameters $a$, $b$, and $c$.
To obtain a stable solution, we need to regularize these motion blur parameters~\cite{tang2017tri}.
The Tikhonov regularization has been extensively used in the literature of image deblurring.
However, we note that motion blur has similar properties to the optical flow in most examples.
For example, the estimated motion blur would have the same property
if the estimated optical flow has piece-wise property.
That is, if $\nabla f_i = 0$ at some regions, we would have $\nabla (a_if_i^2 + b_if_i+c_i) = 0$.
Based on this assumption, we have $b_i = -2a_i f_i$.
As $\nabla f_i = 0$, $f_i$ should be a constant $C$.
This property motivates us to use the following regularization on parameters $a$ and $b$,
\begin{equation}
{\small
	\sum_i \{\beta||a_i||_2^2 + \gamma||b_i-C||_2^2\},
	\label{equ-reg-ab}
}
\end{equation}
where $\beta$ and $\gamma$ denote the weights of each term in the regularization terms.

{\flushleft \textbf{Motion term.}}
The motion term should satisfy:
1) pixels in the same segmentation layer $s_{ik}$ should share a global motion model $f(\theta_{ik})$,
2) neighboring pixels in the same segmentation layer $s_{ik}$ should have similar optical flow.
Thus, our motion term is defined as
\begin{equation}
{\small
	\begin{split}
	& E_{m}(f_{ik}, s_{ik}) = \sum_i \{\sum_x \rho_{\text{aff}}(f_{ik}(x)-f(\theta_{ik})) \\
	& + \sum_x \sum_{r\in \mathcal{N}_x}||f_{i}(x)-f_{i}(r)||^2_2\delta(s_{ik}(x)=s_{ik}(r))\},\\
	\end{split}
	\label{eq-motion}
}
\end{equation}
where $\mathcal{N}_x$ denotes the four nearest neighbors of the pixel $x$,
and $\rho_{\text{aff}}$ is a robust penalty function which enforces that
the pixels in the same segmentation have the same affine motion model~\cite{sun2013fully}.
In addition, $\delta(\cdot)$ denotes the indicator function that is equal to 1
if its expression is true, and 0 otherwise.

{\flushleft \textbf{Spatial term.}}
The spatial regularization term aims to alleviate
the ill-posed inverse problem.
We assume that the spaial term should
1) constrain the pixels with similar colors to lie within the same segmentation layer $s_{ik}$,
and 2) enforce spatial coherence in both latent frames and optical flow.
With these assumptions, the spatial term is defined by
\begin{eqnarray}
{\small
	\begin{split}
	E_{s}(l_i, f_i,s_i) & = \sum_i \{ |\nabla l_i| + \sum_{n=-N}^{N}g_i(x)|\nabla f_{i,i+n}|\\
	& +\sum_{x}\sum_{r\ne x}\omega_{x,r}\delta(s_{ik}(x)\ne s_{ik}(r))\},
	\end{split}
	\label{eq-corrof}
}
\end{eqnarray}
where the weight $g_i(x)$ denotes edge-map~\cite{kim2015generalized} to preserve discontinuities in the optical flow at edges.
In addition, $\omega_{x,r}$ is a weight which measures the similarity between $x$ and $r$.
Similar to the optical flow estimation method~\cite{sun2013fully}, we define it as
\begin{equation}
{\small
	\begin{split}
	\omega_{x,r} = \exp \{ -\frac{||x-r||^2 + ||l_{i}(x)-l_{i}(r)||^2}{\sigma^2} \},
	\end{split}
}
\label{eq-weigh}
\end{equation}
where $\sigma$ is a constant.
For a given pixel $x$, if we know other neighboring pixels $r$ have similar color as $x$, we set them with the same segment.
The effectiveness of the regularization term is
demonstrated in Section~\ref{sec-anal}.

{\flushleft \textbf{Temporal term.}}
Human vision system is sensitive to temporal inconsistencies presented in videos.
To improve temporal coherence,
we first utilize the optical flow to find the corresponding pixels between
neighboring frames in a local temporal window $[i-N, i+N]$ and ensure that the corresponding pixels vary smoothly.
We then enforce that corresponding pixels between neighboring frames
should belong to the same segment.
Thus, the temporal coherence is defined by
\begin{eqnarray}
{\small
	\begin{split}
	E_{t}(l_i, f_i, s_i) & = \sum_i \{ \sum_{n=-N}^{N} \mu_n |l_{i}(x)-l_{i+n}(x')| \\
	& + \sum_{n=-N}^{N} \mu_n |s_{i}(x)-s_{i+n}(x')|\},
	\end{split}
}
\label{eq-temporal}
\end{eqnarray}
where $n$ denotes the index of neighboring images at frame $i$
and $\mu_n$ is a weight for the regularization term.
In addition, $x'= x + f_{i,i+n}$ is the corresponding pixel at the next $n$-th frame for $x$
according to the motion $f_{i,i+n}$.
We use the $L_1$-norm regularization in \eqref{eq-temporal}
for robust estimates
against outliers and occlusions \cite{kim2015generalized}.

\subsection{Inference}
Based on the above analysis, we obtain the proposed video deblurring model.
Although the objective function is non-convex with multiple variables, we can use an alternating minimization method~\cite{kim2015generalized} to solve it.

{\flushleft \textbf{Latent frames estimation.}}
%
With the optical flow $f$, segmentation $s$, and the parameters $a$, $b$ and $c$, the optimization problem with respect to $l_i$ is
\begin{equation}
{\small
	\begin{split}
	& \min_{l_i} \lambda \sum_i \{\|\nabla (k_i l_i) - \nabla y_i\|_2^2 + |\nabla l_i| \\
	& + \sum_{n=-N}^{N} \mu_n |l_{i}(x)-l_{i+n}(x')|\}.
	\end{split}
	\label{eq-sub-l}
}
\end{equation}
Similar to \cite{kim2015generalized}, we optimize the latent frames subproblem~\eqref{eq-sub-l} using the primal-dual update method~\cite{chambolle2011first}.

{\flushleft \textbf{Semantic segmentation.}}
The semantic segmentation estimation can be achieved by solving
\begin{eqnarray}
\begin{small}
\begin{split}
& \min_{s_{i}} \sum_i \{\sum_{x}\sum_{r\ne x}\omega_{x,r}\delta(s_{ik}(x)\ne s_{ik}(r)) \\
& + \sum_x \sum_{r\in \mathcal{N}_x}||f_{i}(x)-f_{i}(r)||^2_2\delta(s_{ik}(x)=s_{ik}(r))\\
& + \sum_{n=-N}^{N} \mu_n |s_{i}(x)-s_{i+n}(x')|\}.
\end{split}
\label{eq-sub-s}
\end{small}
\end{eqnarray}
We optimize this subproblem \eqref{eq-sub-s} using the method in~\cite{sevilla2016optical}.
The semantically segmented regions provide information on a potential optical flow for the motion blurred object, which is used
to guide optical flow estimation instead of directly deblurring on each segment~\cite{bar2007variational,wulff2014modeling}.

Note that we only refine the segmentation results $s_{ik}$ according to possible moving objects including person, rider, car, etc, as like in Figure~\ref{fig-example}(d).
For other background objects (e.g., road, sky, wall), we do not refine their segmentation since these objects are always smooth and their segmentation results cannot affect our deblurring results.

{\flushleft \textbf{Optical flow estimation.}}
After obtaining $l$ and $s$, the optimization problem with respect to $f$ becomes
\begin{eqnarray}
{\small
	\begin{split}
	& \min_{f_i} \lambda \sum_i \{\|\nabla (k_i l_i) - \nabla y_i\|_2^2
	+ \sum_{n=-N}^{N}g_i(x)|\nabla f_{i,i+n}|\\
	& + \sum_x \sum_{r\in \mathcal{N}_x}||f_{i}(x)-f_{i}(r)||^2_2\delta(s_{ik}(x)=s_{ik}(r))\\
	& + \sum_x \rho_{\text{aff}}(f_{ik}(x)-f(\theta_{ik})) + \sum_{n=-N}^{N} \mu_n |l_{i}(x)-l_{i+n}(x')|\}.
	\end{split}
	\label{eq-sub-f}
}
\end{eqnarray}
%
%
%

We solve~\eqref{eq-sub-f} using the method in~\cite{kim2015generalized} and \cite{sun2013fully}.
%
After obtaining $f_i$,
we utilize it to estimate the blur kernel based on the non-linearity assumption,
instead of directly using the bidirectional optical flow as blur kernel.

{\flushleft \textbf{Motion blur trajectory parameters estimation.}}
For each blurry frame $y_i$, we
obtain its corresponding sharp reference $l_i$ and its bidirectional optical flow $f_i$.
With each image pair and the corresponding optical flow, the parameters of the motion blur kernel $a_i$, $b_i$ and $c_i$ are solved by
\begin{eqnarray}
\begin{small}
\min_{a,b} \lambda \sum_i \{\|\nabla (k_i l_i) \rm{-} \nabla y_i \|_2^2
\rm{+} \beta||a_i||_2^2 \rm{+} \gamma \| b_i \rm{-} C \|_2^2\}.
\label{eq-abc}
\end{small}
\end{eqnarray}
This is a least squares minimization problem and
we have the closed-form solutions for the parameters $a$, $b$ and $c$, respectively.
%

\begin{algorithm}[!t]
	\caption{Proposed video deblurring algorithm}
	\label{alg:vd-algorithm}
	\begin{algorithmic}
		\STATE {\textbf{Input:} Blurry frames $y$, duty cycle $\tau$, initialized optical flow $f$ by~\cite{wedel2009improved} and semantic segmentation $s$ by~\cite{ghiasi2016laplacian}.}
		\STATE Repeat the following steps from coarse to fine image pyramid level:
		\STATE {1. Solve for parameters $a$, $b$ and $c$ by minimizing \eqref{eq-abc}.}
		\STATE {2. Solve for optical flow $f$ by minimizing~\eqref{eq-sub-f}.}
		\STATE 3. Estimate blur kernel based on PWNLK model~\eqref{eq-pwnl}. 
		\STATE 4. Solve for latent image $l$ by minimizing~\eqref{eq-sub-l}.
		\STATE 5. Solve for segmentation $s$ by minimizing~\eqref{eq-sub-s}.
		\STATE \textbf{Output:} latent frames $l$, blur kernels $k$, optical flow $f$ and segmentation $s$.
	\end{algorithmic}
\end{algorithm}

Similar to the existing methods, we use the coarse-to-fine method with an image pyramid~\cite{kim2015generalized} to achieve better performance.
Algorithm~\ref{alg:vd-algorithm} summarizes the main steps of the proposed video deblurring on one image pyramid level.

\begin{figure}[t]\scriptsize
	\begin{center}
		\begin{tabular}{@{}cccccc@{}}
			\includegraphics[width =0.15\linewidth]{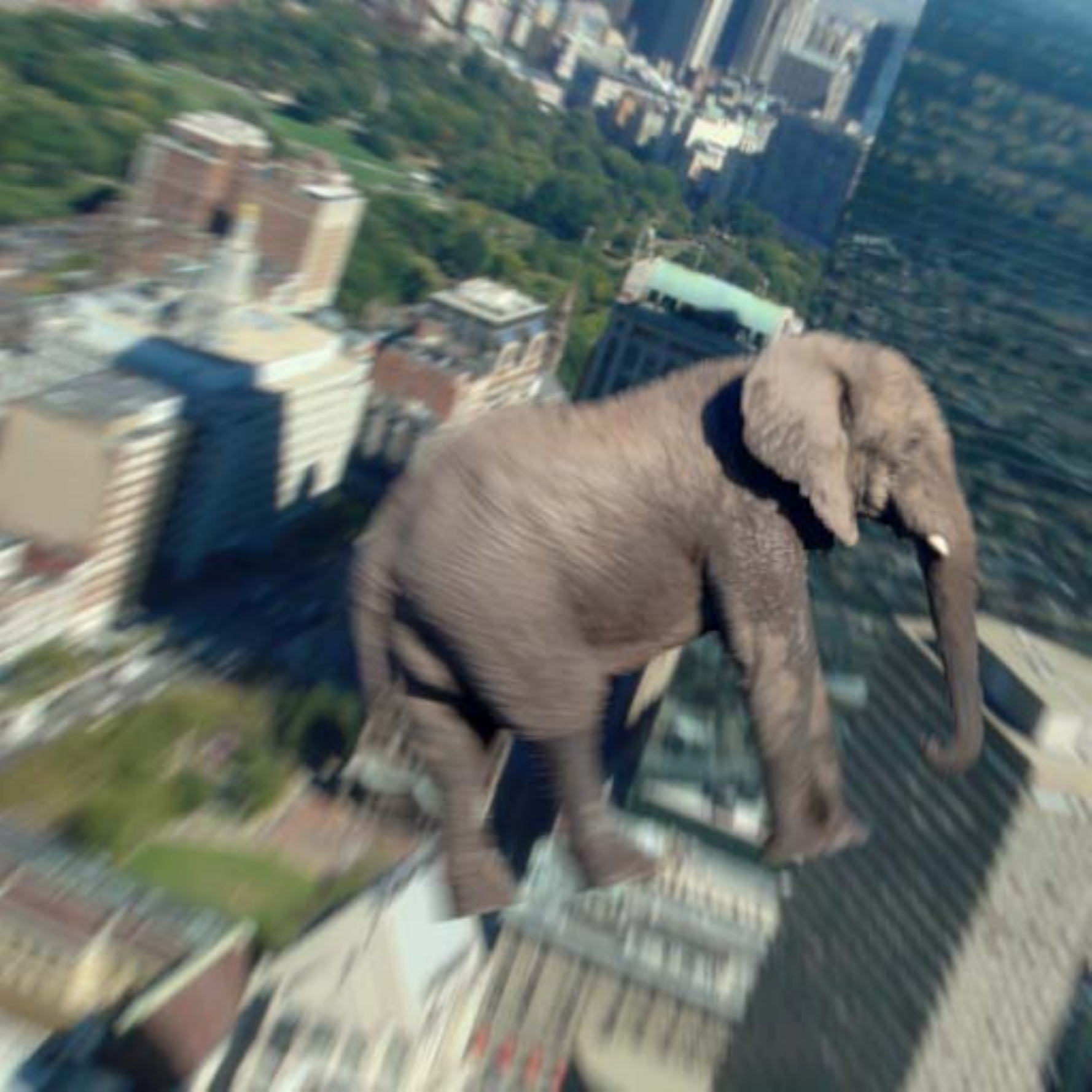} & \hspace{-0.4cm}
			\includegraphics[width = 0.15\linewidth]{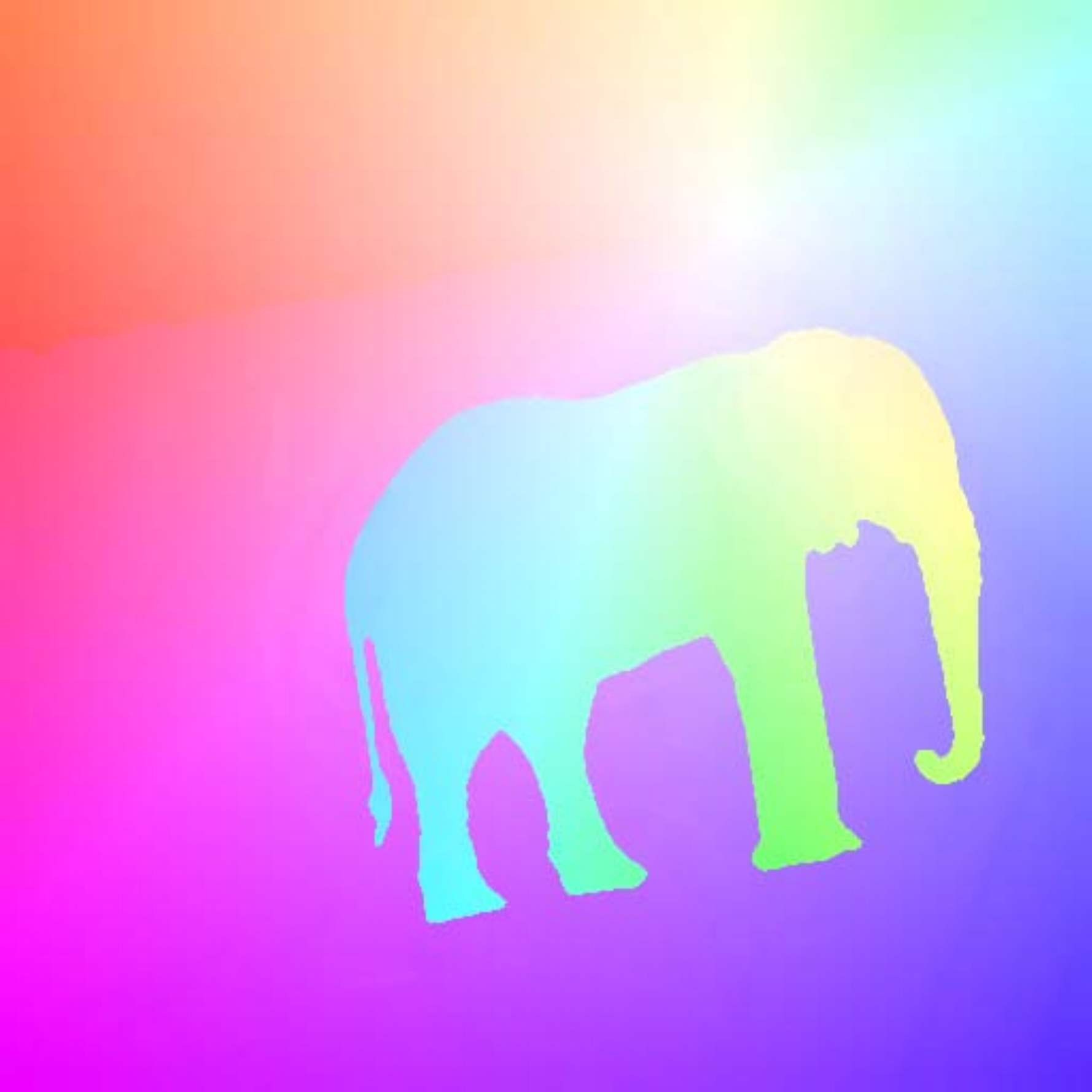} & \hspace{-0.4cm}
			\includegraphics[width = 0.15\linewidth]{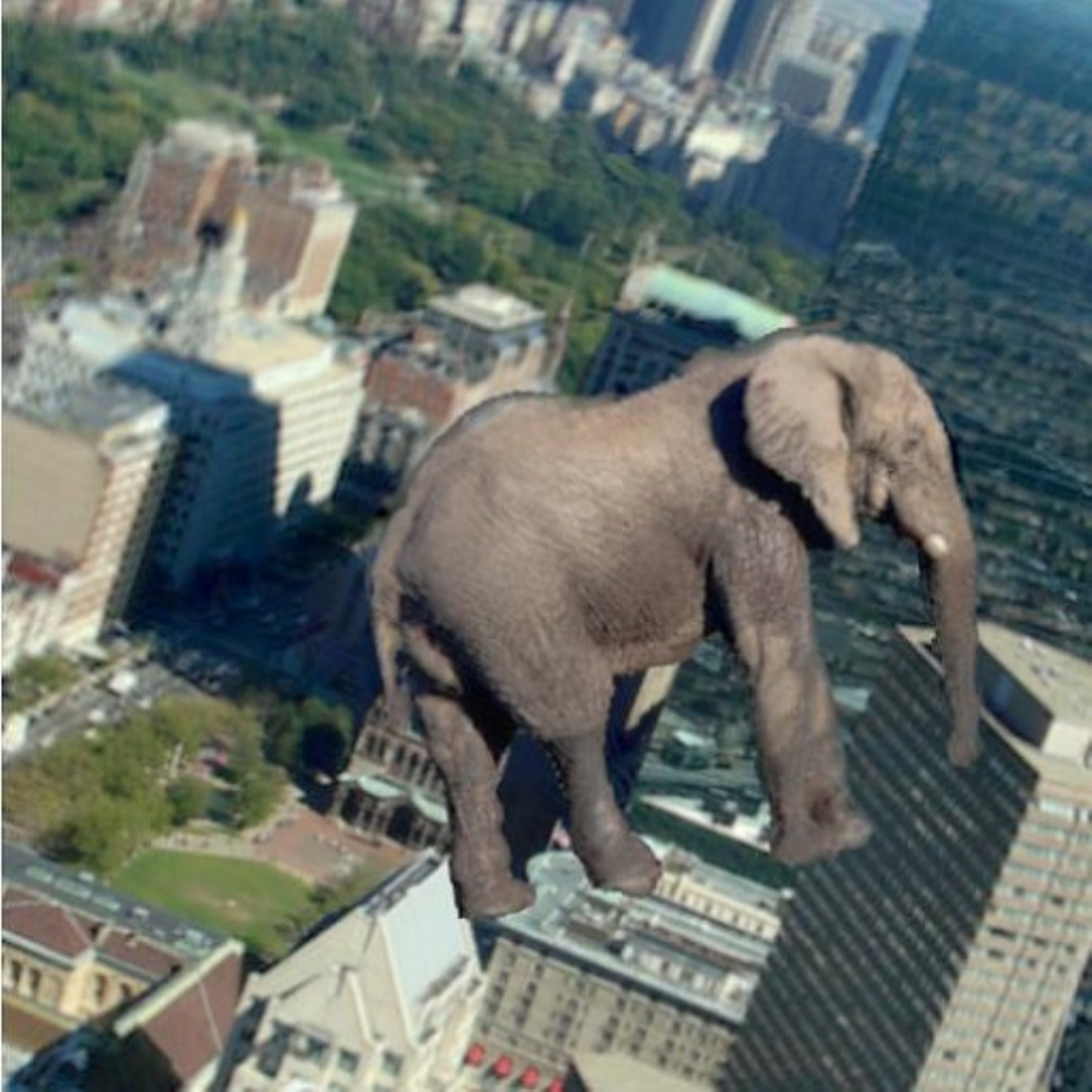} & \hspace{-0.4cm}
			\includegraphics[width = 0.15\linewidth]{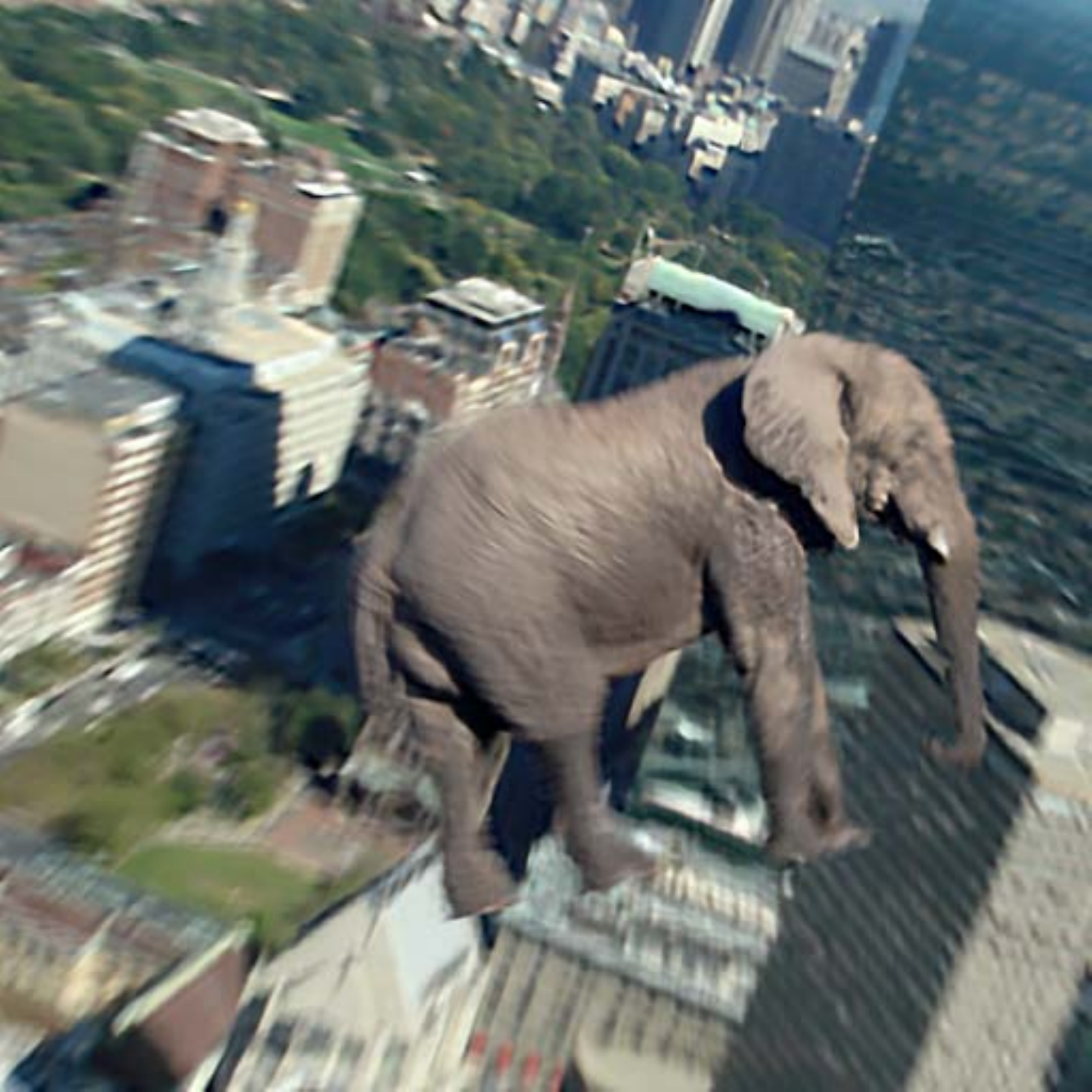} & \hspace{-0.4cm}
			\includegraphics[width = 0.15\linewidth]{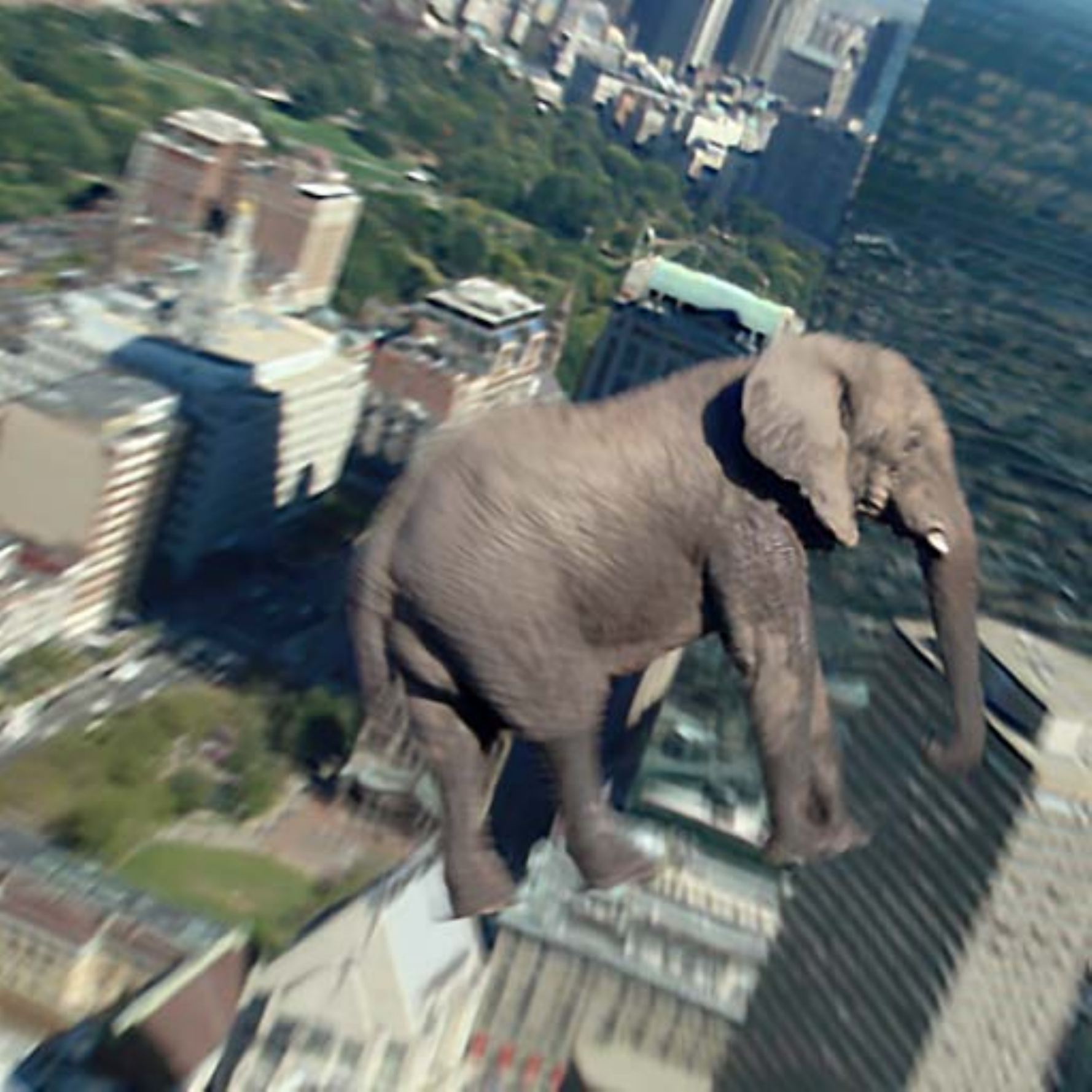} & \hspace{-0.4cm}
			\includegraphics[width = 0.15\linewidth]{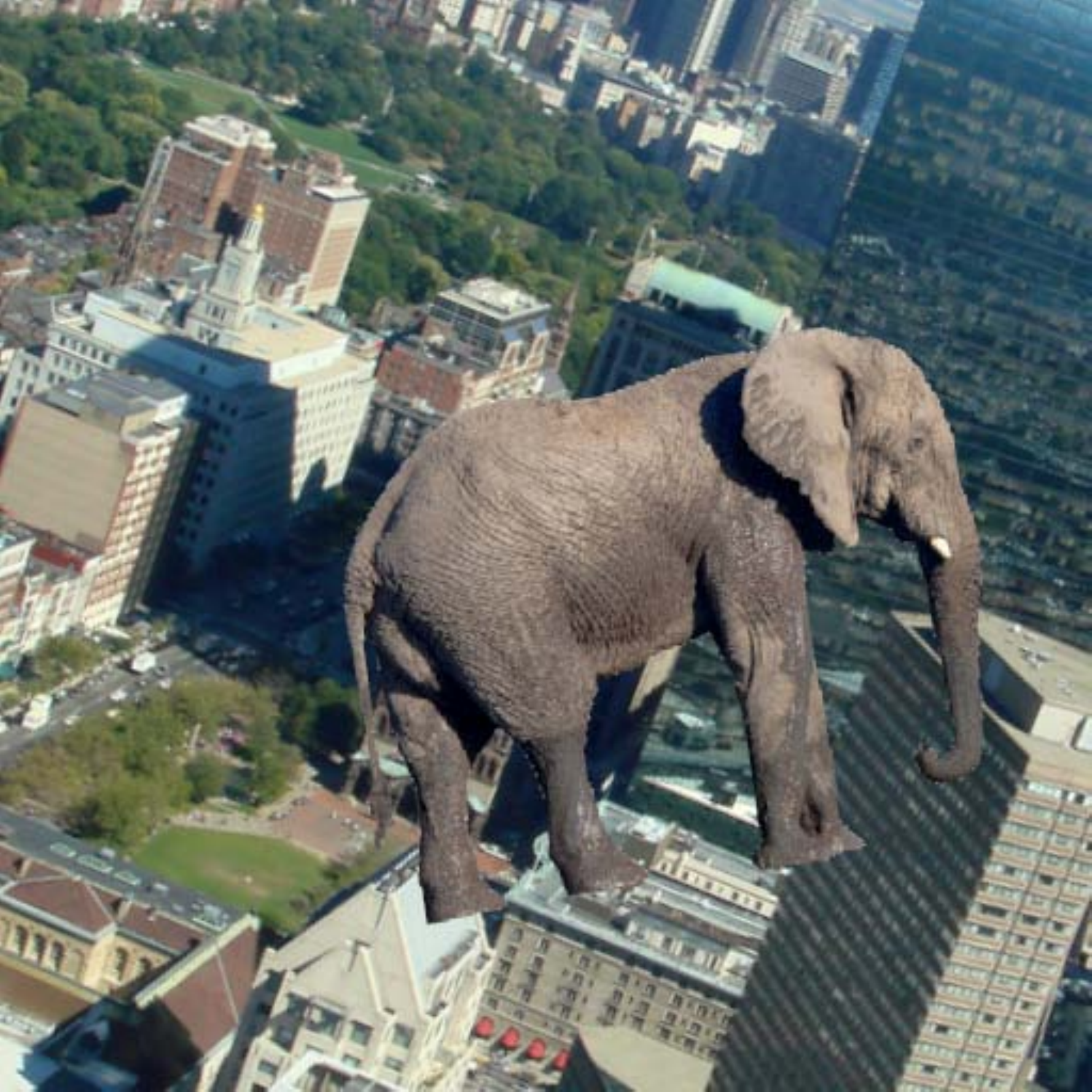} \\
			(a) Input & \hspace{-0.4cm} (b) Flow & \hspace{-0.4cm} (c) 16.05dB & \hspace{-0.4cm} (d) 17.98dB & \hspace{-0.4cm} (e) 18.13dB & \hspace{-0.4cm} (f) Truth \\
		\end{tabular}
	\end{center}
    \vspace{-2mm}
	\caption{The limitation of linear assumption in~\cite{kim2015generalized}.
		(a) Blurred input.
		(b) Ground truth optical flow.
		(c) Deblurred result by segmentation based method~\cite{wulff2014modeling}.
		(d) and (e) are the deblurred results using flow in (b) based on linear assumption [15] and our non-linear model, respectively.
		(d) Ground truth image.
	}
	\label{fig-deblur-GTflow}
\end{figure}

\section{Experimental Results}

In this section, we first analyze and show the effects of the semantic segmentation and PWNLK model.
We then evaluate the proposed method on both synthetic and real-world blurry videos.
We compare the proposed algorithm with the state-of-the-art methods,
based on motion transformation~\cite{cho2012video},
uniform kernel~\cite{vsroubek2012robust},
piece-wise kernel \cite{wulff2014modeling},
and pixel-wise linear kernel by Kim and Lee \cite{kim2015generalized}.
%
%
%

{\flushleft \bf{Parameter settings.}}
In all experiments, we set the parameters
$\lambda=\mu_n=250$, $\beta=\gamma=0.5\lambda$, $\sigma=7$,
and $N=2$.
We initialize the parameters of the quadratic bidirectional optical flow as $a=c=0$ and $b=1$.
For fair comparisons, we use the TV-$\ell^1$ based method~\cite{wedel2009improved} to initialize optical flow as like in~\cite{kim2015generalized}.
We also use the state-of-the-art semantic segmentation method~\cite{ghiasi2016laplacian} to segment images first, and
refine the results based on the proposed algorithm.
In addition, we use the method in~\cite{kim2015generalized} to estimate the camera
duty cycle $\tau$.

\begin{figure}[t]\small
	\begin{center}
			\includegraphics[width = 0.48\textwidth]{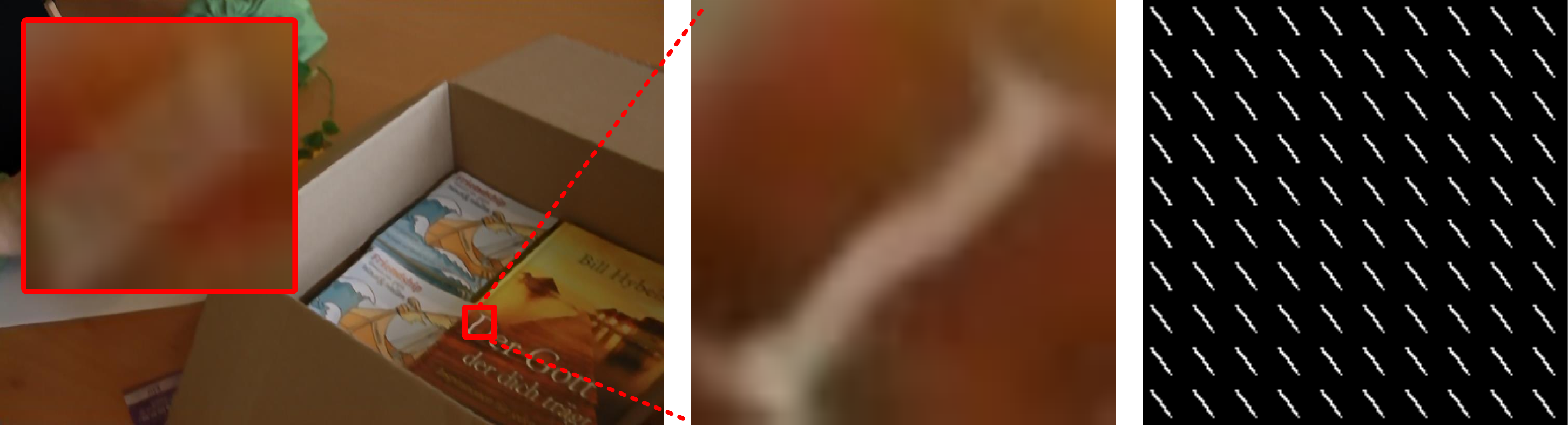} \\
			(a) Deblurred result by the linear approximation method \cite{kim2015generalized} \\
			\includegraphics[width = 0.48\textwidth]{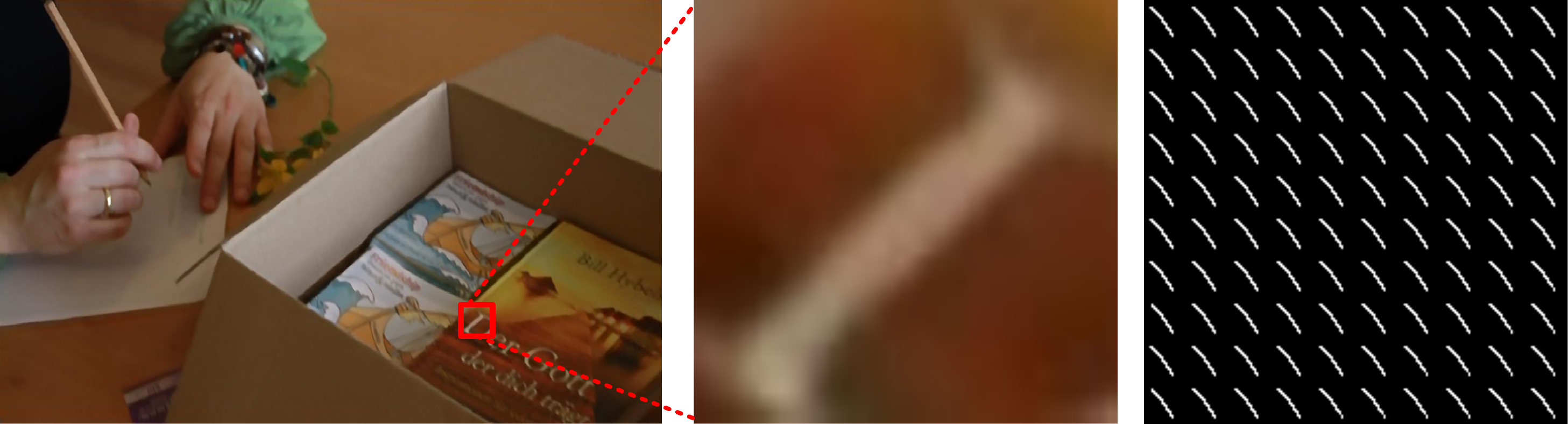} \\
			(b) Deblurred result by our non-linear approximation method.\\
	\end{center}
    \vspace{-2mm}
	\caption{Effects of PWNLK.
		(a) Deblurred results and estimated kernel by linear approximation method \cite{kim2015generalized}.
		(b) Deblurred results and estimated kernel by the proposed non-linear
		approximation approach~\eqref{eq-pwnl}.
		The highlighted area in the red rectangle is the corresponding blurry input.
		The recovered kernel in (a) is almost straight, which results in the deblurred result has some distortion artifacts. In contrast, the estimated kernel by the proposed PWNLK model is more close to real situation, and results in the recovered image is visually more pleasing.}
	\label{fig:role_k}
\end{figure}
\begin{figure}[t]\scriptsize
	\begin{center}
		\begin{tabular}{@{}ccc@{}}
			\includegraphics[width = 0.155\textwidth,height = 0.1\textwidth]{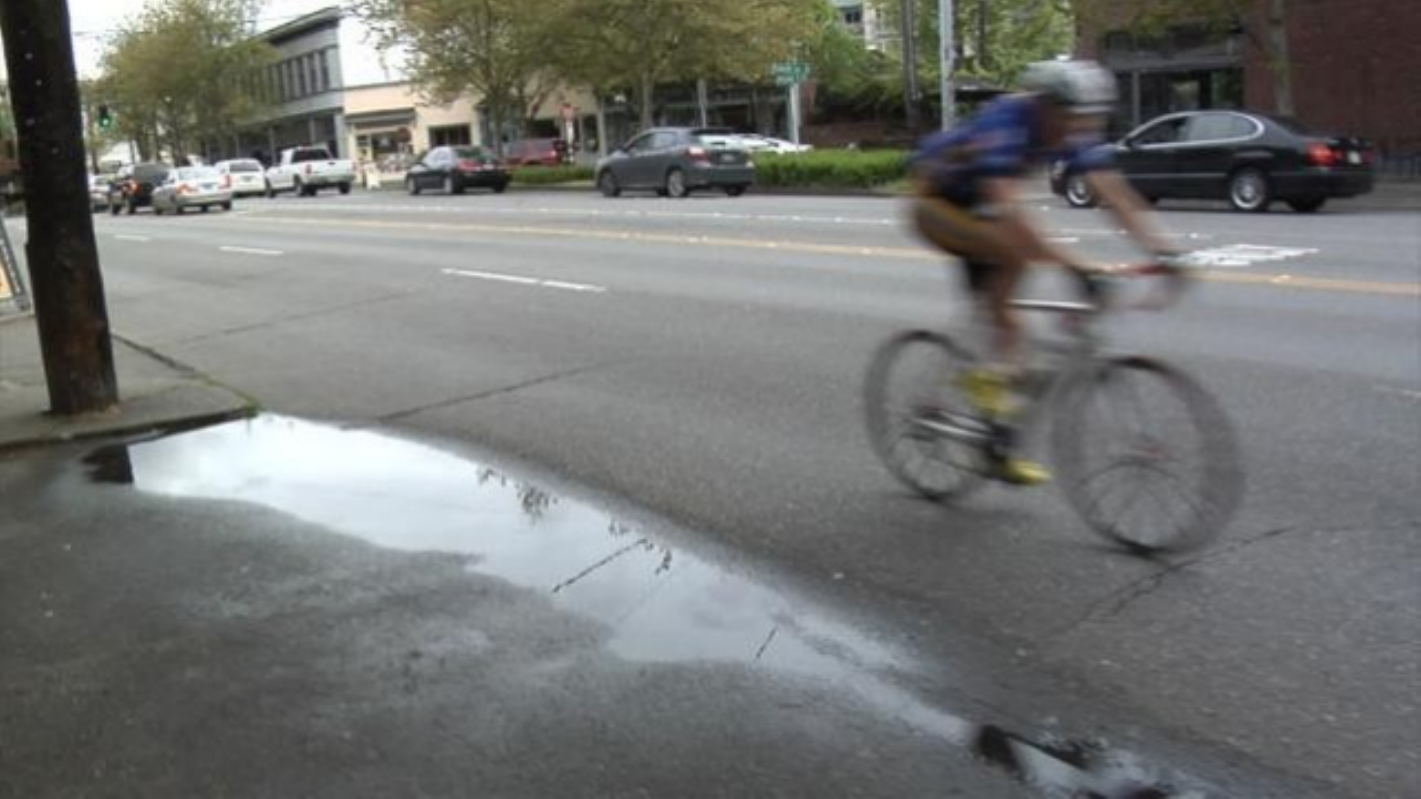} & \hspace{-0.4cm}
			\includegraphics[width = 0.155\textwidth,height = 0.1\textwidth]{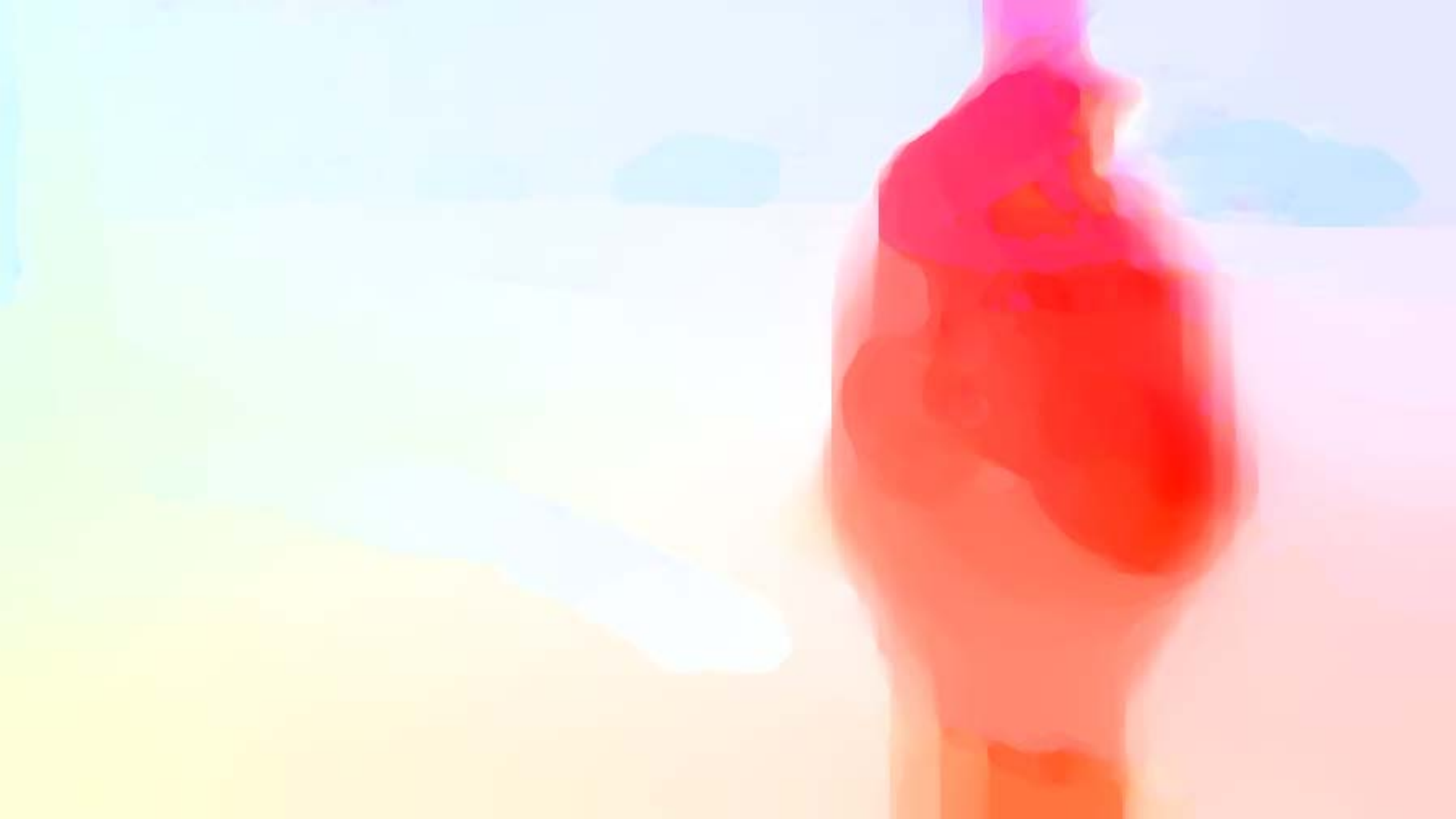} & \hspace{-0.4cm}
			\includegraphics[width = 0.155\textwidth,height = 0.1\textwidth]{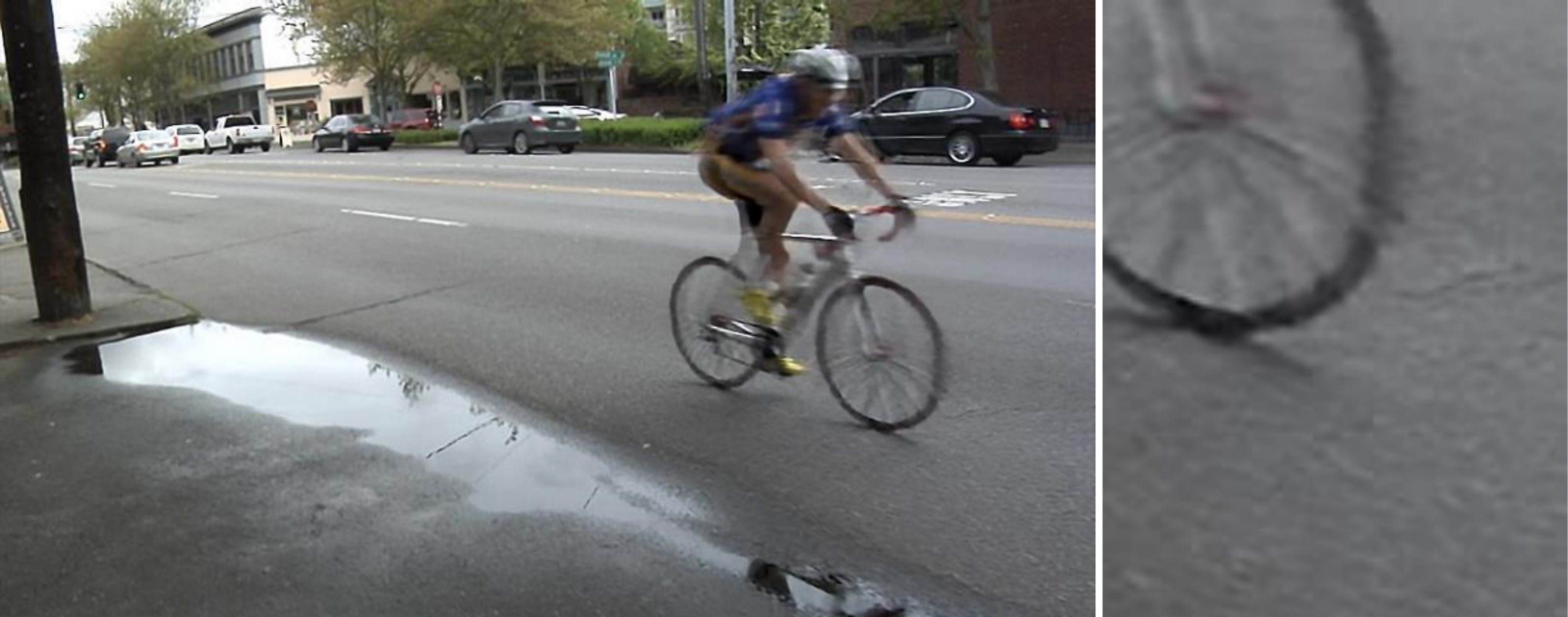} \\
			(a) Blurry frame &  \hspace{-0.4cm}
			(b) Optical flow by~\cite{kim2015generalized} & \hspace{-0.4cm}
			(c) Without segmentation \\
			\includegraphics[width = 0.155\textwidth,height = 0.1\textwidth]{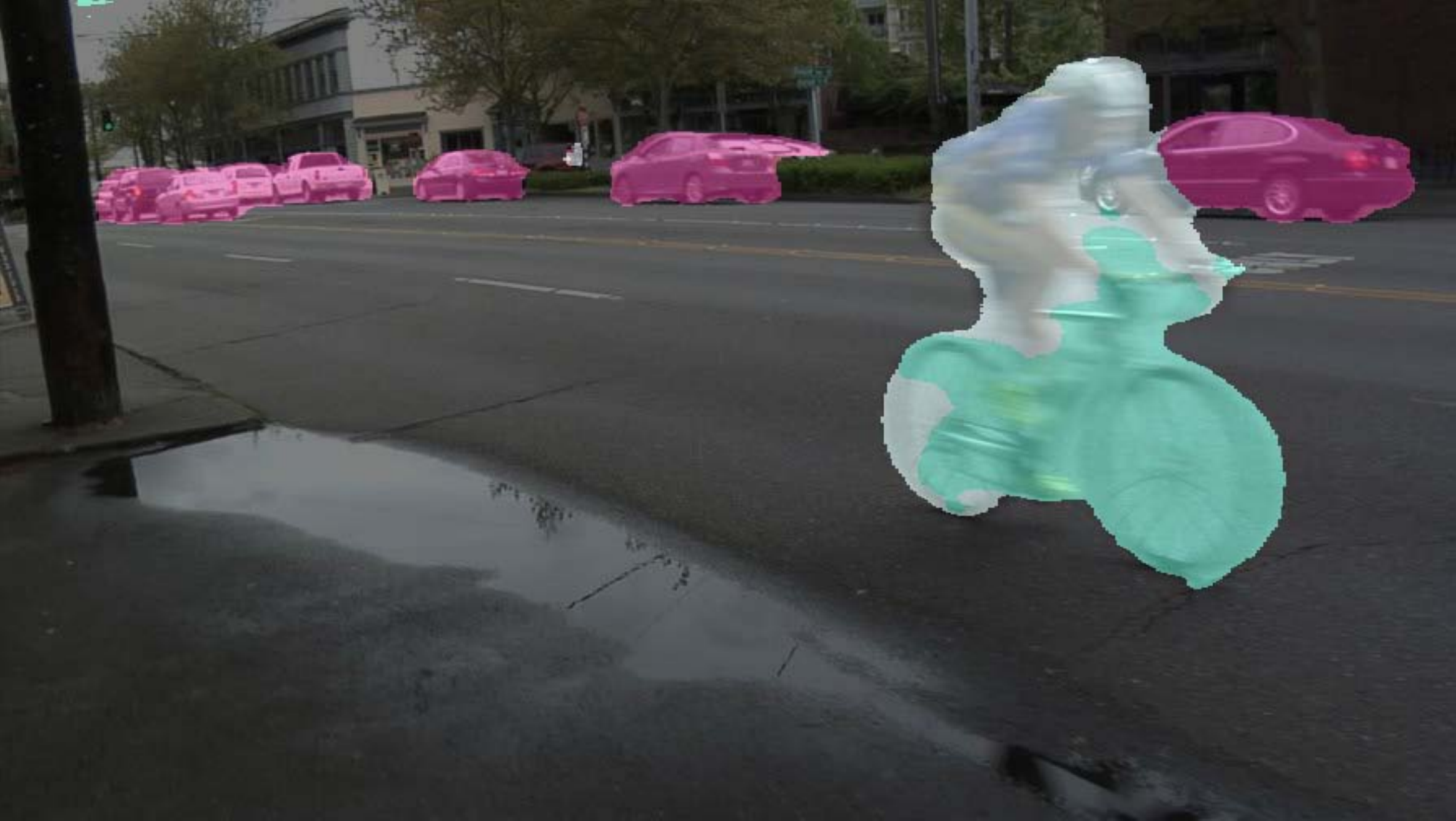} & \hspace{-0.4cm}
			\includegraphics[width = 0.155\textwidth,height = 0.1\textwidth]{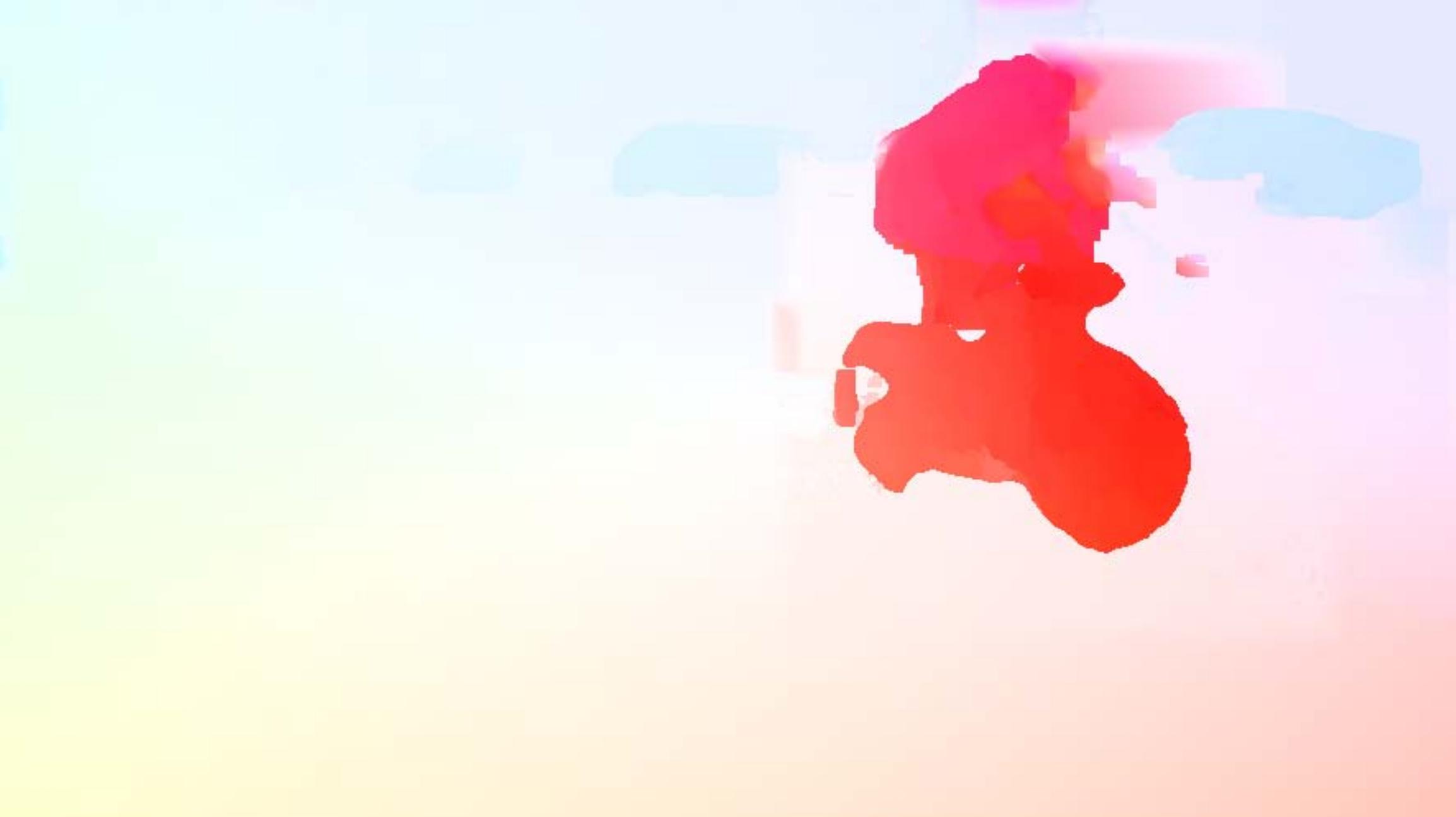} & \hspace{-0.4cm}
			\includegraphics[width = 0.155\textwidth,height = 0.1\textwidth]{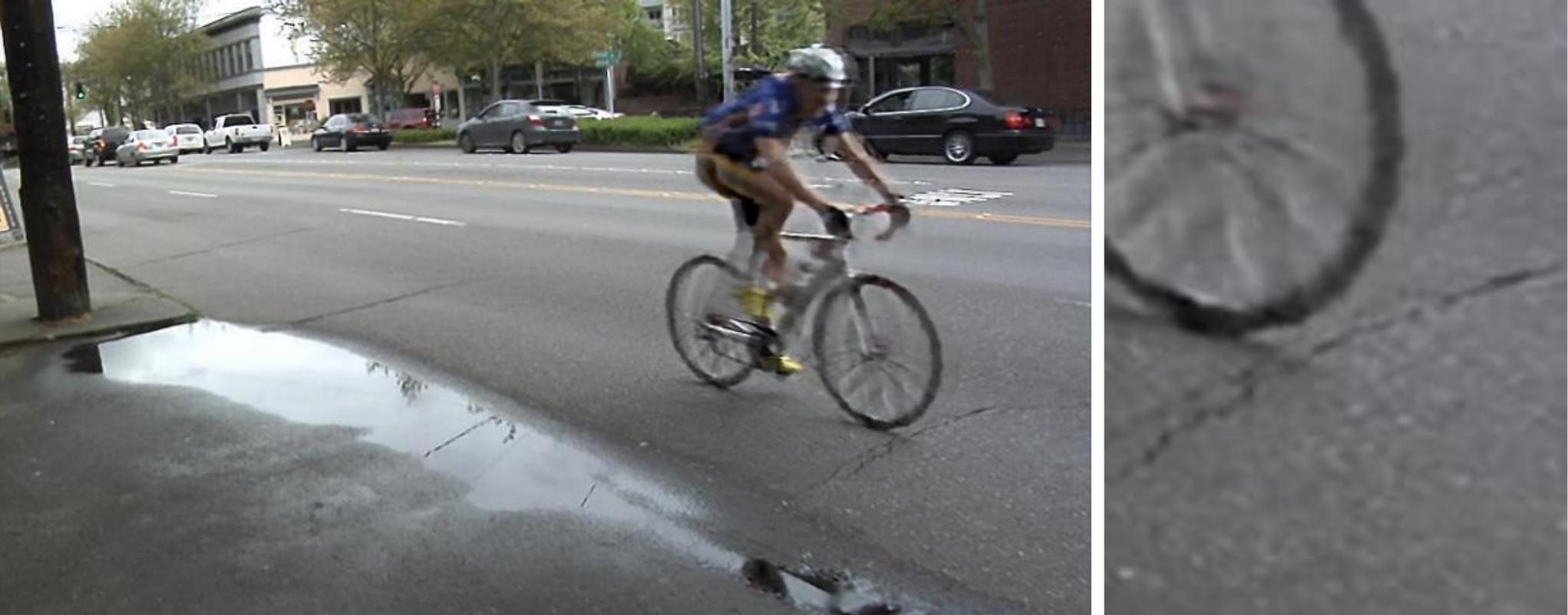} \\
			(d) Our segmentation & \hspace{-0.4cm}
			(e) Our optical flow & \hspace{-0.4cm}
			(f) With segmentation
		\end{tabular}
	\end{center}
    \vspace{-2mm}
	\caption{Effects of semantic segmentation on deblurring.
		(a) Blurry input.
		(b) and (c) are estimated optical flow and deblurred result by~\cite{kim2015generalized}.
		(d) Our segmentation results (semantic color coded using \cite{sevilla2016optical}).
		(e) and (f) are estimated optical flow and deblurred result with the proposed semantic segmentation.
		The background and road regions in (c) are over-smoothed due to the inaccurate estimated optical flow in (b).
	}
	\label{fig-role-segmen-bike}
\end{figure}
\subsection{Analysis of Proposed Method}
\label{sec-anal}
{\flushleft \textbf{Effects of PWNLK model.}}
We note that~\cite{kim2015generalized} directly uses the linear bidirectional optical flow to restore the clear images.
As mentioned in Figure~\ref{fig:ker-from-flow}, this method is less effective since motion trajectories in videos are different from optical flow.
Figure~\ref{fig-deblur-GTflow}(a) shows an example where the blurred image is
generated by affine transformation~\cite{wulff2014modeling}.
%
We first show the deblurred result by the layer based method~\cite{wulff2014modeling} in Figure~\ref{fig-deblur-GTflow}(c).
Note that there are significant artifacts around the \textit{elephant} boundary since
the inaccurate segmentation.
As shown in Figure~\ref{fig-deblur-GTflow}(d), the restored image
generated by the ground truth optical flow (Figure~\ref{fig-deblur-GTflow}(b)) using the pixel-wise linear kernel method~\cite{kim2015generalized} contains significant ring artifacts,
which demonstrates that the linear bidirectional optical flow cannot model motion blur well.
%

Figure~\ref{fig:role_k} shows an example which is able to demonstrate the effectiveness of the PWNLK model.
We use the same optical
flow to estimate the pixel-wise linear and non-linear kernel.
We note that the linear assumption of motion blur for each pixel does not hold in real images
as shown in Figure~\ref{fig:role_k}(a).
The estimated motion blur kernel
using linear approximation
for the zoomed-in region is almost straight and the corresponding deblurred results
contain distortion artifacts on the line of letter \textit{D}.
%
%
%
The trajectories of the estimated motion kernel by the
proposed non-linear approximation method
coincide well with the real motion blur trajectories and the corresponding deblurred
image is much clearer and contains fewer artifacts as shown in Figure~\ref{fig:role_k}(b),
%
%
which indicate that the proposed blur
model~\eqref{eq-motion-blur-model} can better approximate motion trajectories in real scenes.
%


\begin{figure}[t]\scriptsize
	\begin{center}
		\begin{tabular}{@{}cccc@{}}
			\includegraphics[width = 0.11\textwidth,height = 0.08\textwidth]{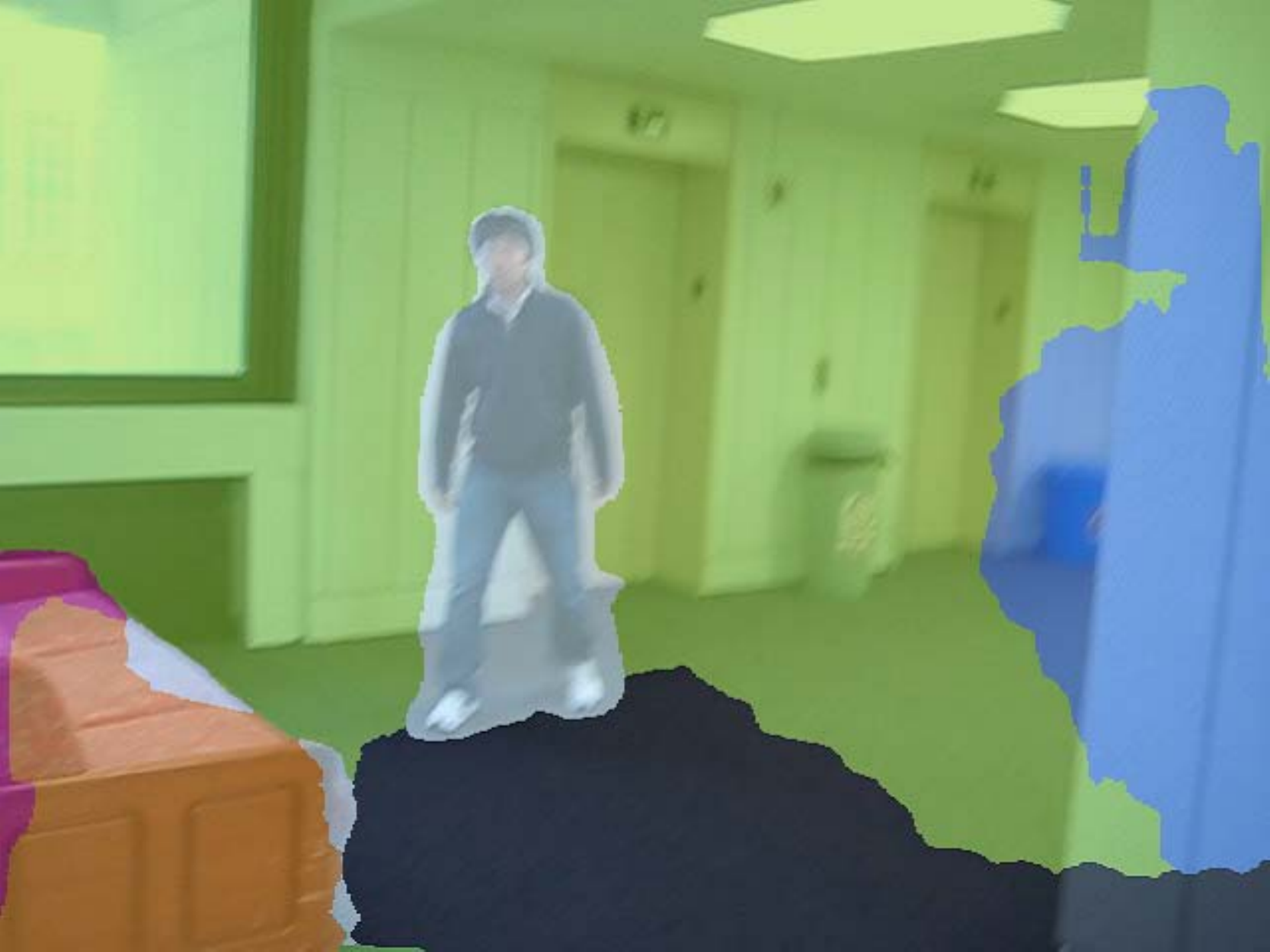} & \hspace{-0.4cm}
			\includegraphics[width = 0.11\textwidth,height = 0.08\textwidth]{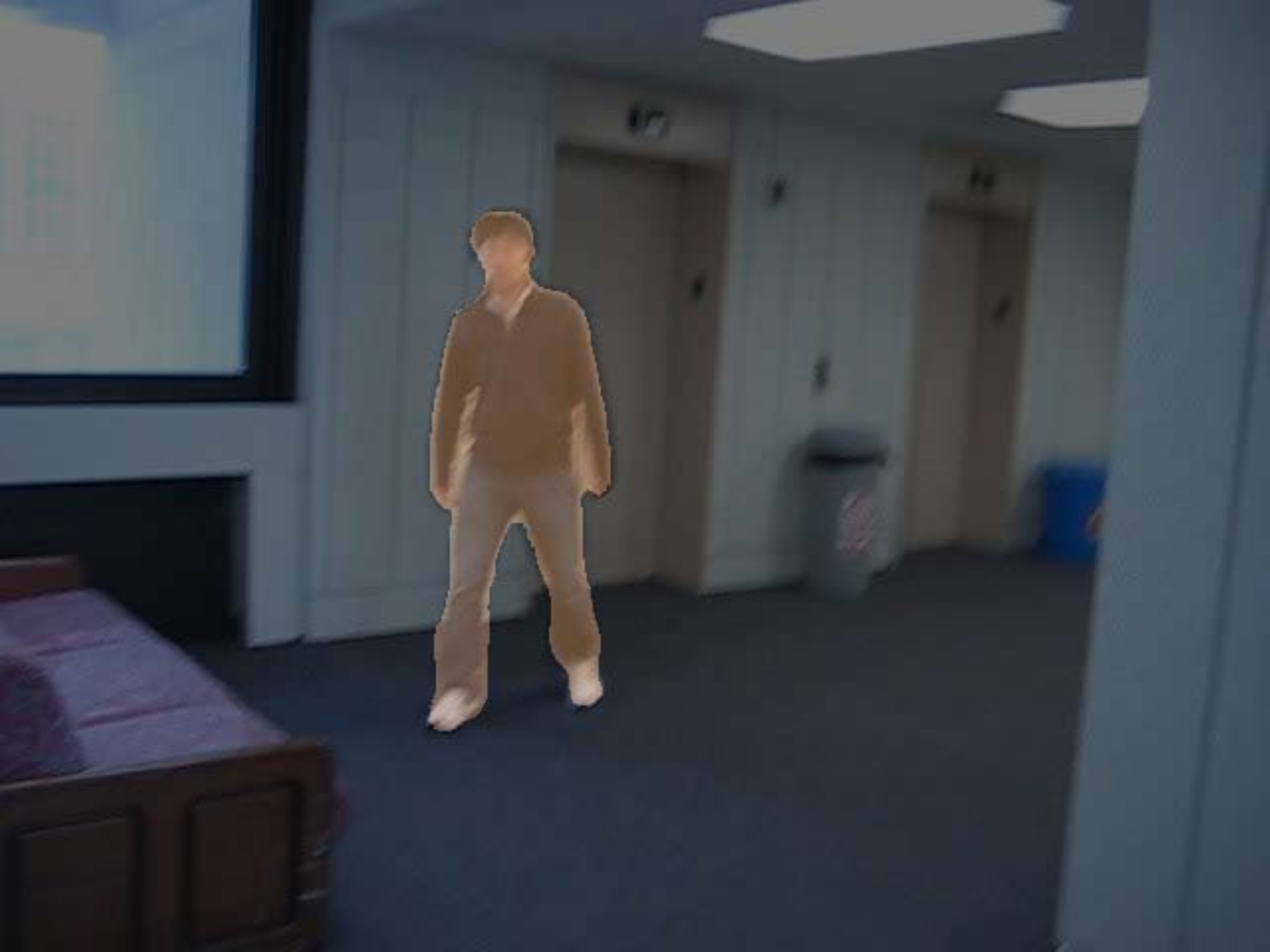} & \hspace{-0.4cm}
			\includegraphics[width = 0.11\textwidth,height = 0.08\textwidth]{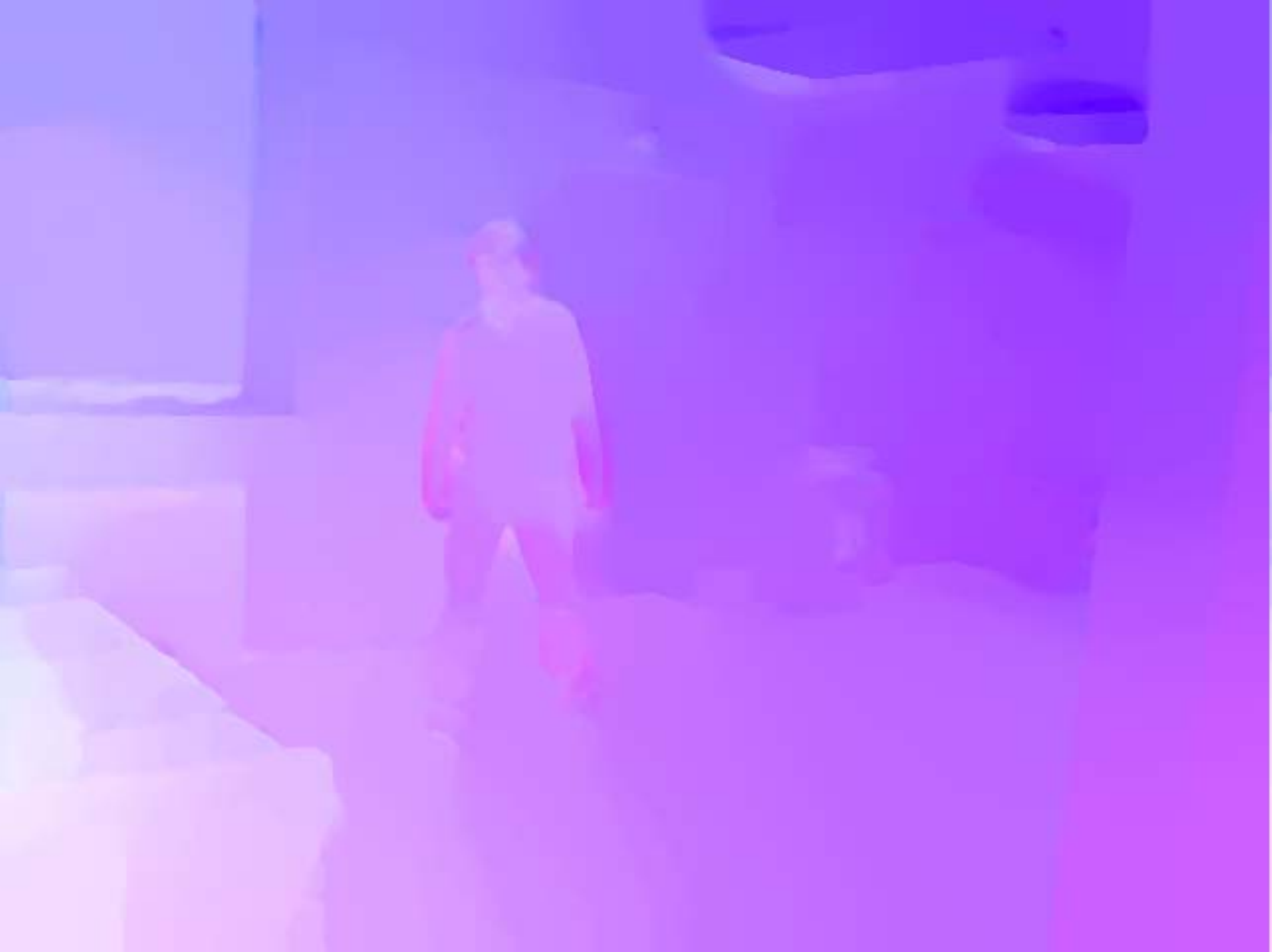} & \hspace{-0.4cm}
			\includegraphics[width = 0.11\textwidth,height = 0.08\textwidth]{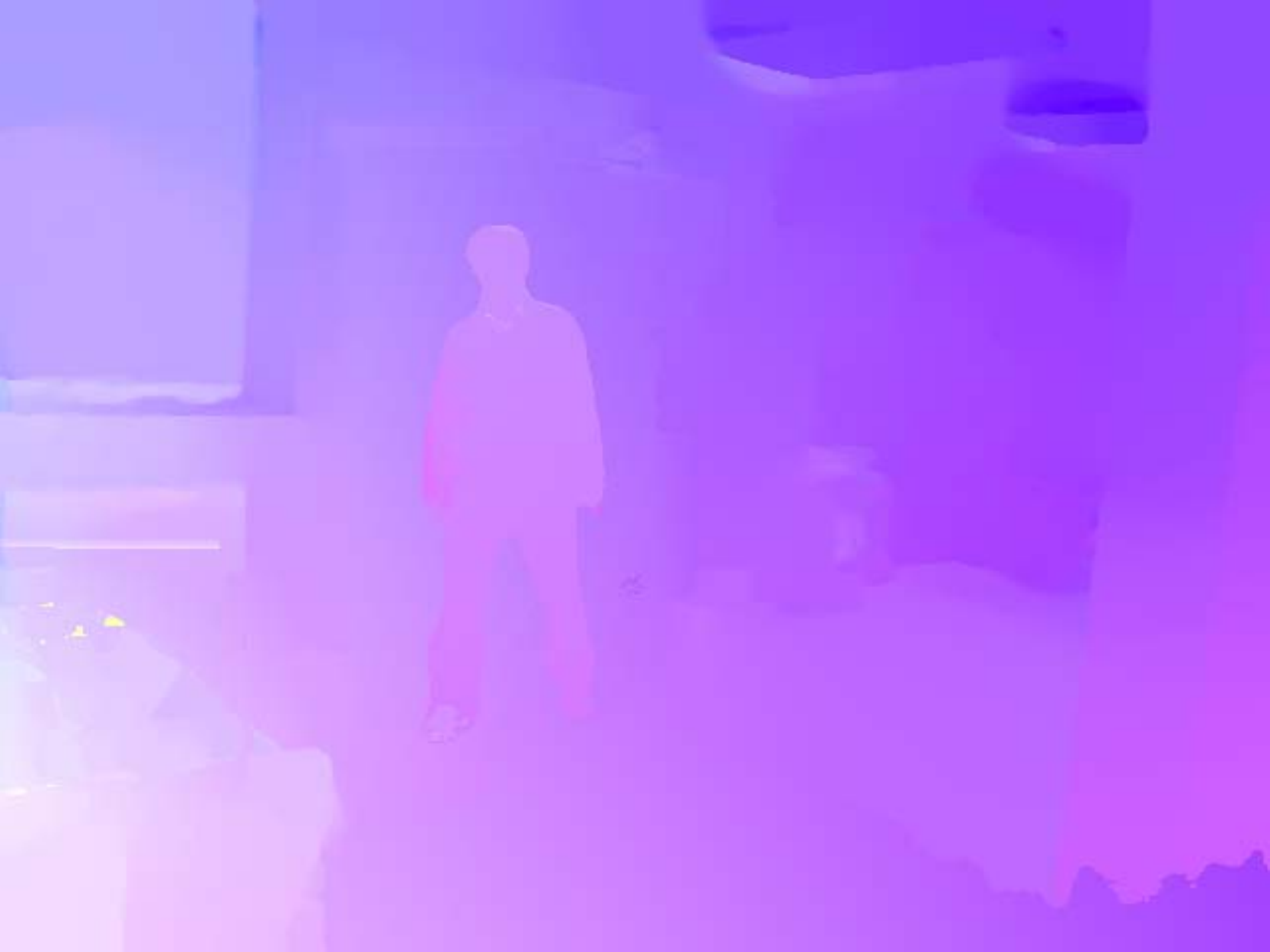} \\
			\includegraphics[width = 0.11\textwidth,height = 0.08\textwidth]{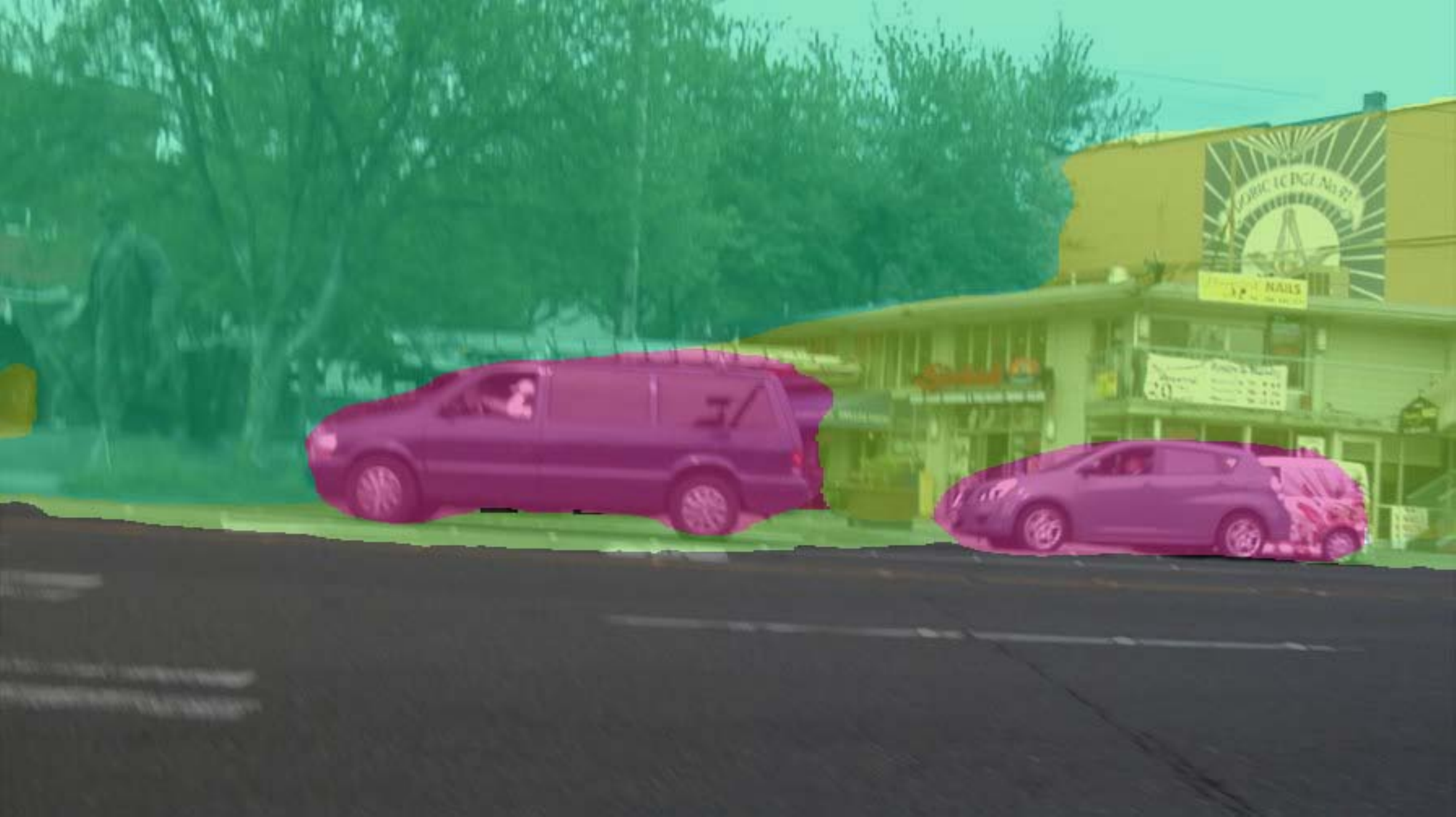} & \hspace{-0.4cm}
			\includegraphics[width = 0.11\textwidth,height = 0.08\textwidth]{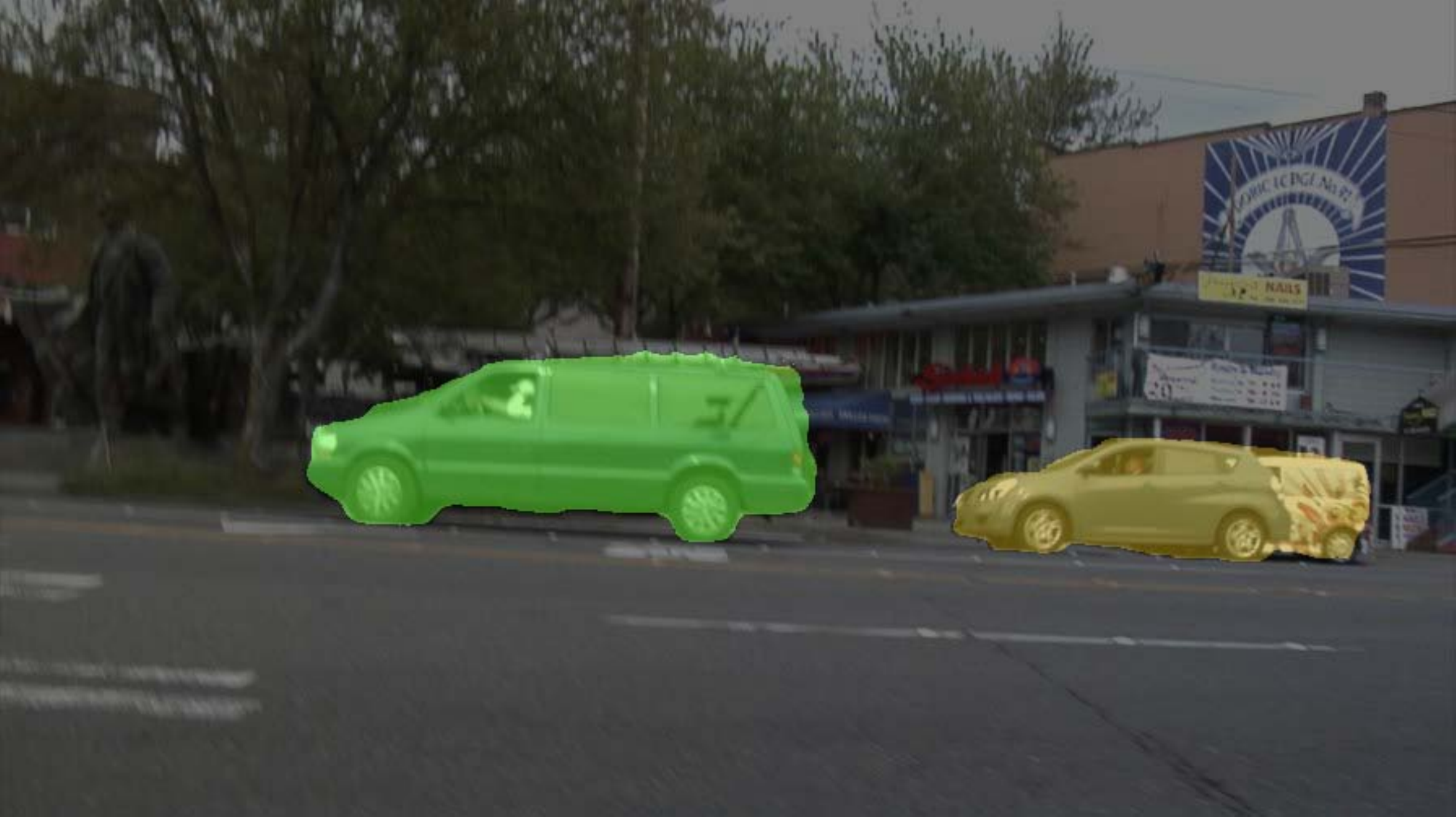} & \hspace{-0.4cm}
			\includegraphics[width = 0.11\textwidth,height = 0.08\textwidth]{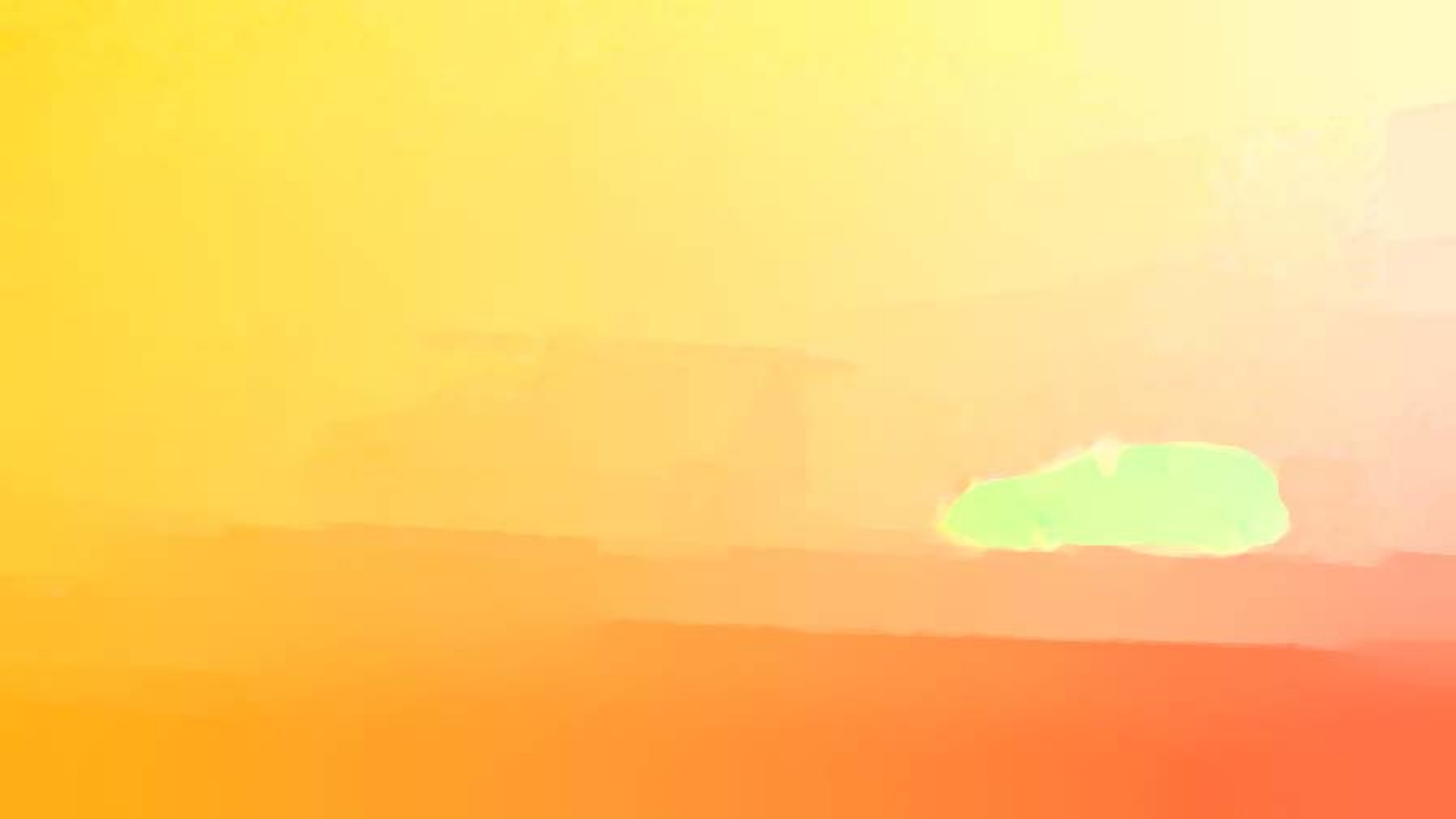} & \hspace{-0.4cm}
			\includegraphics[width = 0.11\textwidth,height = 0.08\textwidth]{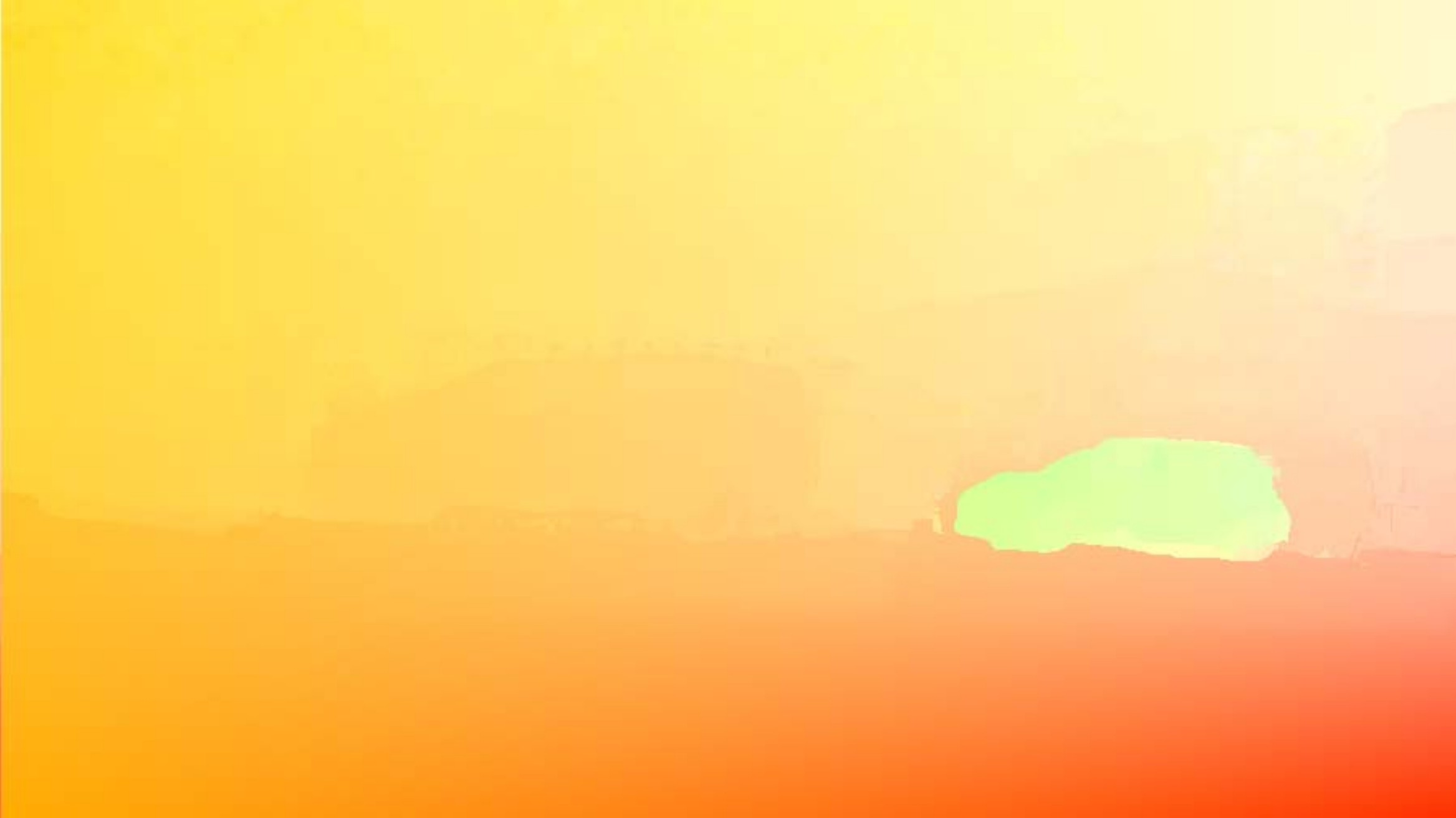} \\
			\includegraphics[width = 0.11\textwidth,height = 0.08\textwidth]{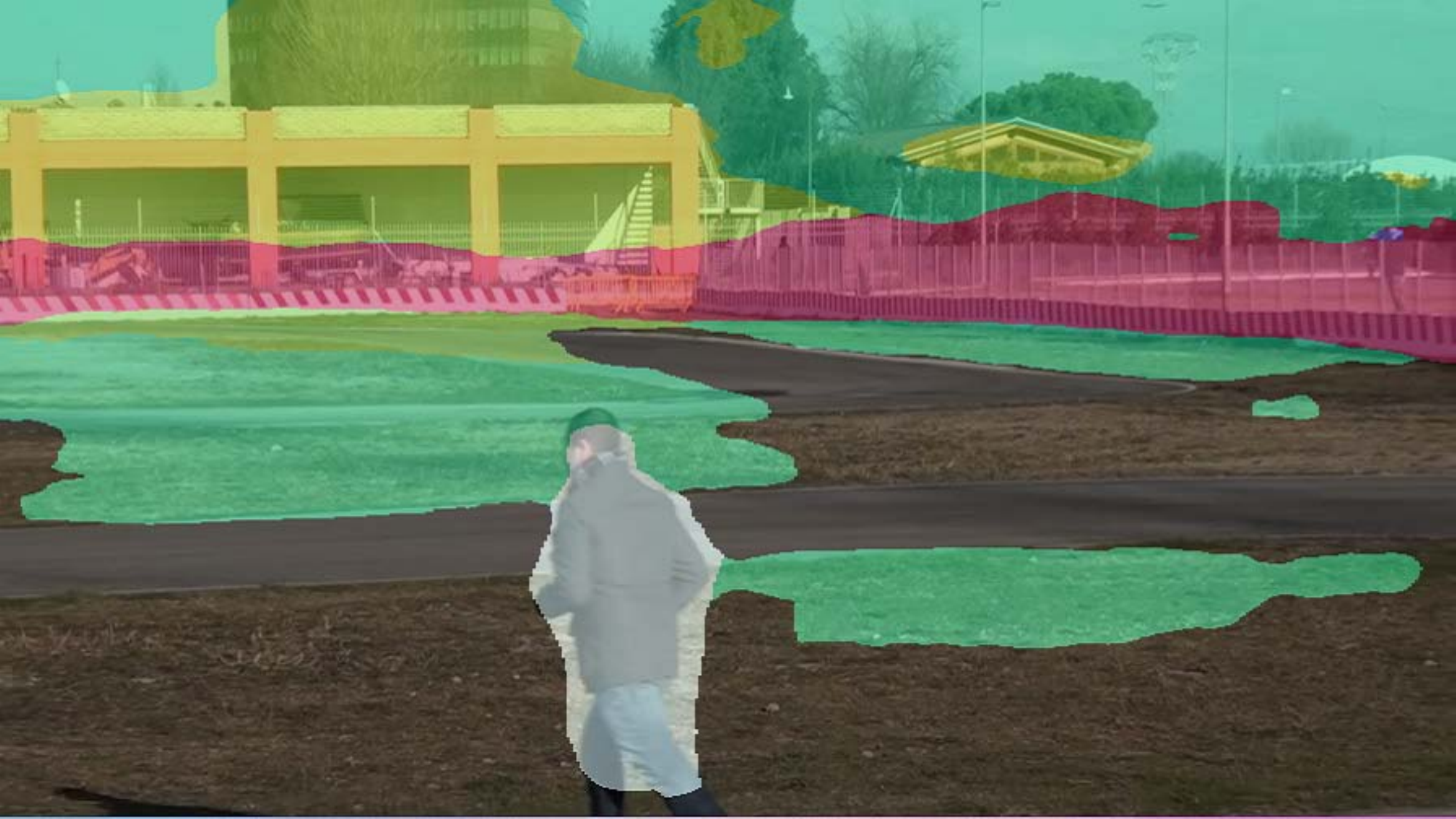} & \hspace{-0.4cm}
			\includegraphics[width = 0.11\textwidth,height = 0.08\textwidth]{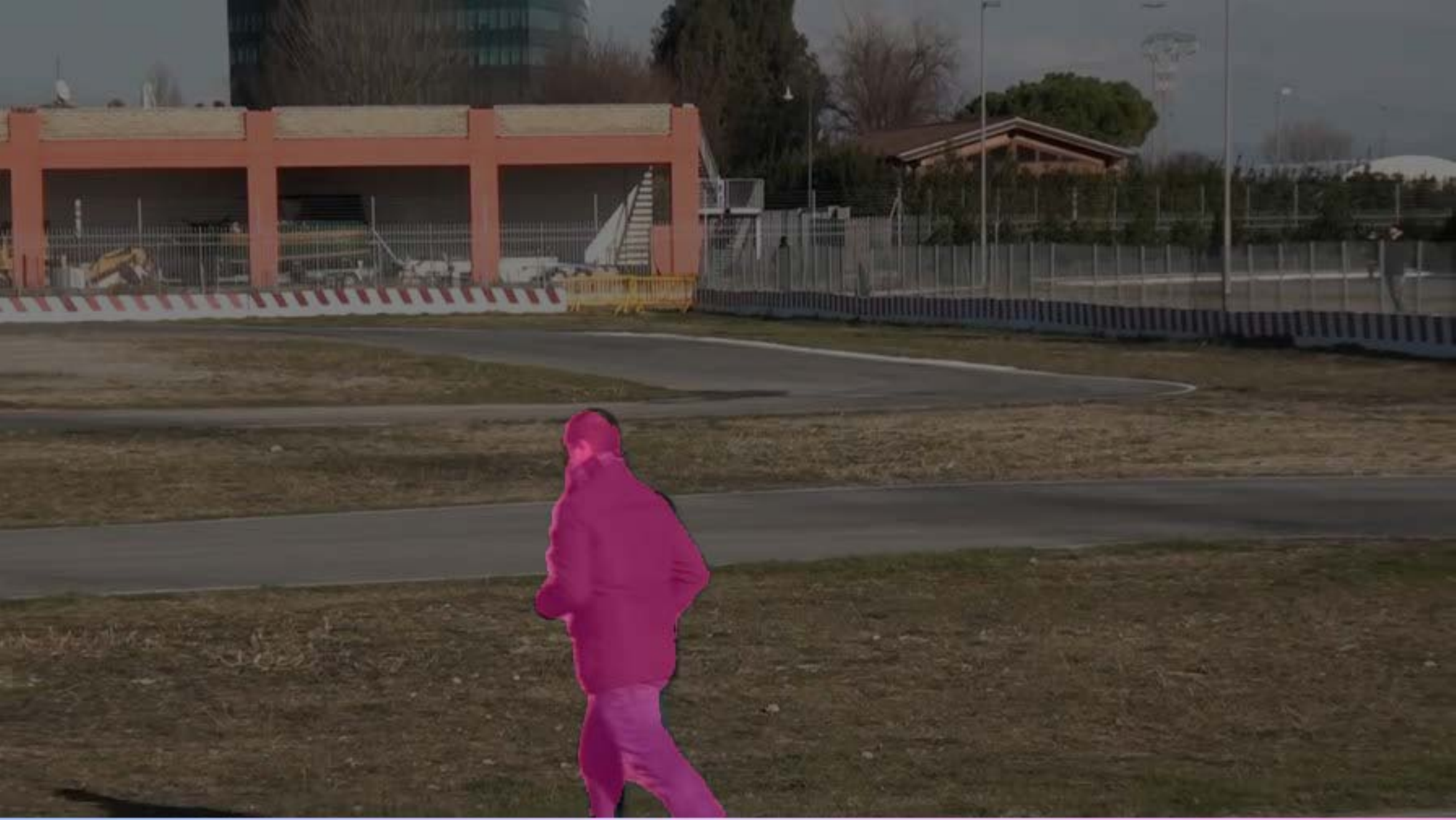} & \hspace{-0.4cm}
			\includegraphics[width = 0.11\textwidth,height = 0.08\textwidth]{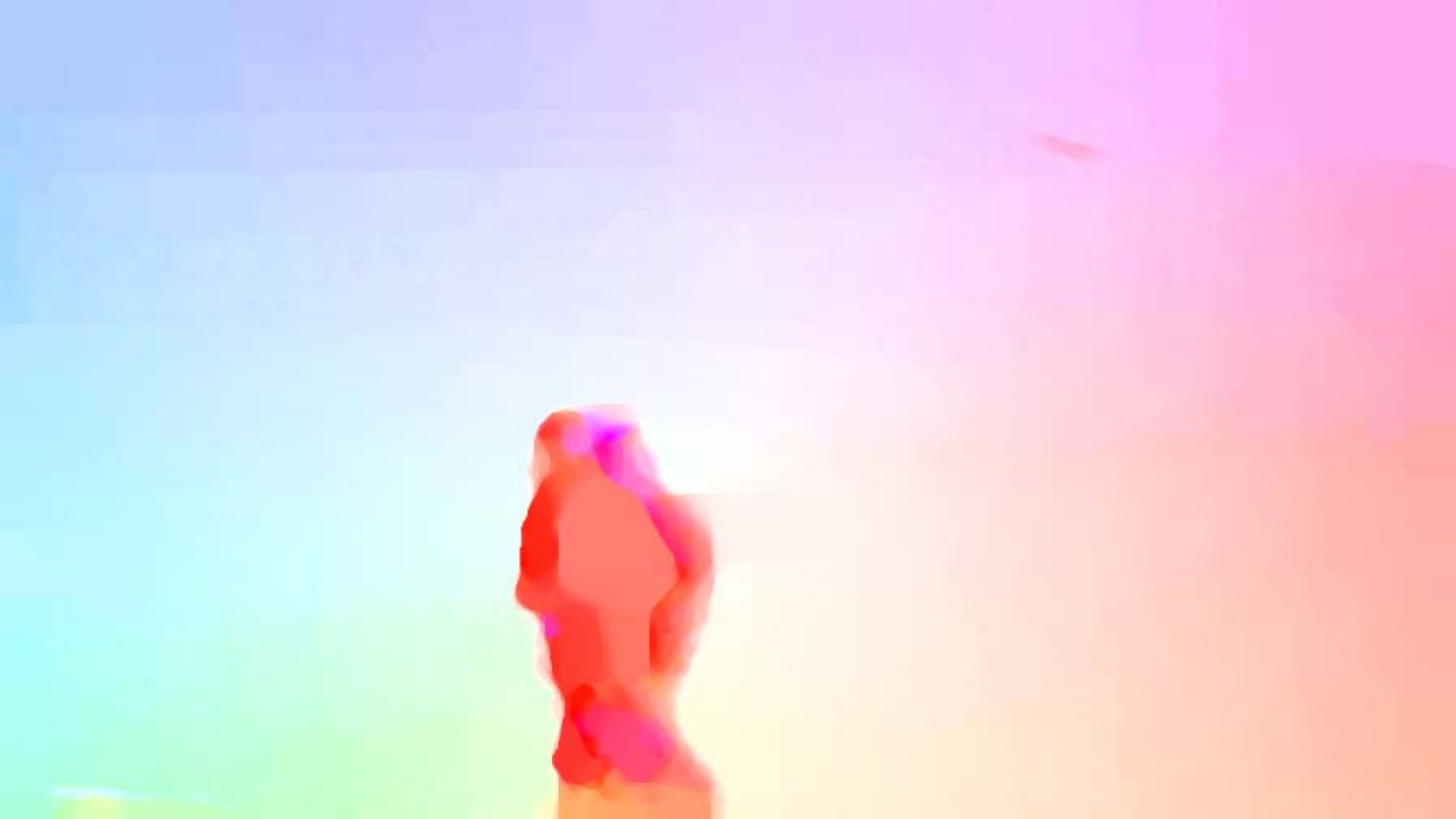} & \hspace{-0.4cm}
			\includegraphics[width = 0.11\textwidth,height = 0.08\textwidth]{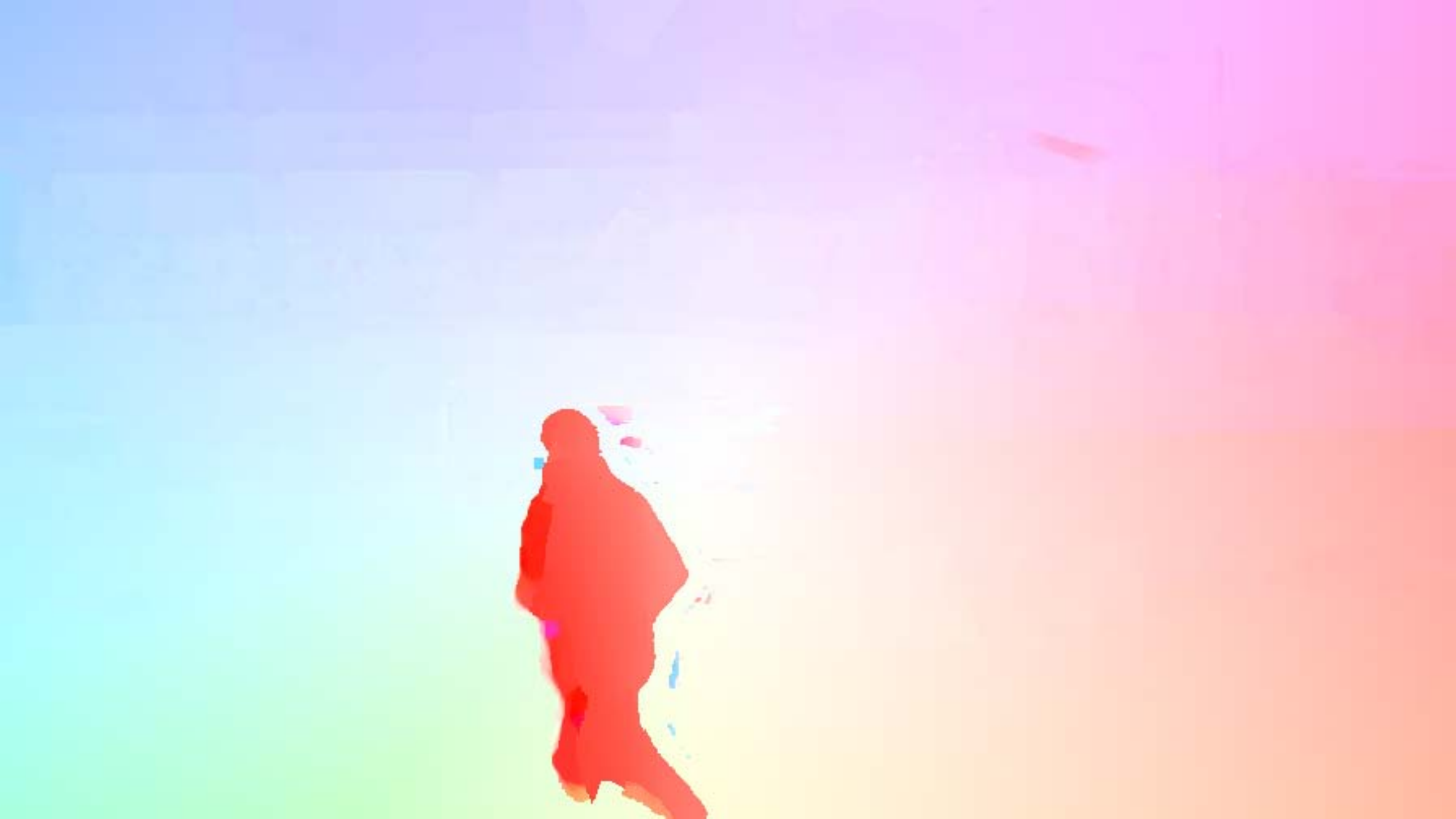} \\
			(a) & \hspace{-0.4cm}
			(b) & \hspace{-0.4cm}
			(c) & \hspace{-0.4cm}
			(d)
		\end{tabular}
	\end{center}
    \vspace{-2mm}
	\caption{Qualitative analysis of semantic segmentation.
		(a) Blurry input and initialized segmentation results~\cite{ghiasi2016laplacian}.
		(b) Our refined segmentation.
		(c) Optical flow by~\cite{kim2015generalized}.
		(d) Our optical flow.
	}
	\label{fig-role-segmen-more}
\end{figure}

{\flushleft \textbf{Effects of semantic segmentation.}}
Semantic segmentation improves video deblurring in multiple ways
as it is used to help estimate optical flow from which the blur kernel is estimated.
First, it provides region information about object boundaries.
Second, as different objects (layers) move differently, semantic segments are
used to constrain optical flow estimation of each region.
As shown in Figure~\ref{fig-role-segmen-bike}(b),
the estimated optical flow is over-smoothed around
the bicycle when semantic segmentation is not used.
Consequently, the deblurred results for the background and road regions are over-smoothed.
In contrast, the semantic segmentation results by the proposed algorithm describe
boundaries well and help generate accurate optical flow.
As shown in Figure~\ref{fig-role-segmen-bike}(f),
the deblurred images by the proposed algorithm are clear with fine details.

In addition, we carry out more experiments to examine the effects of semantic segmentation for optical flow estimation.
Although the initialized segmentations are inaccurate as shown in Figure~\ref{fig-role-segmen-more}(a),
the proposed algorithm can precisely segment the moving objects (Figure~\ref{fig-role-segmen-more}(b)) and provide more
accurate motion boundaries information for optical flow estimation,
and thereby facilitates video deblurring.
\subsection{Real Datasets}
\label{sec-real}
\begin{figure}[t]\small
	\begin{center}
		\begin{tabular}{@{}cccc@{}}
			\includegraphics[width = 0.16\textwidth,height = 0.18\textwidth]{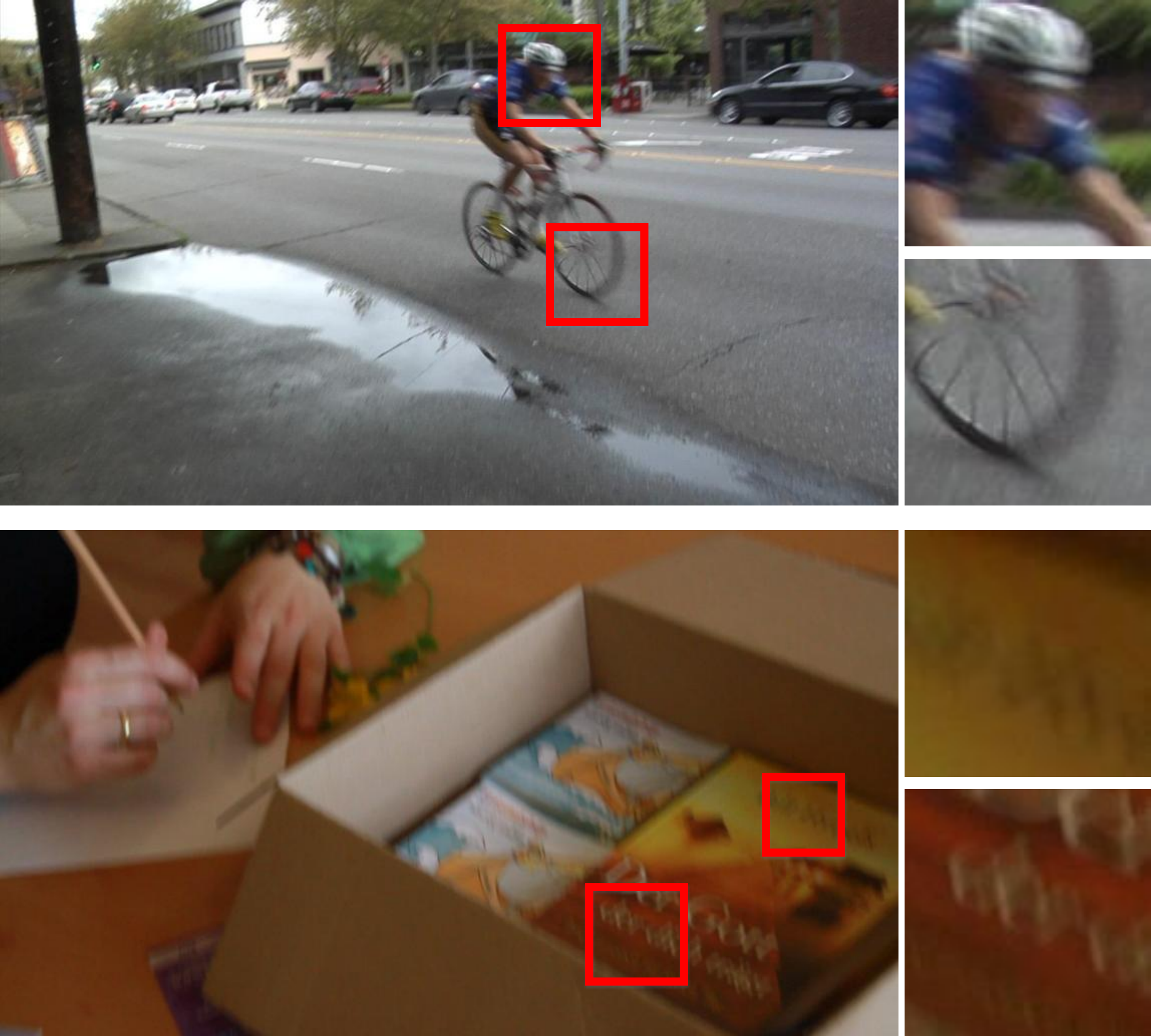} & \hspace{-0.4cm}
			\includegraphics[width = 0.16\textwidth,height = 0.18\textwidth]{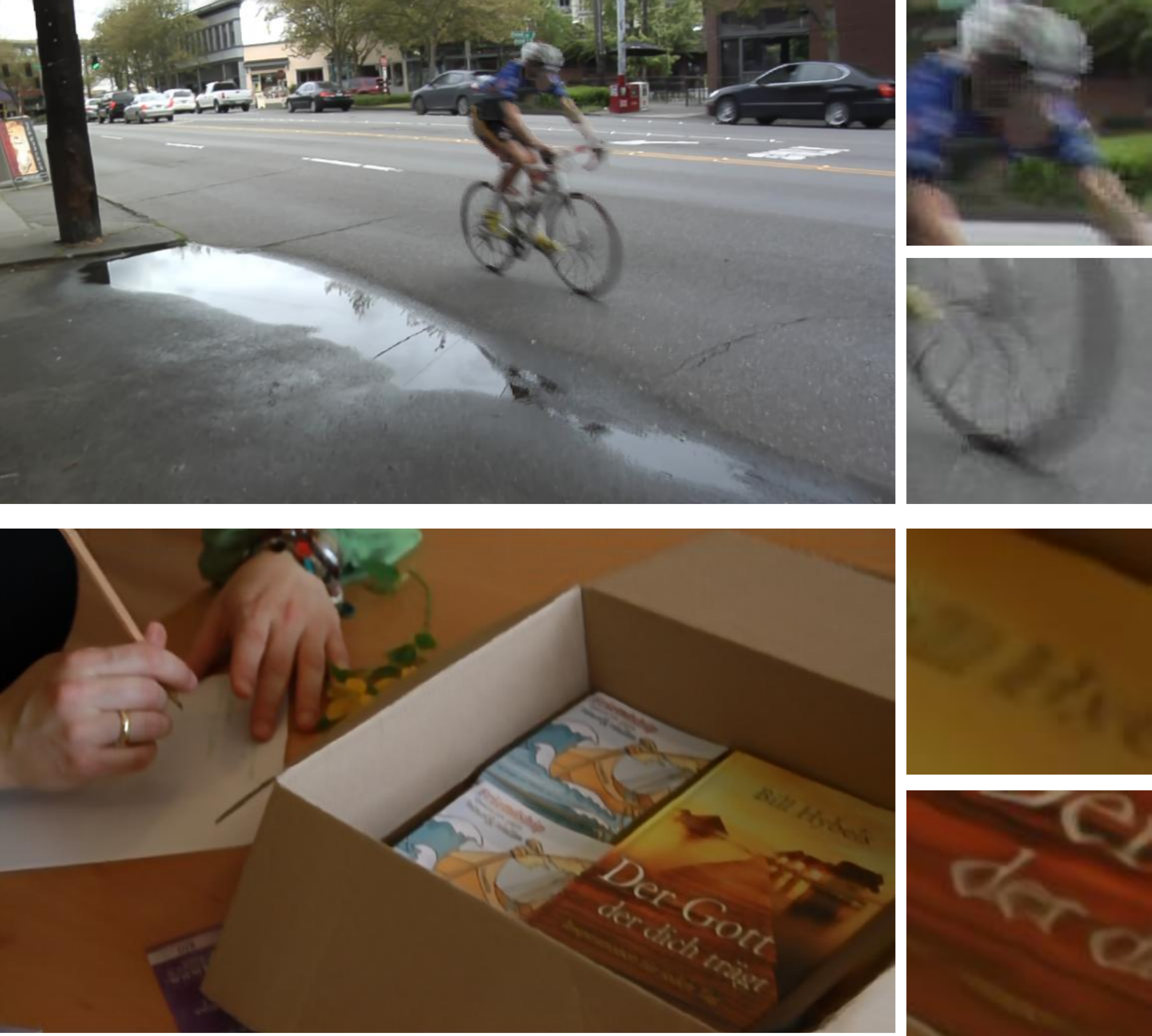} & \hspace{-0.4cm}
			\includegraphics[width = 0.16\textwidth,height = 0.18\textwidth]{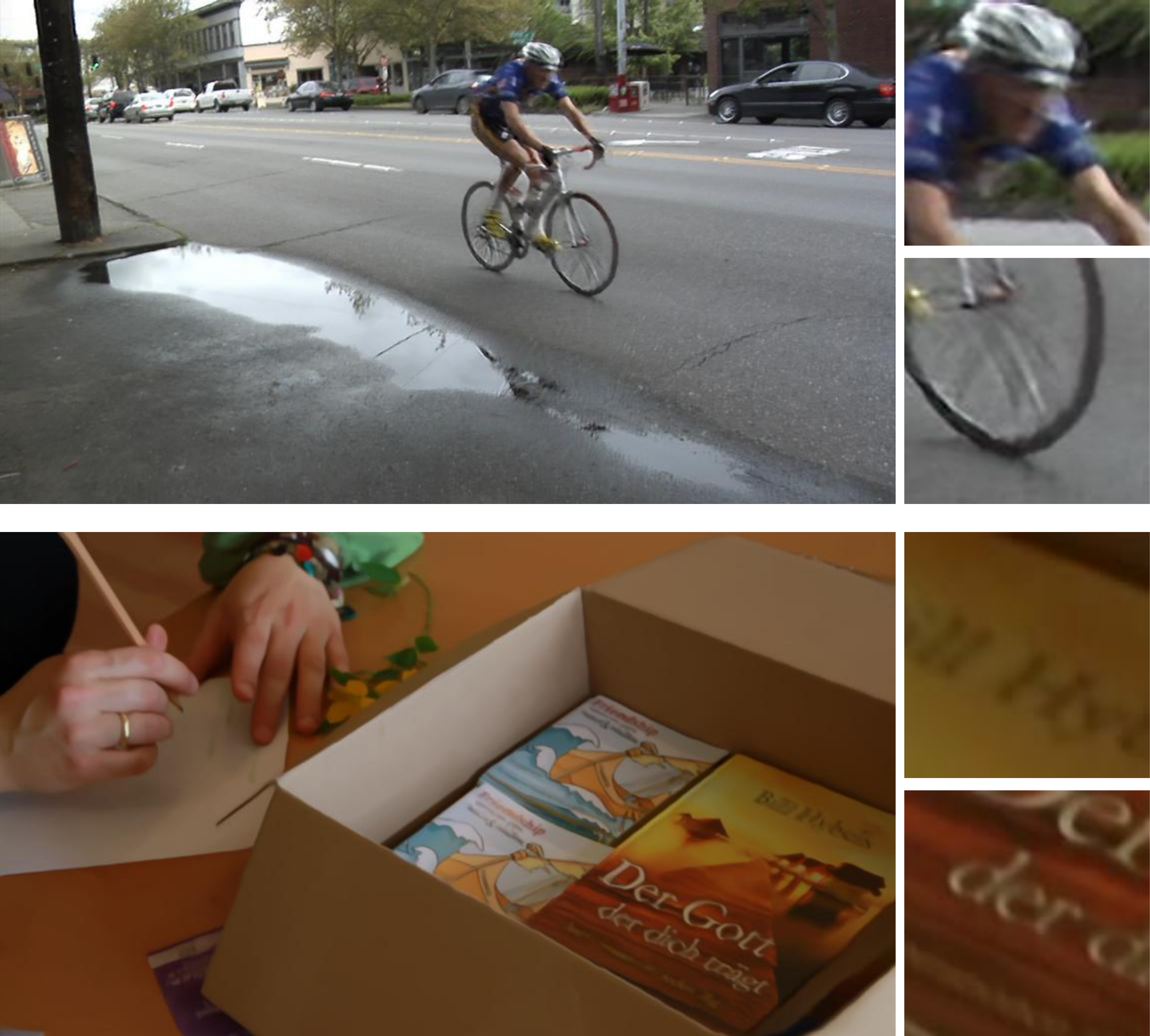} \\
			(a) Blurry frame & \hspace{-0.4cm}
			(b) Cho~\etal \cite{cho2012video} & \hspace{-0.4cm}
			(c) Our results
		\end{tabular}
	\end{center}
	\caption{Comparisons with transformation based
		method~\cite{cho2012video}.}
	\label{fig-tra}
\end{figure}

We evaluate the proposed algorithm
against the state-of-the-art video deblurring methods \cite{cho2012video,vsroubek2012robust,wulff2014modeling,kim2015generalized}
on real sequences from  \cite{cho2012video,wulff2014modeling}.
%
%
%
We first compare our algorithm with the
transformation based method by Cho \etal \cite{cho2012video}.
As shown in the first row of Figure~\ref{fig-tra}(b), the method \cite{cho2012video}
does not recover the moving bicycle because the object
motion is large and there are no sharp images in the nearby frames.
%
In contrast, the proposed algorithm is able to deal with the blur caused by the moving objects and generates a clear image as shown in the first row of Figure~\ref{fig-tra}(c).
The transformation based approach \cite{cho2012video}
does not handle large camera motion blur as shown in the second row of Figure~\ref{fig-tra}(b).
The recovered texts for the \textit{Books} sequence contain
significant distortion artifacts
since this transformation based method \cite{cho2012video} introduces
incorrect patch matches
%
if the clear images or sharp patches are not available.
In contrast, the proposed method based on the estimated optical flow
does not require clear images or patches.
The deblurred result is visually more pleasing especially for the texts.
\begin{figure}[t]\scriptsize
	\begin{center}
		\begin{tabular}{@{}cccc@{}}
			\includegraphics[width = 0.16\textwidth,height = 0.18\textwidth]{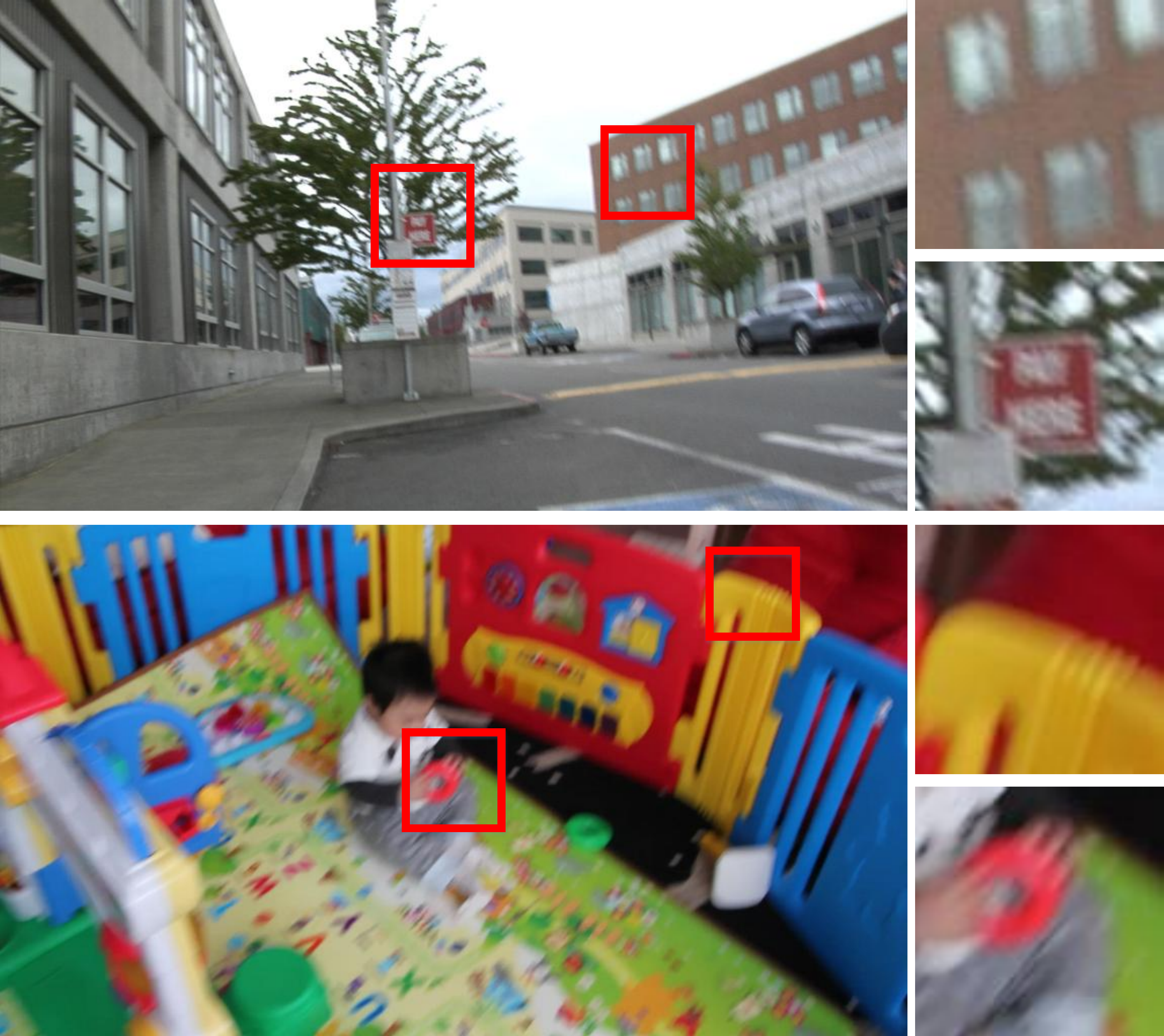} & \hspace{-0.5cm}
			\includegraphics[width = 0.16\textwidth,height = 0.18\textwidth]{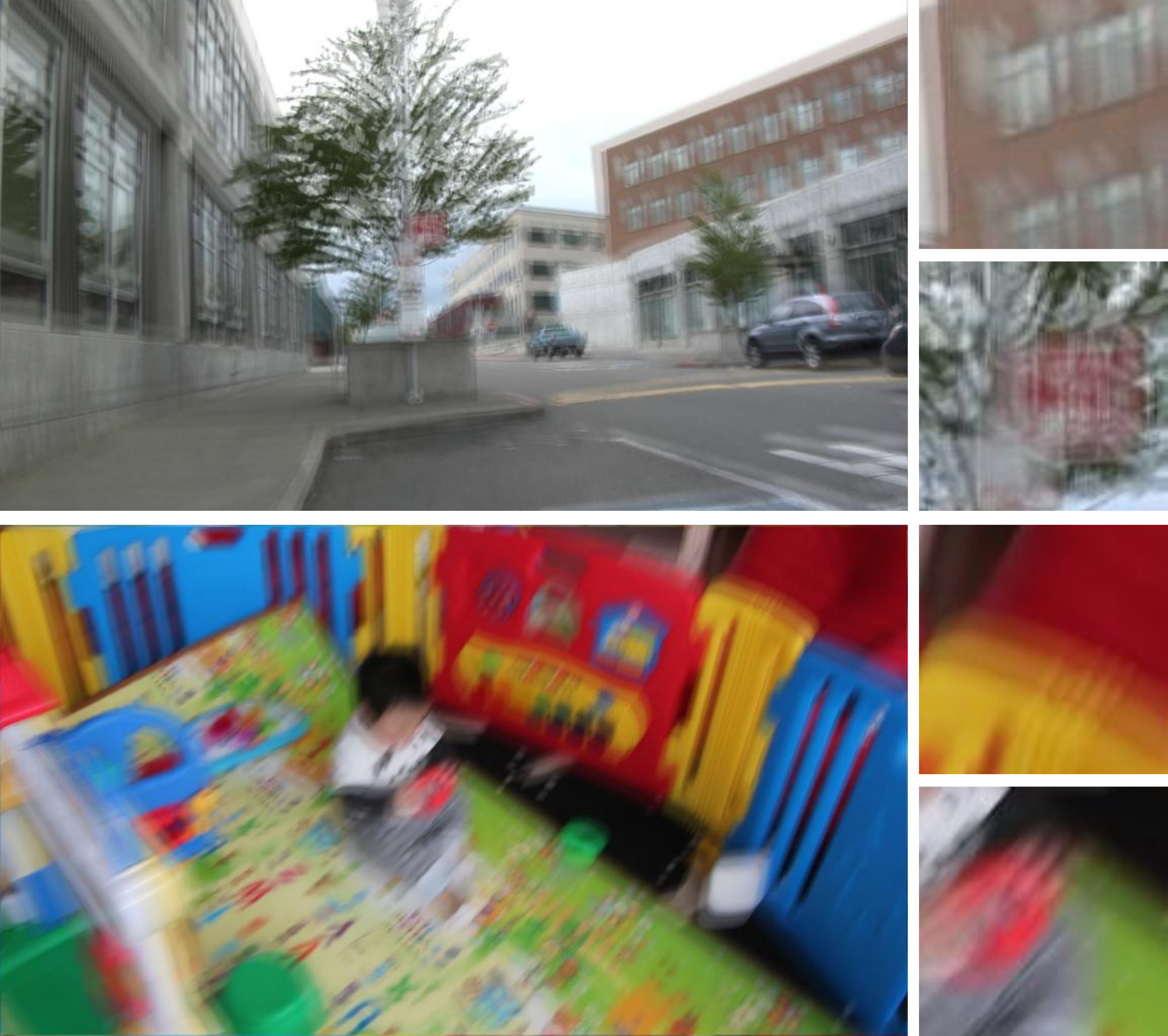} & \hspace{-0.5cm}
			\includegraphics[width = 0.16\textwidth,height = 0.18\textwidth]{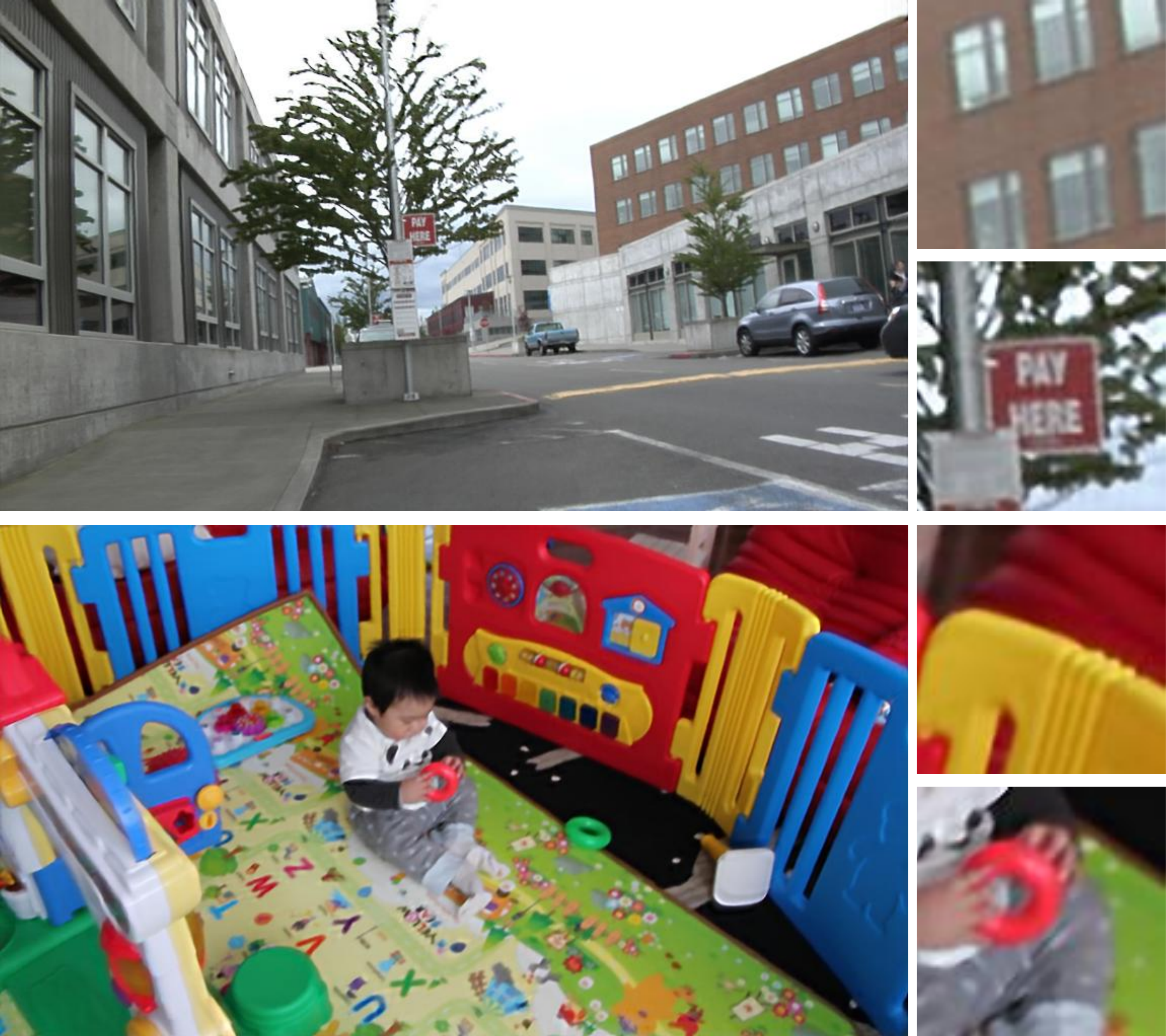} \\
			(a) Blurry frames & \hspace{-0.5cm}
			(b) \v{S}roubek and Milanfar \cite{vsroubek2012robust} & \hspace{-0.5cm}
			(c) Our results
		\end{tabular}
	\end{center}
	\caption{Comparisons with uniform kernel based
		method \cite{vsroubek2012robust}.}
	\label{fig-multi}
\end{figure}
\begin{figure}[t]\footnotesize
	\begin{center}
		\begin{tabular}{@{}cccc@{}}
			\includegraphics[width = 0.16\textwidth]{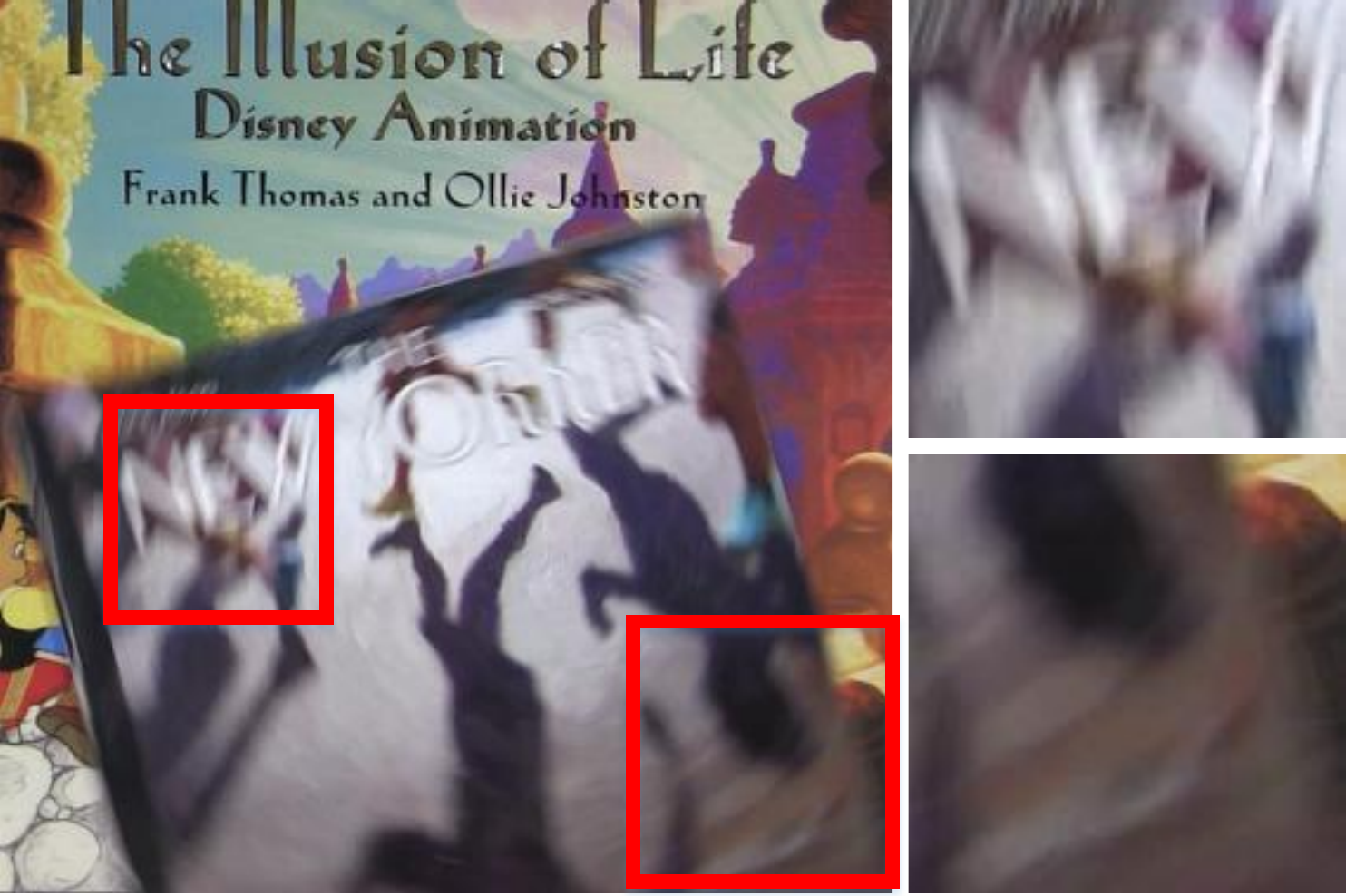} & \hspace{-0.4cm}
			\includegraphics[width = 0.16\textwidth]{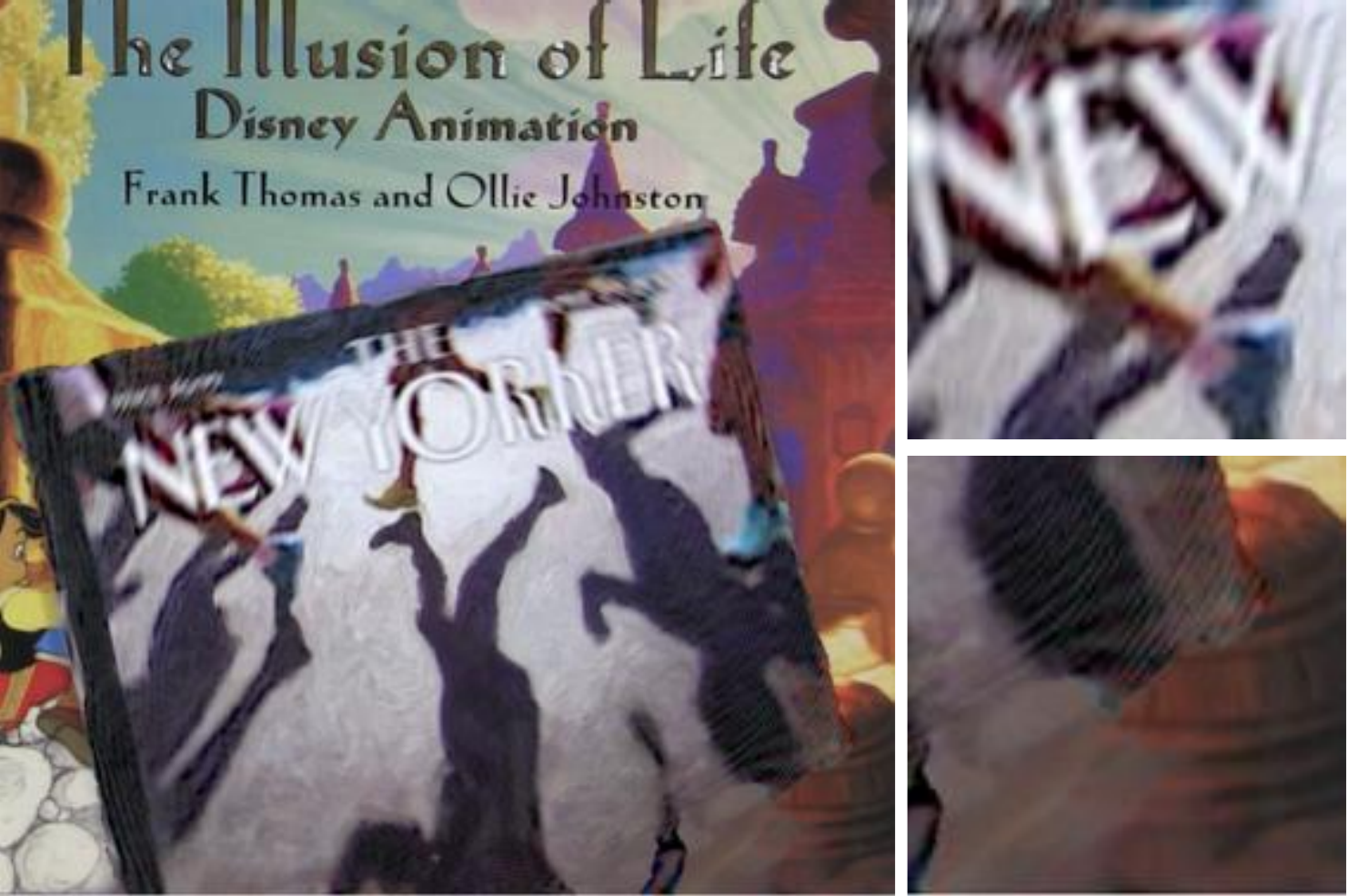} & \hspace{-0.4cm}
			\includegraphics[width = 0.16\textwidth]{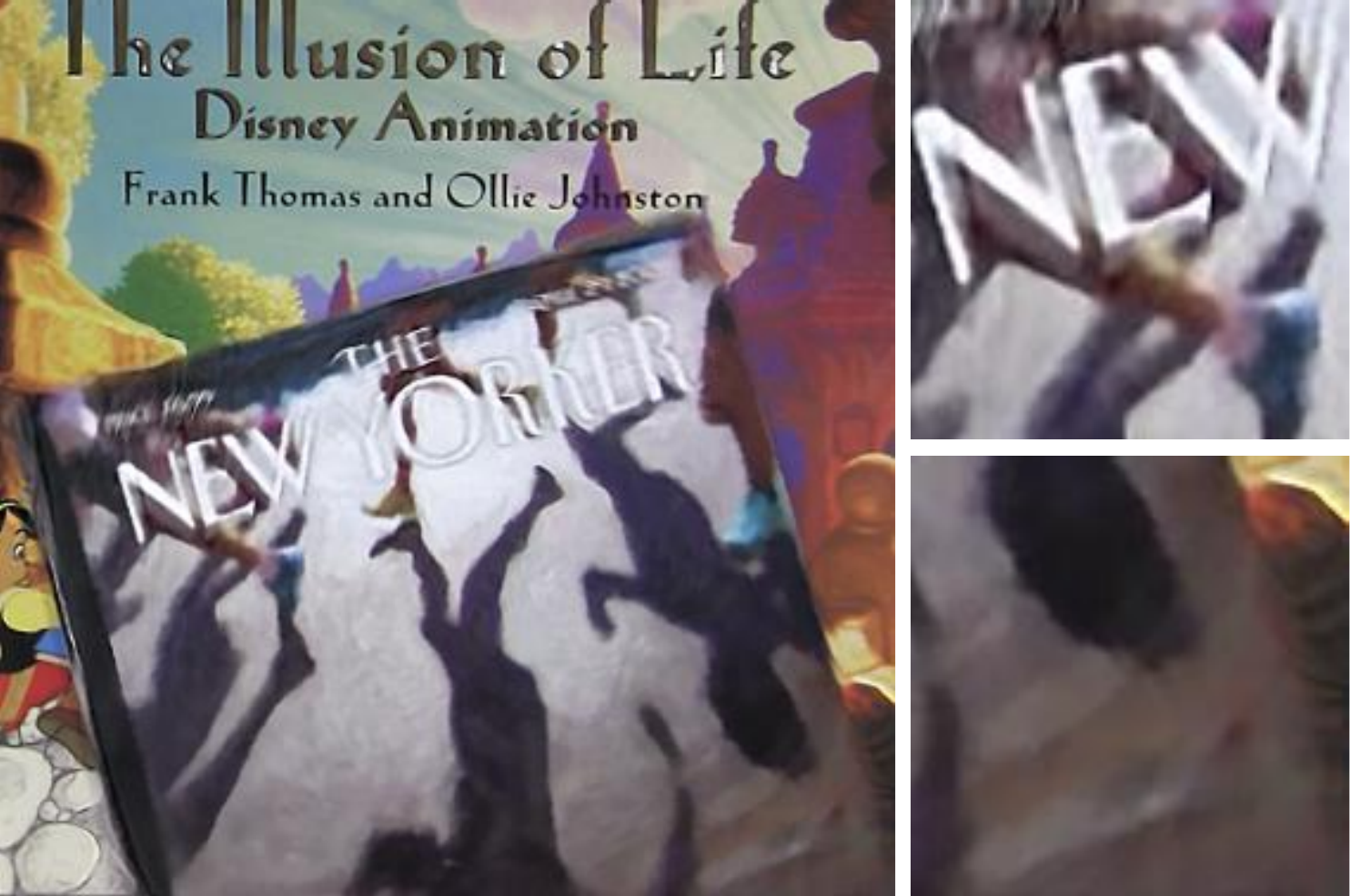} \\
			(a) Blurry frames & \hspace{-0.4cm}
			(b) Wulff and Black \cite{wulff2014modeling} & \hspace{-0.4cm}
			(c) Our results
		\end{tabular}
	\end{center}
	\caption{Comparisons with piece-wise kernel (segmentation) based video deblurring
		method \cite{wulff2014modeling}.}
	\label{fig-layer}
\end{figure}
\begin{figure}[t]\footnotesize
	\begin{center}
		\begin{tabular}{@{}ccc@{}}
			\includegraphics[width = 0.16\textwidth,height = 0.16\textwidth]{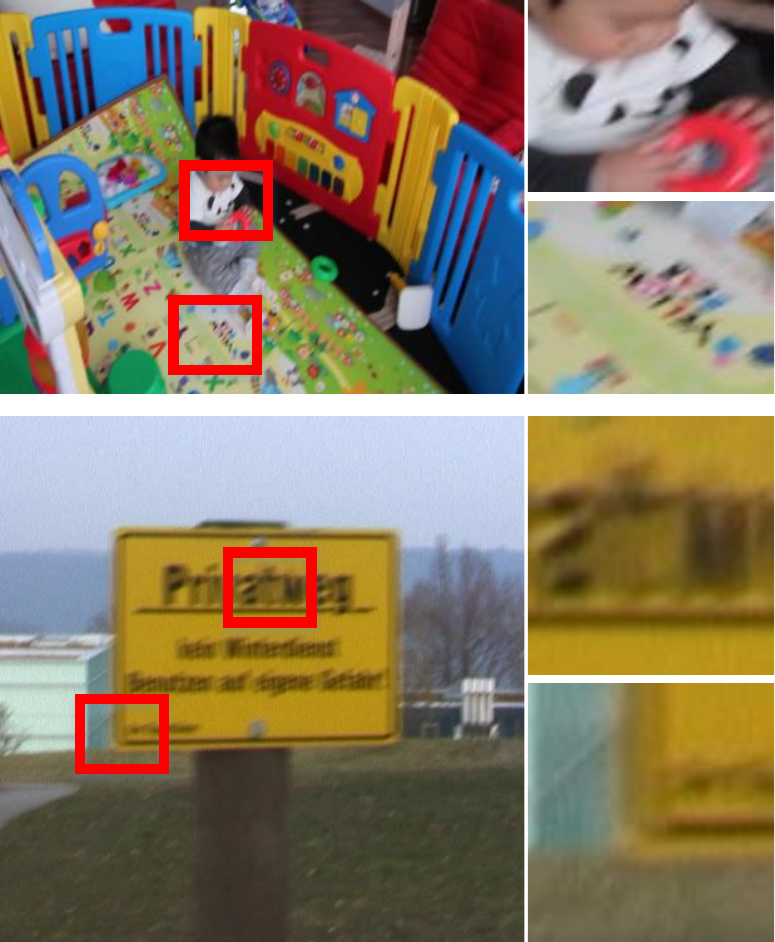} & \hspace{-0.4cm}
			\includegraphics[width = 0.16\textwidth,height = 0.16\textwidth]{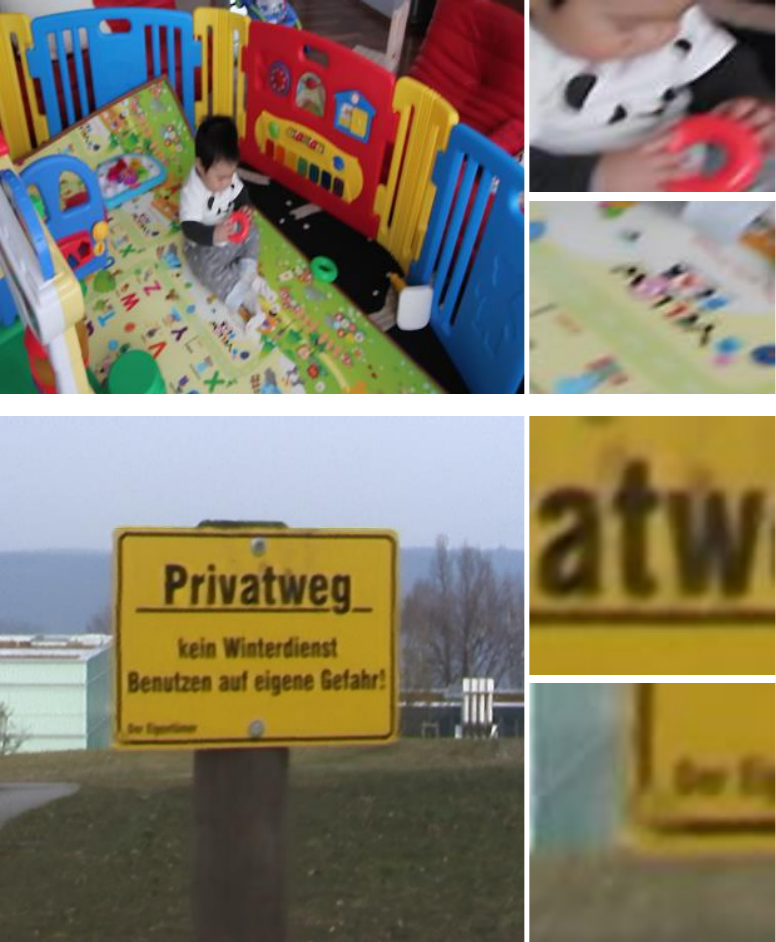} & \hspace{-0.4cm}
			\includegraphics[width = 0.16\textwidth,height = 0.16\textwidth]{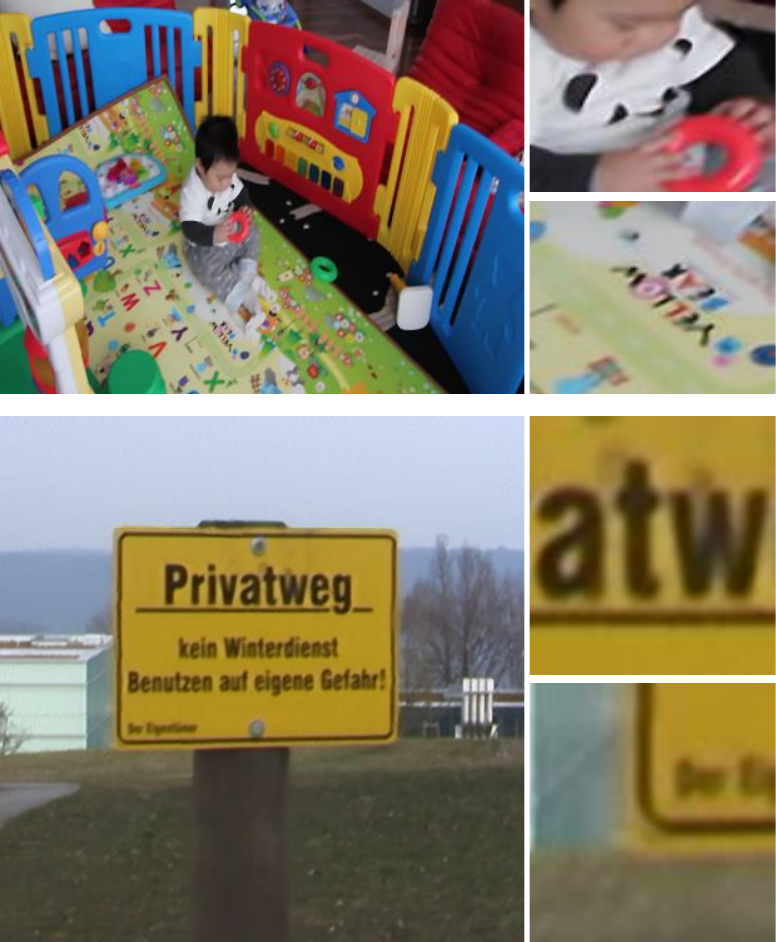} \\
			(a) Blurry frames & \hspace{-0.4cm}
			(b) Kim and Lee \cite{kim2015generalized} & \hspace{-0.4cm}
			(c) Our results
		\end{tabular}
	\end{center}
	\caption{Comparisons with pixel-wise linear kernel based video deblurring method \cite{kim2015generalized}.}
	\label{fig-kim}
\end{figure}
\begin{figure*}[!t]\scriptsize
	\begin{center}
		\begin{tabular}{@{}cccccccc@{}}
			\includegraphics[width = 0.15\textwidth,height = 0.19\textwidth]{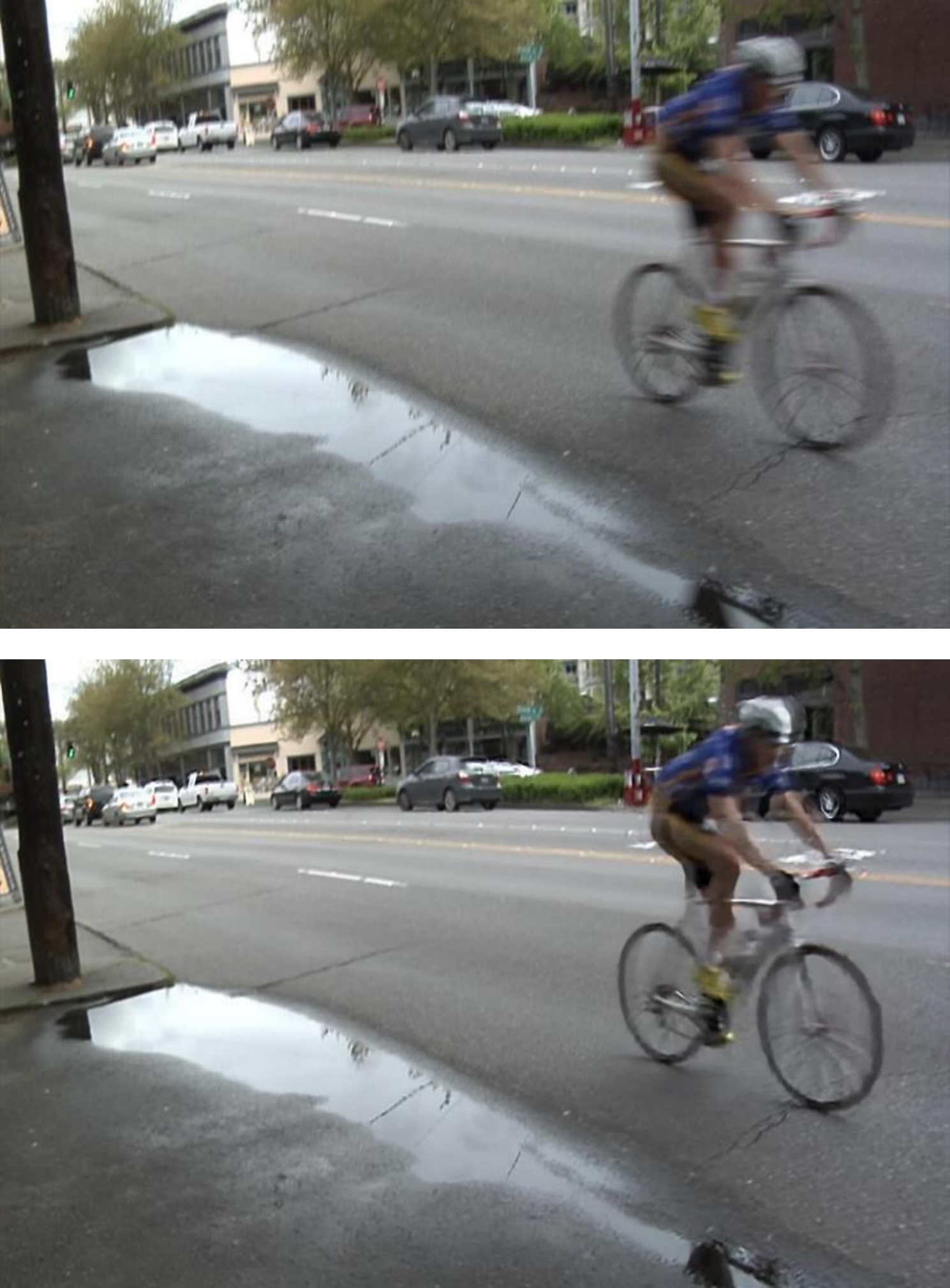} & \hspace{-0.4cm}
			\includegraphics[width = 0.12\textwidth,height = 0.19\textwidth]{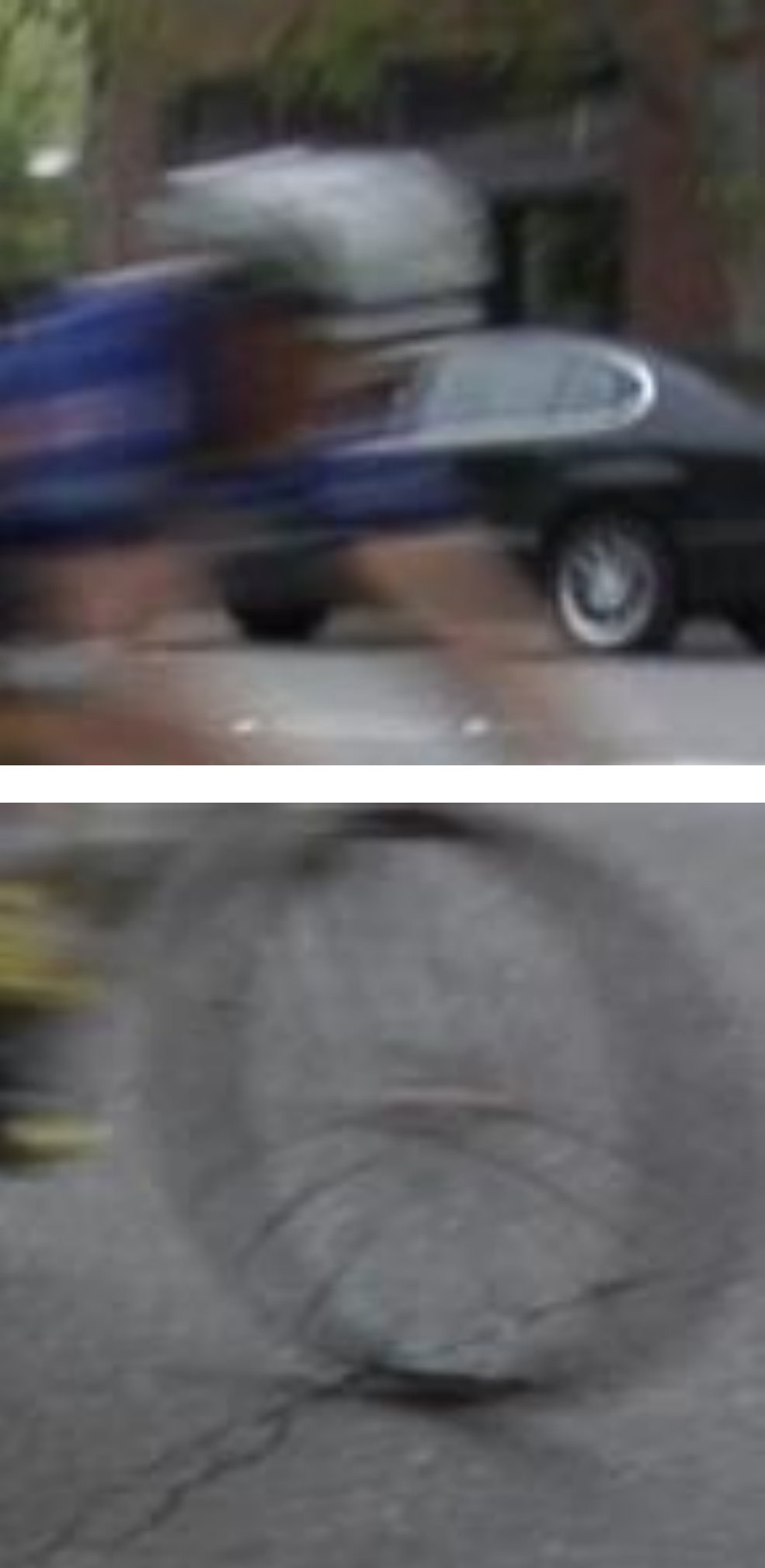} & \hspace{-0.4cm}
			\includegraphics[width = 0.12\textwidth,height = 0.19\textwidth]{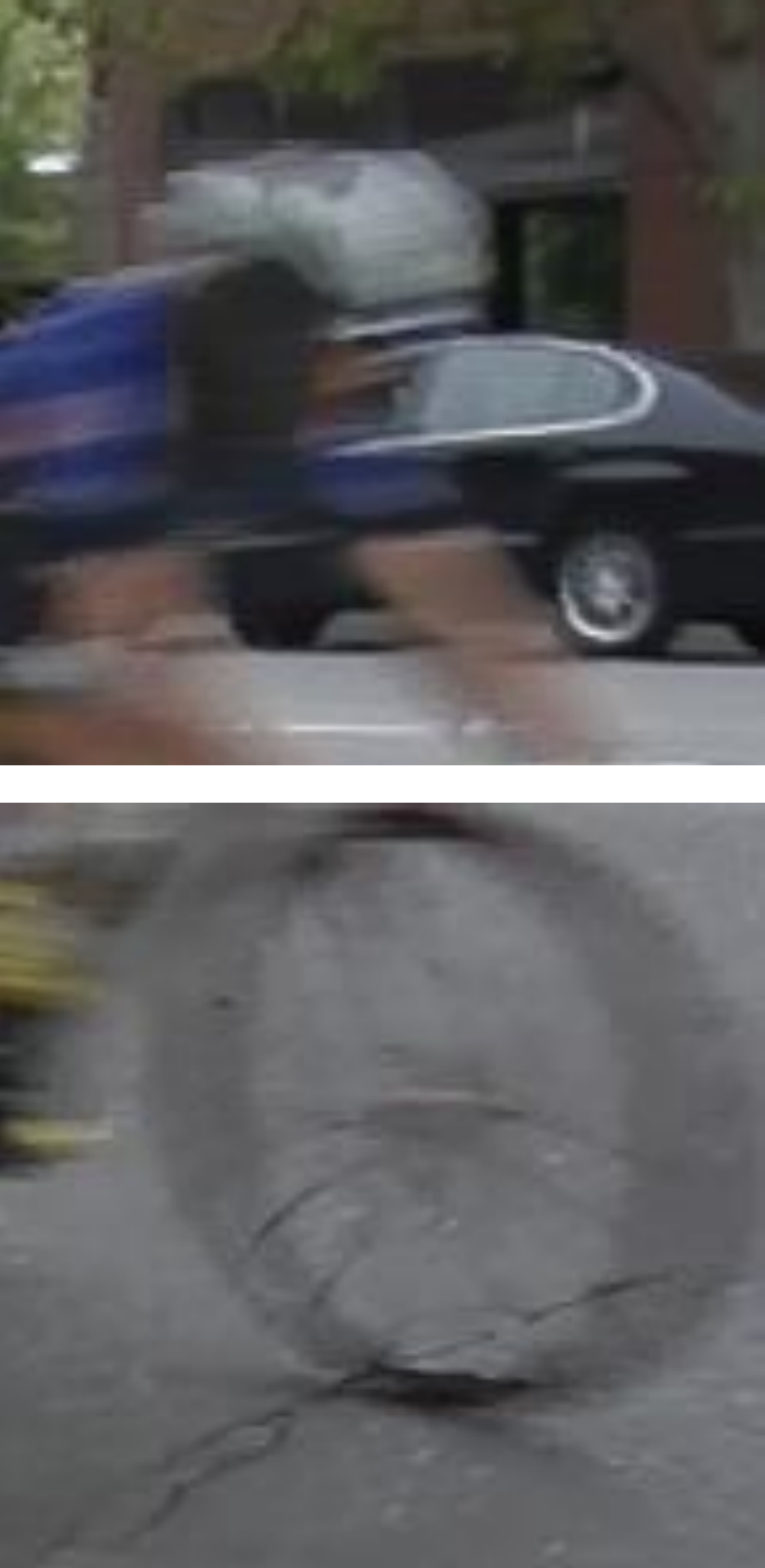} & \hspace{-0.4cm}
			\includegraphics[width = 0.12\textwidth,height = 0.19\textwidth]{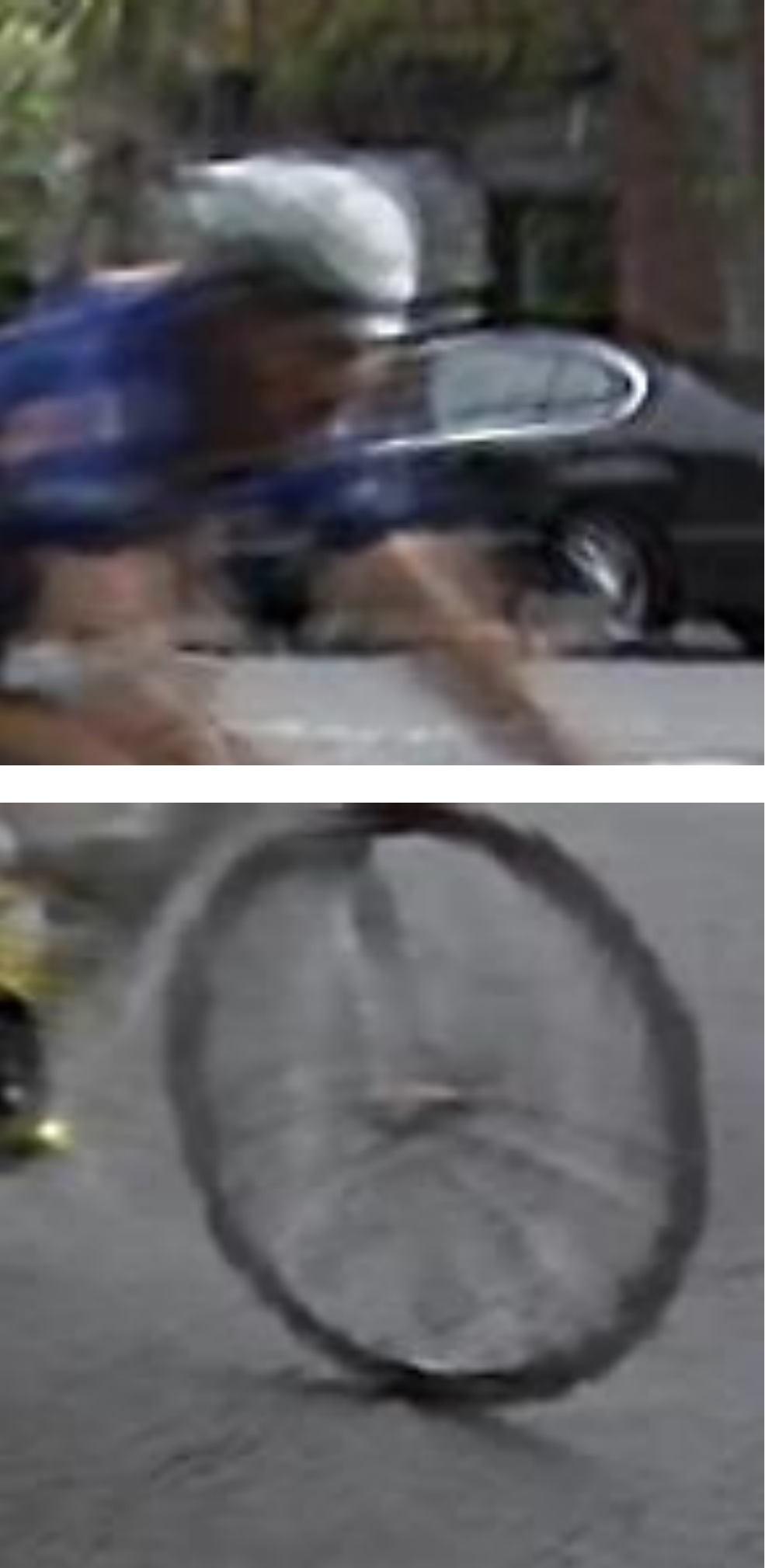} & \hspace{-0.4cm}
			\includegraphics[width = 0.12\textwidth,height = 0.19\textwidth]{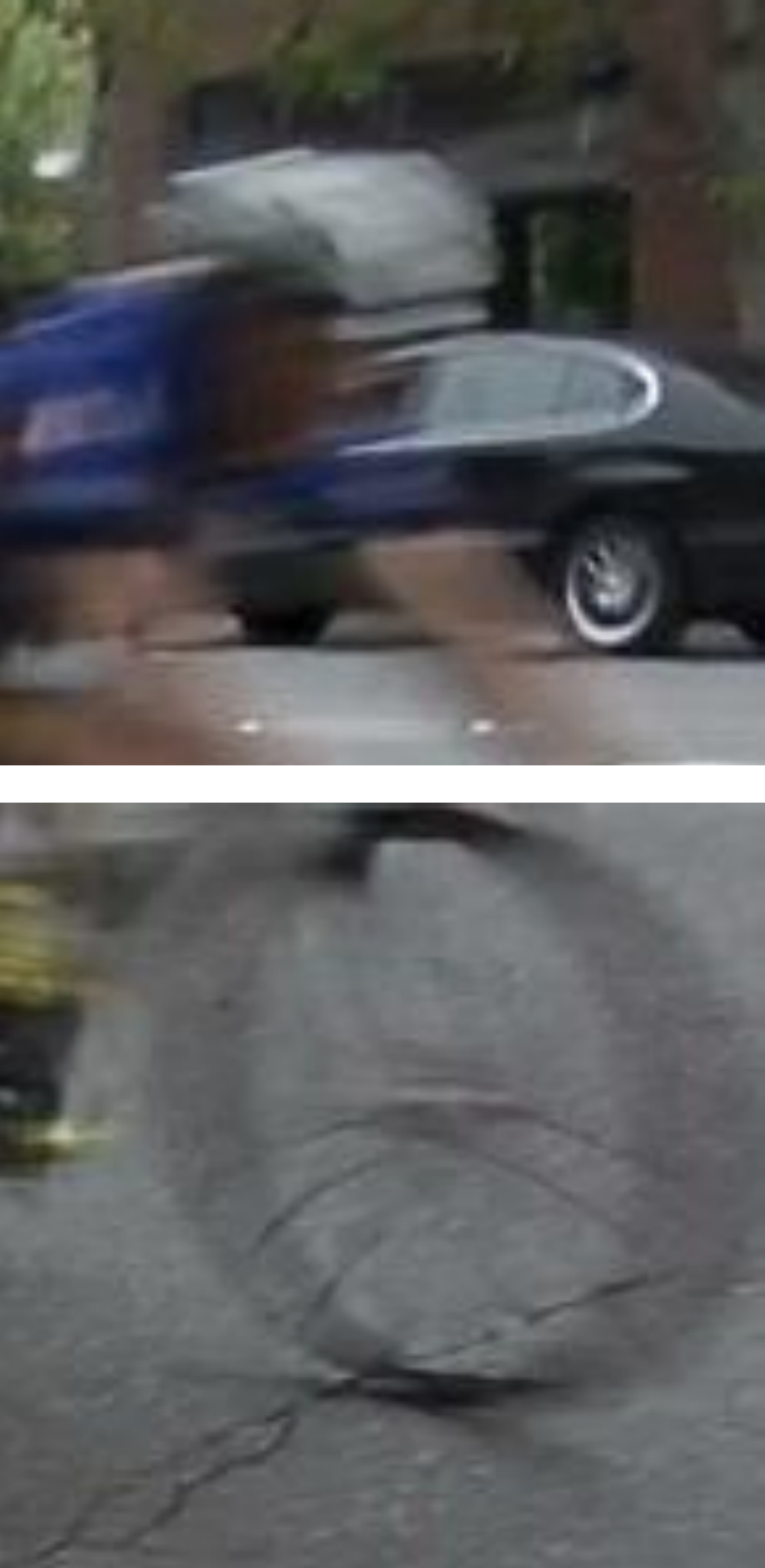} & \hspace{-0.4cm}
			\includegraphics[width = 0.12\textwidth,height = 0.19\textwidth]{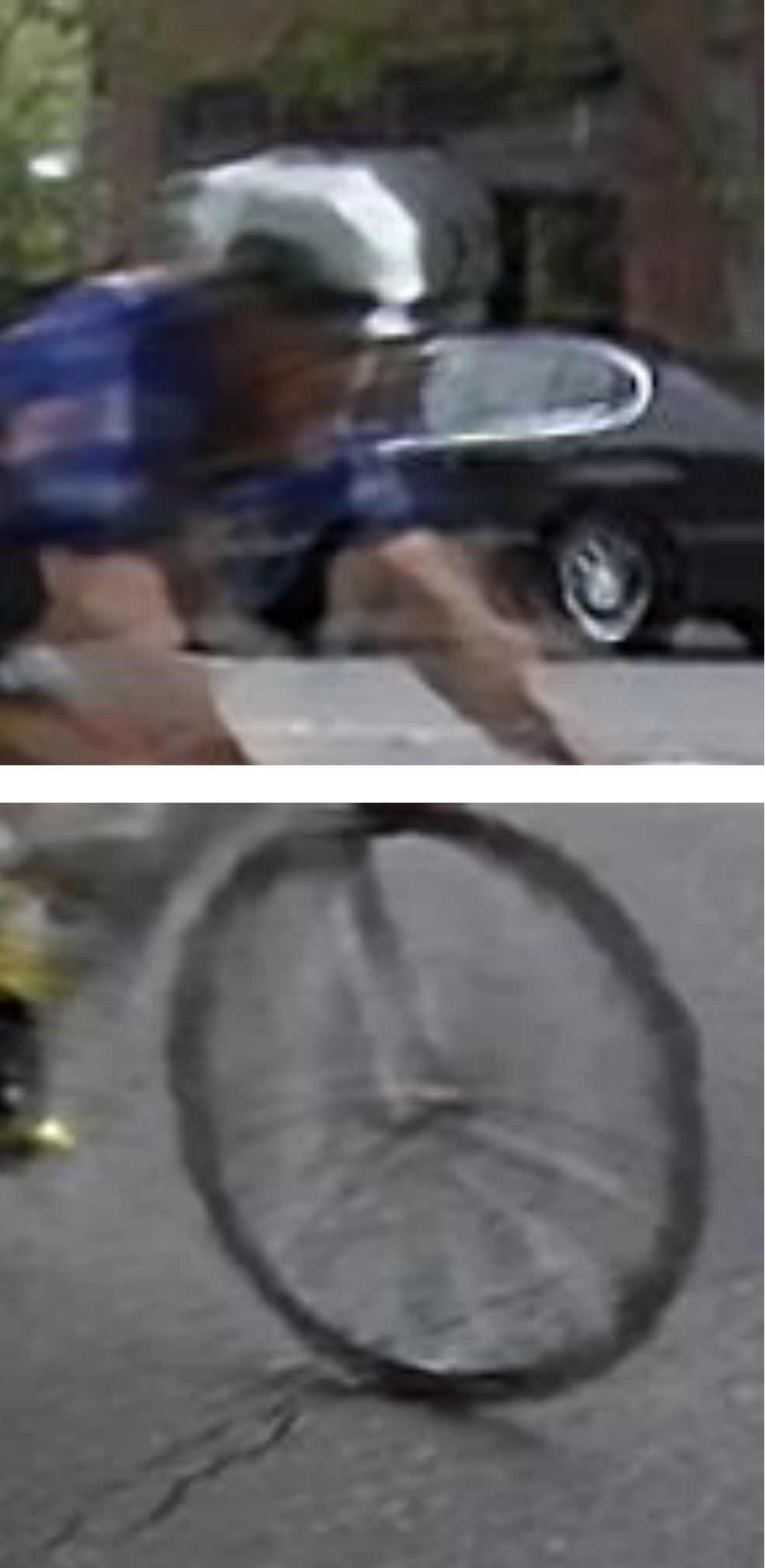} & \hspace{-0.4cm}
			\includegraphics[width = 0.12\textwidth,height = 0.19\textwidth]{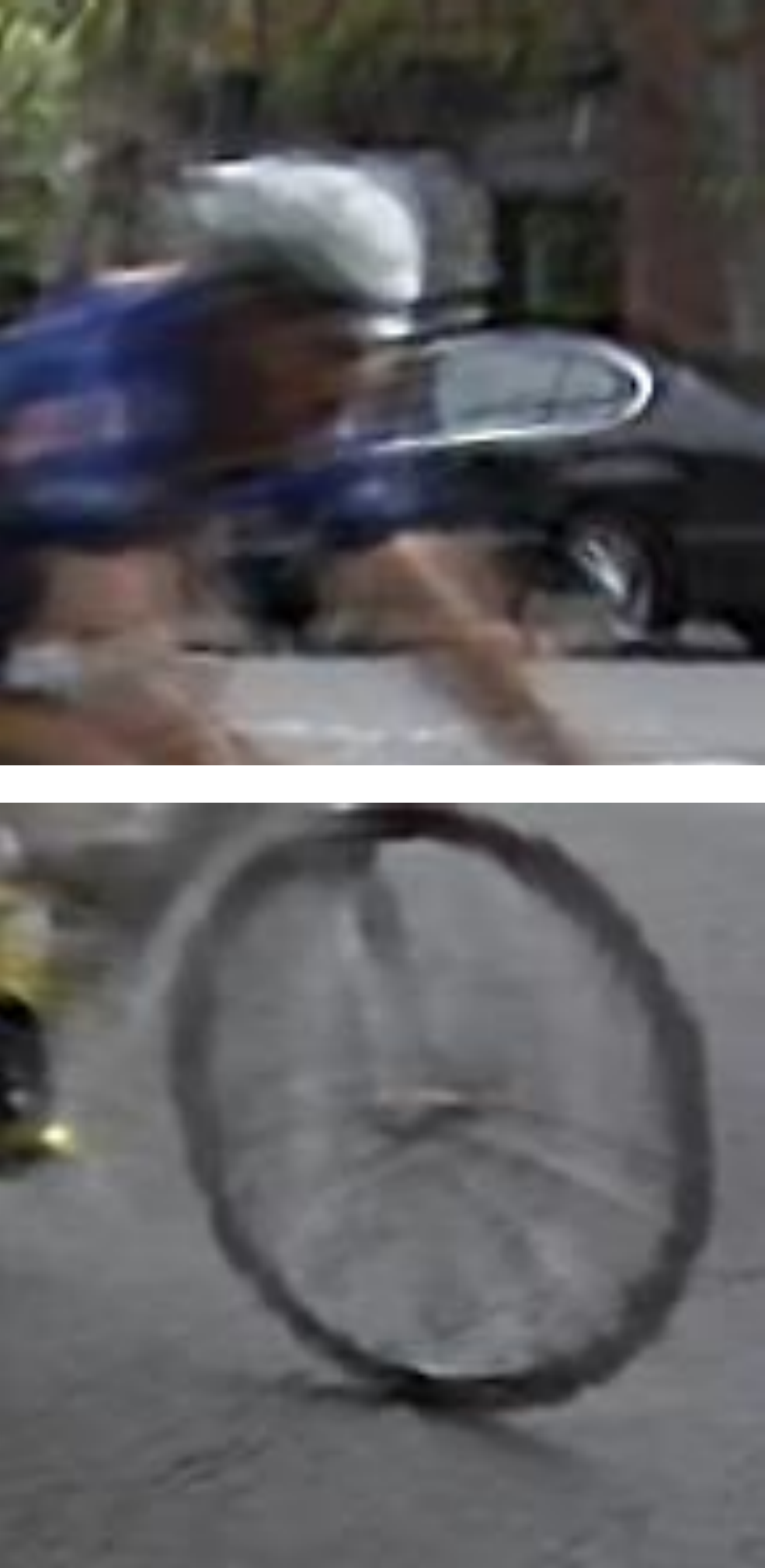} & \hspace{-0.4cm}			
			\includegraphics[width = 0.12\textwidth,height = 0.19\textwidth]{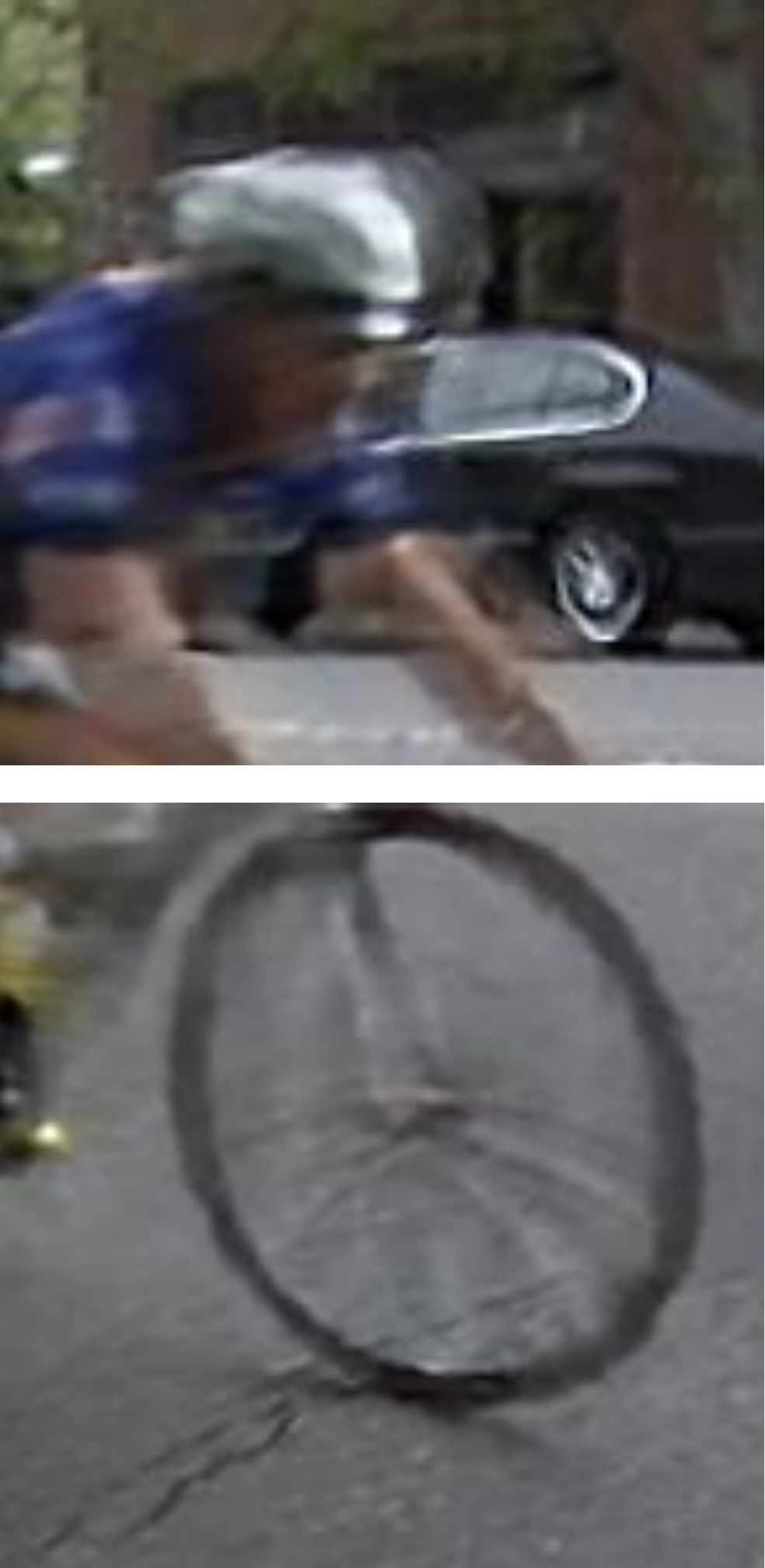} \\
			(a) Input / Our result & \hspace{-0.4cm}
			(b) Input & \hspace{-0.4cm}
			(c) Cho~\etal \cite{cho2012video} & \hspace{-0.4cm}
			(d) Kim and Lee \cite{kim2015generalized}  & \hspace{-0.4cm}
			(e) Su~\etal~\cite{su2016deep} & \hspace{-0.4cm}
			(f) Without PWNLK & \hspace{-0.4cm}
			(g) Without segment & \hspace{-0.4cm}
			(h) Our results
		\end{tabular}
	\end{center}
    \vspace{-2mm}
	\caption{Deblurred results with and without the PWNLK model and semantic segmentation.}
	\label{fig-without}
\end{figure*}

We compare the proposed algorithm with the uniform kernel based multi-image deblurring
method \cite{vsroubek2012robust}.
%
%
On the \textit{Street} sequence,
the sign PAY HERE and the structure of the windows can be clearly recognized
from the deblurred image by the proposed algorithm,
while the one by the multi-image based method does not recover such details.
%
Furthermore, our method recovers clear edges and details in the \textit{Kid} sequence.
However, the multi-image based deblurring method does not generate clear images.
The main reason is that the uniform kernels estimated
by the multi-image based method do not account for complex scenes with non-uniform blur.
In addition, the deblurred results of this multi-image deblurring method
depend on  whether the alignments of adjacent frames are accurate or not.

We show the deblurred results by the proposed method and segmentation based video deblurring approach~\cite{wulff2014modeling} in Figure~\ref{fig-layer}.
Although the deblurred image by~\cite{wulff2014modeling} is sharp,
it contains some distortion artifacts around the image boundaries
due to the inaccurate segmentations (\eg, the boundary of the \textit{Magazine}
on the right-bottom corner in Figure~\ref{fig-layer}(b)).
In contrast, the deblurred image in Figure~\ref{fig-layer}(c) shows that
proposed method is able to recover the clear edge of the \textit{Magazine}.
In addition, the recovered text \textit{NEW} in the foreground layer by Wulff and Black~\cite{wulff2014modeling} is blurry  compared to the result
generated by the proposed algorithm.

We compare the proposed algorithm with the state-of-the-art video deblurring method
based on pixel-wise linear kernel by Kim and Lee \cite{kim2015generalized}.
The deblurred results by \cite{kim2015generalized} contain blurry edges
and distortion artifacts as shown in Figure \ref{fig-kim}(b).
For example, due to the inaccurate kernel estimation, the deblurred result by \cite{kim2015generalized}
has distortion artifacts around the left-bottom corner of the \textit{Sign}
in the second row of Figure~\ref{fig-kim}(b).
%
%
In contrast, as the proposed motion blur model is able to approximate the true motion blur trajectories, the recovered images contain fine details.
Note that in Figure~\ref{fig-kim}(c), the deblurred texts in both first and second rows by the
proposed algorithm are clearer and sharper.

Finally, we show the deblurred results with and without the PWNLK model
and semantic segmentation, and compare with the state-of-the-art
transformation based~\cite{cho2012video}, deconvolution based~\cite{kim2015generalized} and deep learning based~\cite{su2016deep} video deblurring methods
in Figure~\ref{fig-without}.
The state-of-the-art video deblurring methods \cite{cho2012video,su2016deep}
do not generate clear images as shown in Figure~\ref{fig-without}(c) and (e).
Pixel-wise linear kernel based method~\cite{kim2015generalized} can generate sharp image, but the road region is over-smoothed as show in the bottom line in Figure~\ref{fig-without}(d).
In Figure~\ref{fig-without}(f), the road region is successfully recovered, but there are some visual artifacts around the tire due to imperfect kernel estimation.
Figure~\ref{fig-without}(g) shows the deblurred result
without performing semantic segmentation.
Although the tire is deblurred well, the road region is over-smoothed.
Compared to the image shown in (h),
the visual quality of (f) and (g) is lower, which indicates
the importance of the proposed PWNLK model~\eqref{eq-motion-blur-model} and
semantic segmentation regularization.

\subsection{Limitations}
\begin{figure}[t]\scriptsize
	\begin{center}
		\begin{tabular}{@{}cccc@{}}
			\includegraphics[width = 0.23\linewidth]{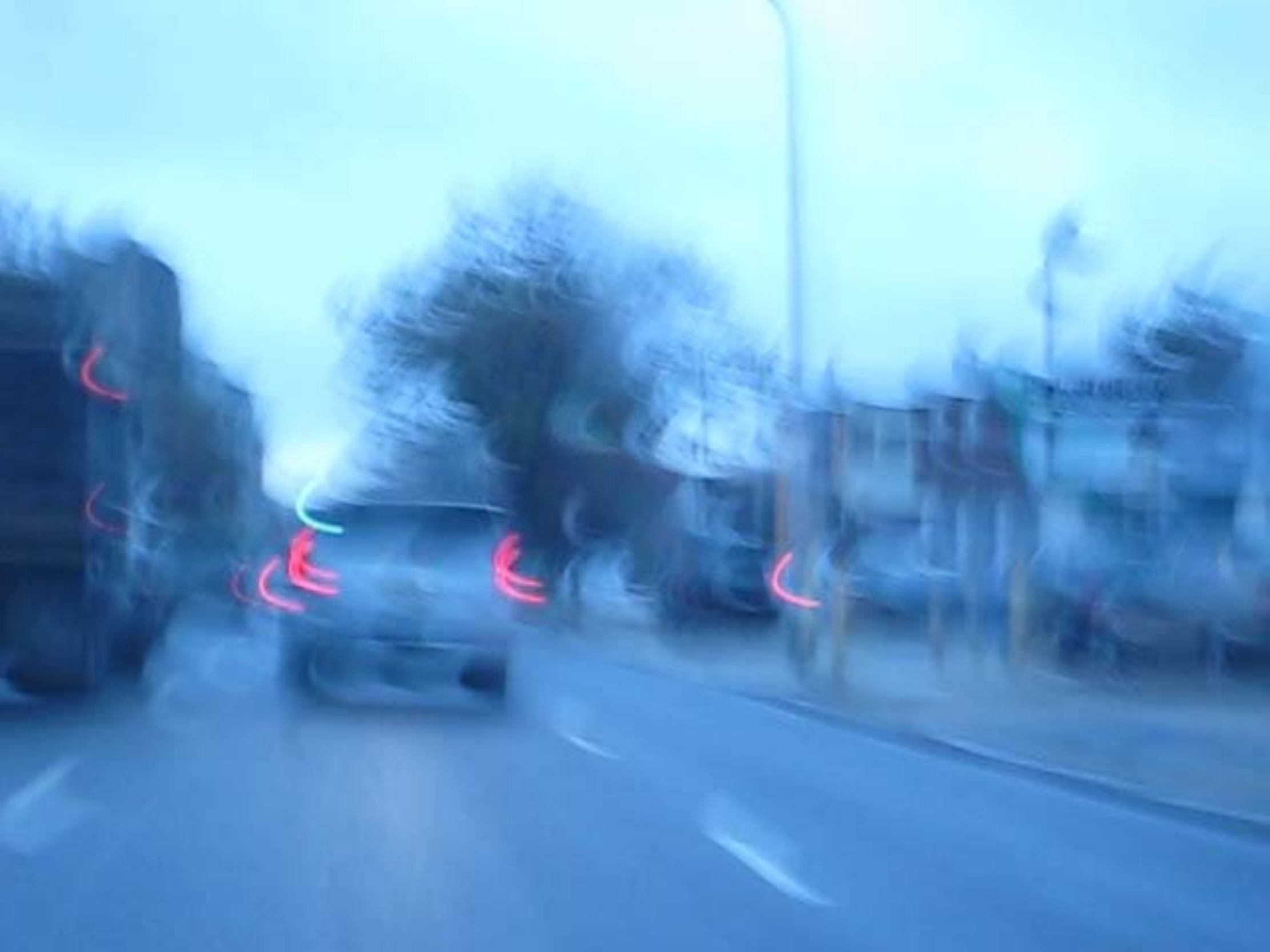}& \hspace{-0.4cm}
			\includegraphics[width = 0.23\linewidth]{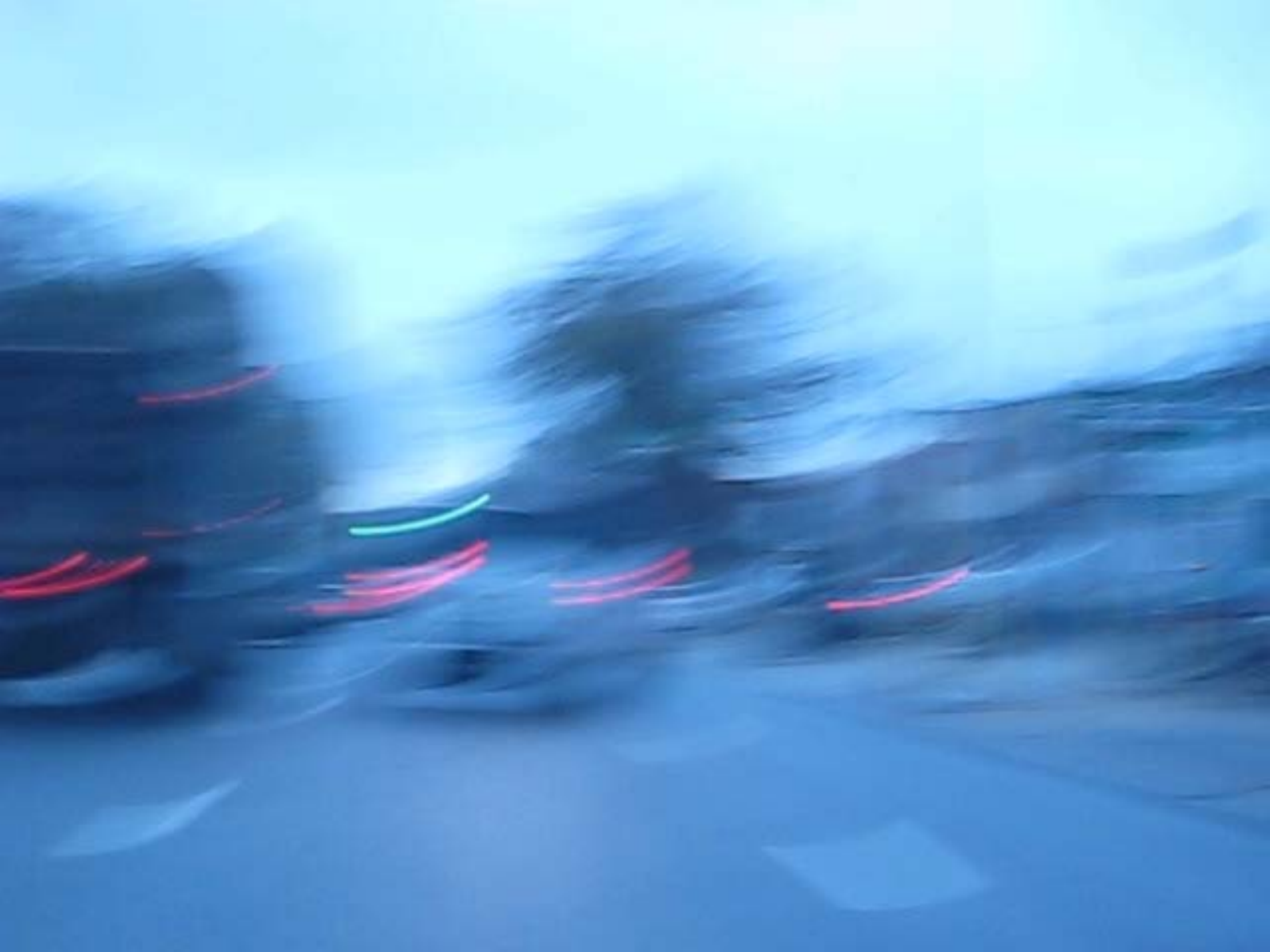}& \hspace{-0.4cm}
			\includegraphics[width = 0.23\linewidth]{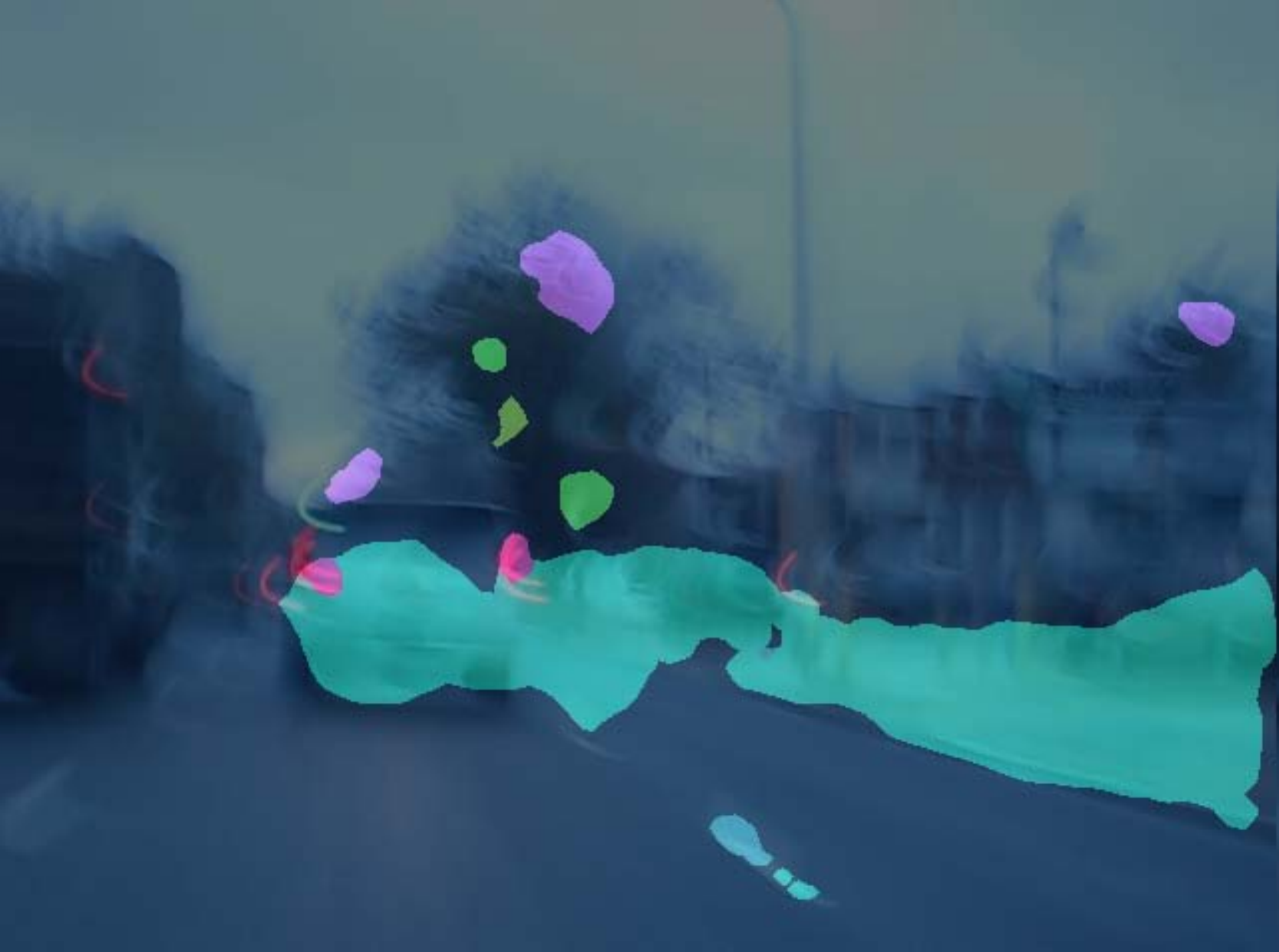}& \hspace{-0.4cm}
			\includegraphics[width = 0.23\linewidth]{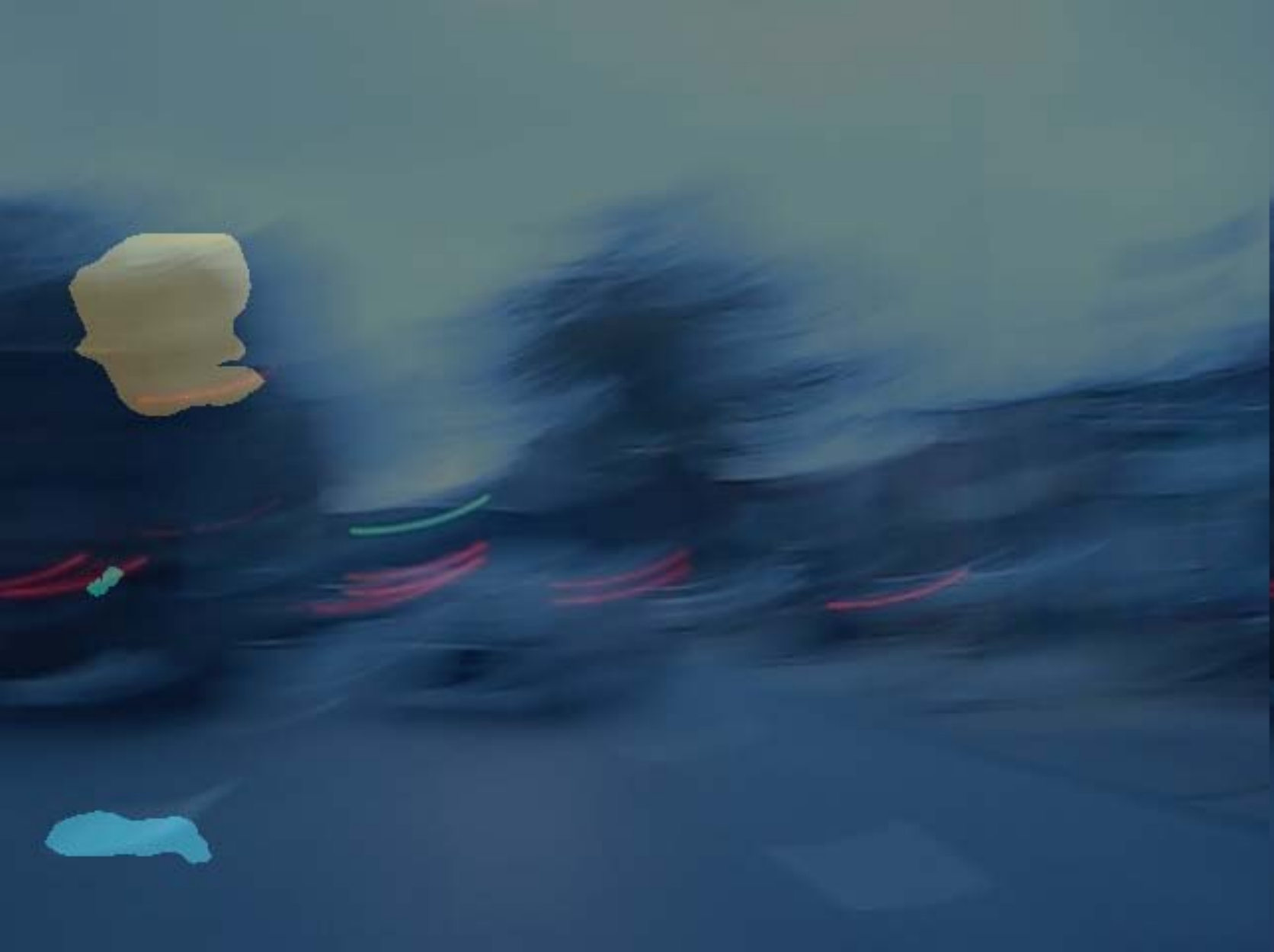}\\
			(a) Frame $l_{i-1}$ & \hspace{-0.4cm} (b) Frame $l_{i}$ & \hspace{-0.4cm} (c) Segment $l_{i-1}$ & \hspace{-0.4cm} (d) Segment $l_{i}$\\
			\includegraphics[width = 0.23\linewidth]{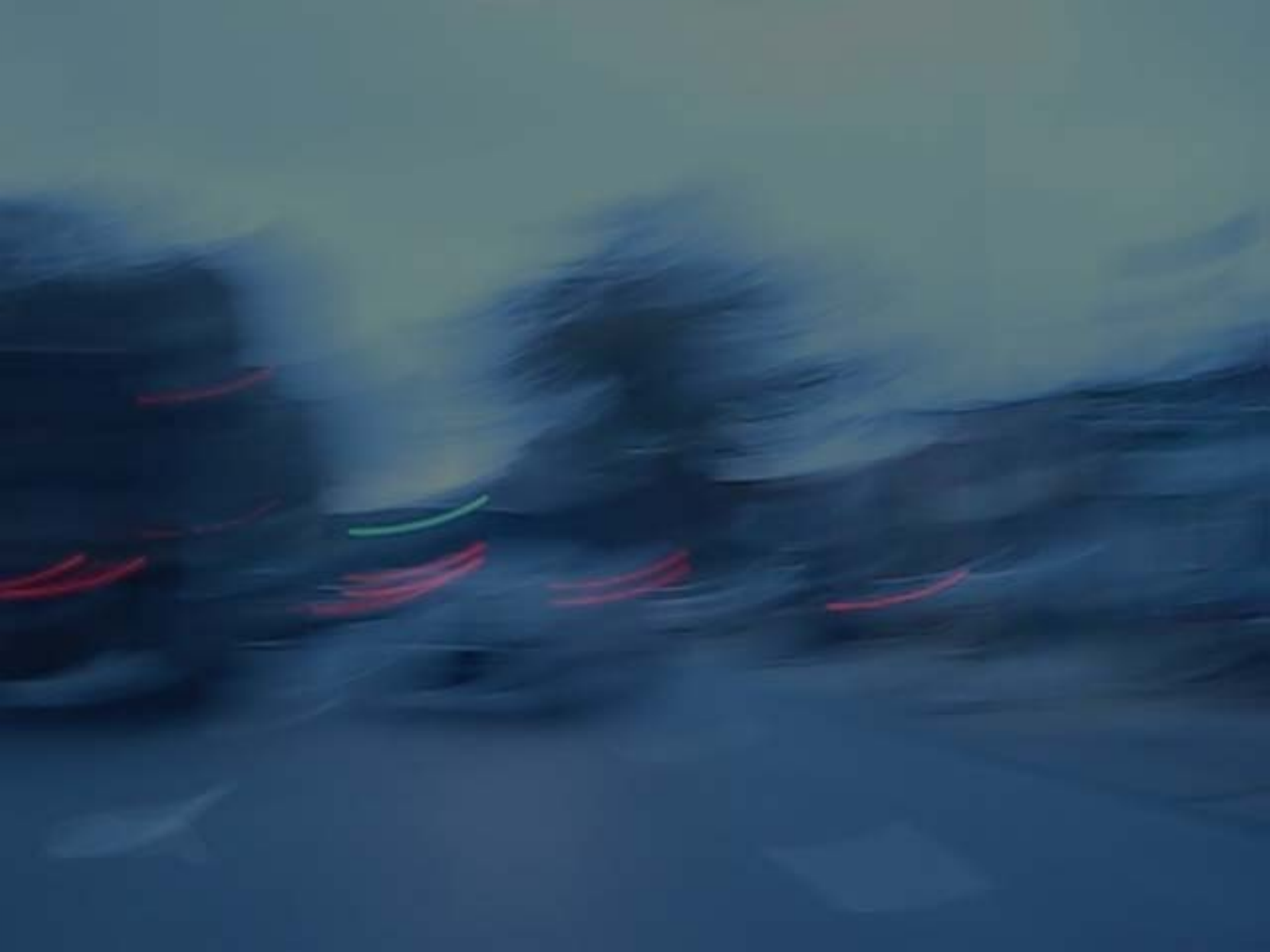}& \hspace{-0.4cm}
			\includegraphics[width = 0.23\linewidth]{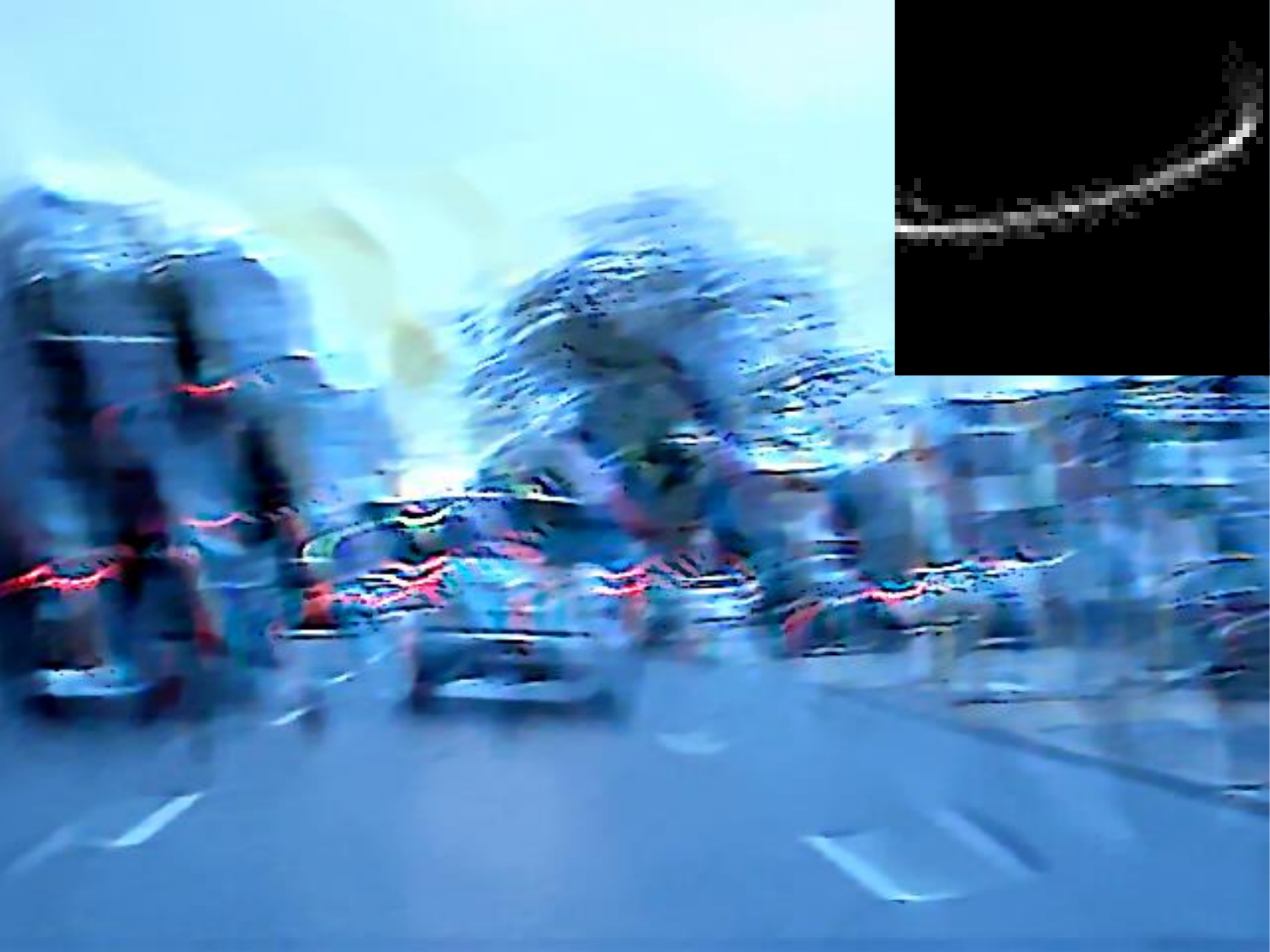}& \hspace{-0.4cm}
			\includegraphics[width = 0.23\linewidth]{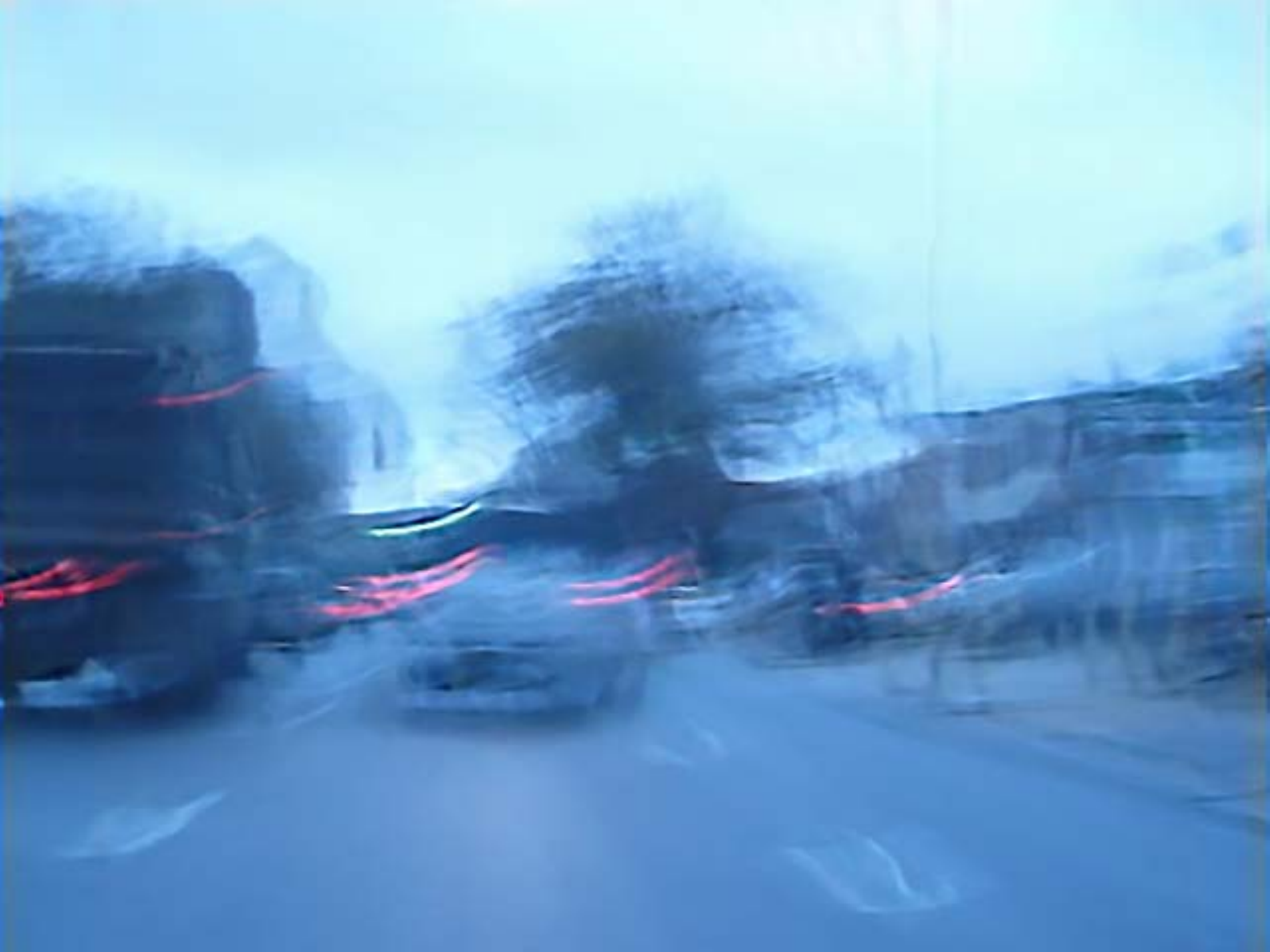}& \hspace{-0.4cm}
			\includegraphics[width = 0.23\linewidth]{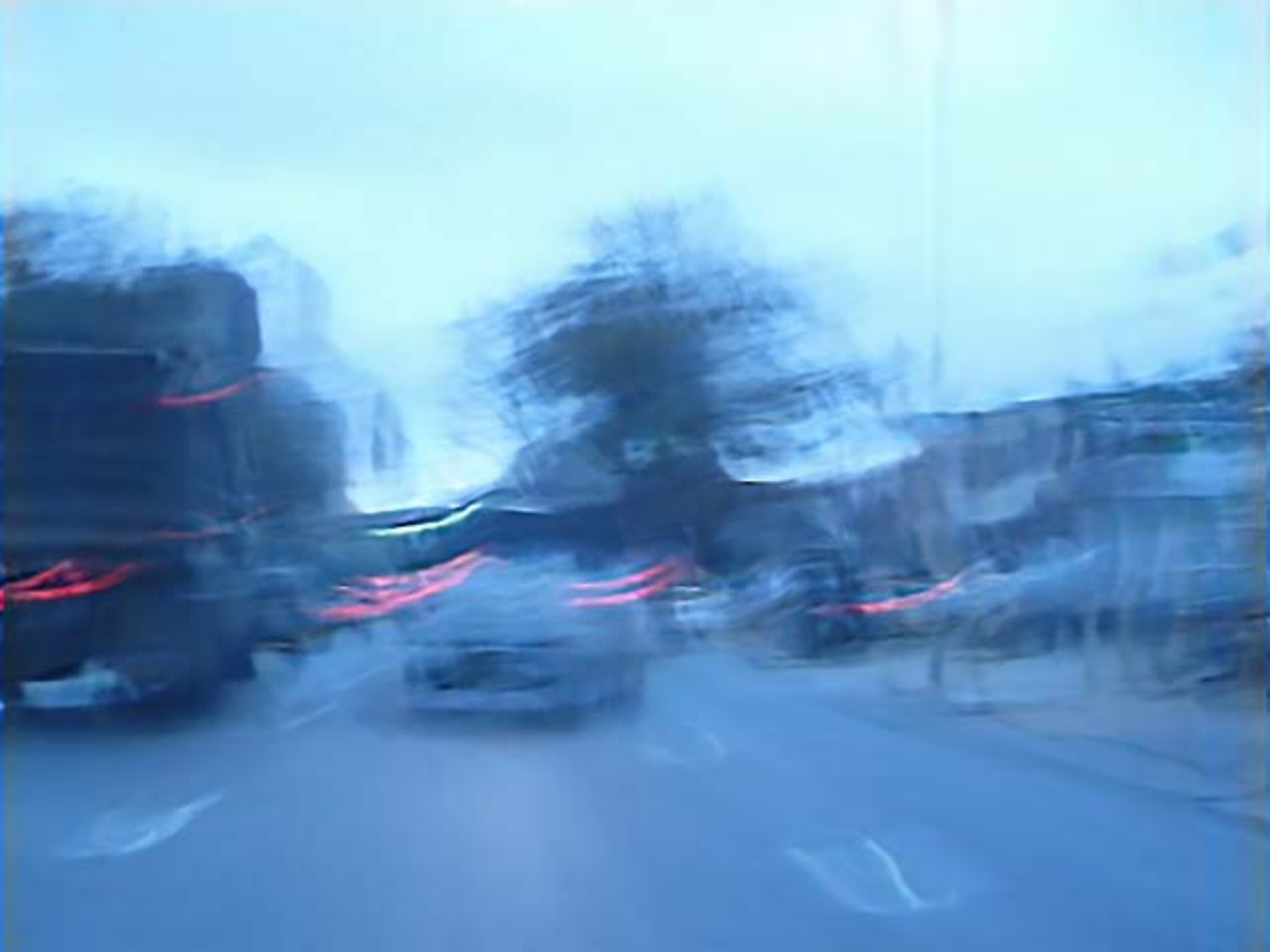} \\
			(e) Our segment & \hspace{-0.4cm} (f) Xu \cite{xuunnatural} & \hspace{-0.4cm} (g) Kim \cite{kim2015generalized} & \hspace{-0.4cm} (h) Our result\\
		\end{tabular}
	\end{center}
\vspace{-2mm}
	\caption{
		{Failure cases. (a) and (b) are blurred inputs $l_{i-1}$ and $l_i$. (c) and (d) are initialized segmentation results on frames $l_{i-1}$ and $l_i$. (e) Our final segmentation on frame $l_i$. (e)-(g) are deblurred results by \cite{xuunnatural}, \cite{kim2015generalized} and our method on frame $l_i$.}
	}
	\label{fig:failure}
\end{figure}

Our algorithm does not performs well when the input video contains significant blur along with bad initial segmentations.
Figure~\ref{fig:failure}(c) and (d) are the initial segmentation results for the consecutive blurry frame
Figure~\ref{fig:failure}(a) and (b), respectively.
Since the assumed spatial and temporal constraints in \eqref{eq-corrof} and \eqref{eq-temporal} do not
hold in the segmented image, the final segmentation result
in Figure~\ref{fig:failure}(e) does not have any semantic information. Thus,
our method degenerates to traditional optical flow estimation in \cite{kim2015generalized} and generate similar deblurred results as shown
in Figure~\ref{fig:failure}(g) and (h).
%

\section{Conclusions}
In this paper, we propose an effective video deblurring algorithm
by exploiting semantic segmentation and PWNLK model.
The proposed segmentation applies different motion model to different object layers,
which can significantly improve
the optical flow estimation, especially at object boundaries.
The PWNLK model is based on the non-linear assumption and is able to model the relationship between motion blur and optical flow.
In addition, we analyze that conventional uniform, homography, piece-wise, pixel-wise linear based blur kernels cannot model the complex spatially variant blur
caused by the combination of camera shakes, objects motions and depth variations.
Extensive experimental results on synthetic and real videos
show that the proposed algorithm performs favorably
in video deblurring against the state-of-the-art methods.


{\flushleft \textbf{Acknowledgments.}}
This work is supported in part by the National Key R\&D Program of China (No. 2016YFB0800403), National Natural Science Foundation of China
(No. 61422213, U1636214), Key Program of the Chinese Academy of Sciences (No. QYZDB-SSW-JSC003). %
Ming-Hsuan Yang is supported in part by the NSF CAREER (No. 1149783), gifts from Adobe and Nvidia.
Jinshan Pan is supported by the 973 Program (No. 2014CB347600), NSFC (No. 61522203), NSF of Jiangsu Province (No. BK20140058), National Key R\&D Program of China (No. 2016YFB1001001).

{\small
\bibliographystyle{ieee}
\bibliography{egbib_videodeblur}
}

\end{document}